%% file: main.tex
\documentclass{article}

\let\paperaddcontentsline\addcontentsline
\usepackage{iclr2027_conference,times}
\let\addcontentsline\paperaddcontentsline
\iclrfinalcopy 

\usepackage[utf8]{inputenc} 
\usepackage[T1]{fontenc}    
\usepackage{hyperref}       
\usepackage{url}            
\usepackage{booktabs}       
\usepackage{amsfonts}       
\usepackage{nicefrac}       
\usepackage{microtype}      
\usepackage{xcolor}         
\usepackage{svg}
\usepackage{graphicx}
\usepackage[most]{tcolorbox}
\usepackage{subcaption}
\usepackage{multirow}
\usepackage[ruled,vlined]{algorithm2e}
\usepackage{enumitem}
\usepackage{titletoc}
\usepackage{titlesec}
\usepackage{caption}
\usepackage{fontawesome5}

\usepackage{footmisc}
\titlespacing\section{0pt}{1.8ex plus 0.0ex minus 0.3ex}{1.3ex plus 0.0ex minus 0.2ex}
\titlespacing\subsection{0pt}{1.8ex plus 0.1ex minus 0.3ex}{1.2ex plus 0.1ex}
\titlespacing\paragraph{0pt}{-0.2ex}{1.1ex plus 0.3ex}

\usepackage{pifont}
\newcommand{\cmark}{\ding{51}}
\newcommand{\xmark}{\ding{55}}
\newcommand{\pmark}{$\circ$}      

\newcommand{\logic}[1]{{\color{teal}\texttt{#1}}}

\newcommand{\projectpage}[1]{
  \href{#1}{\faGlobe\ \textcolor{blue}{Project Page}}
}

\newcommand{\githubrepo}[1]{
  \href{#1}{\faGithub\ \textcolor{blue}{GitHub}}
}

\newtcolorbox{examplebox}[1][]{
  enhanced,
  breakable,
  colback=white,
  colframe=blue!50!black,
  colbacktitle=blue!8,
  coltitle=black,
  boxrule=0.6pt,
  arc=0.6mm,
  fonttitle=\bfseries,
  title=#1,
  attach boxed title to top left={xshift=2mm,yshift=-2mm},
  top=2.0mm,
  bottom=2.0mm,
  left=2.0mm,
  boxsep=1.0mm,
  before skip=2.0mm,
  boxed title style={
    boxrule=0.5pt,
    colframe=blue!50!black,
    colback=blue!8,
    arc=0.6mm,
    top=0.5mm,
    bottom=0.5mm,
  }
}

\title{How Deep Can LLMs Learn to Reason?\\ Expressiveness Is Key}

\author{%
  \textbf{Tianle Wang}$^{1}$ \quad
  \textbf{Zhaoyang Wang}$^{2}$ \quad
  \textbf{Guangchen Lan}$^{1}$ \quad
  \textbf{Xinpeng Wei}$^{3}$ \\[0.1em]
  \textbf{Sipeng Zhang}$^{4}$ \quad
  \textbf{Guanwen Qiu}$^{1}$ \quad
  \textbf{Abulhair Saparov}$^{1}$ \\[0.4em]
  $^{1}$Purdue University \quad
  $^{2}$UNC Chapel Hill \quad
  $^{3}$Georgia Tech \quad
  $^{4}$UC San Diego \\[0.4em]
  \texttt{wang7327@purdue.edu}\\[0.4em]
  {\small
\projectpage{https://wtl666wtl.github.io/projects/scalelogic}
\quad
\githubrepo{https://github.com/wtl666wtl/ScaleLogic}
}
}

\begin{document}

\maketitle
\lhead{Preprint}

\begin{abstract}
Reinforcement learning (RL) has been applied to improve large language model (LLM) reasoning, yet the systematic study of how training scales with task difficulty has been hampered by the lack of controlled, scalable environments. Observed LLM shortcomings in long-horizon reasoning have raised the prospect that they are fundamental to the autoregressive transformer architecture. To address this, we introduce \textsc{ScaleLogic}, a synthetic logical reasoning framework that offers independent control over two axes of difficulty: the depth of the required proof planning (i.e., the \emph{horizon}) and the expressiveness of the underlying formal system. Our proposed framework supports a wide range of logics: from simple implication-only logic (``if-then'') towards more expressive first-order reasoning with conjunction (``and''), disjunction (``or''), negation (``not''), and universal quantification (``for all''). Using this framework, we show that the RL training compute $T$ follows a power law with respect to reasoning depth $D$ ($T \propto D^{\gamma}$, $R^{2} > 0.99$), and the scaling exponent $\gamma$ increases monotonically with expressiveness, from $1.05$ to $2.60$. On downstream mathematics and general reasoning benchmarks, more expressive training settings yield both larger performance gains (up to $+10.66$ points) and more compute-efficient transfer compared to less expressive settings, demonstrating that \emph{what} a model is trained on, not just \emph{how much} it is trained, shapes downstream transfer. We further show that the power-law relationship holds across multiple RL methods, and curriculum-based training substantially improves scaling efficiency. More broadly, our results suggest that LLM shortcomings in long-horizon reasoning are not fundamental to the underlying architecture, and could be addressed by improved training methodology and data.
\end{abstract}

\section{Introduction}

Recent advances in reinforcement learning (RL) for large language model (LLM) reasoning have shown that training on verifiable domains such as mathematics and coding can elicit enhanced reasoning behavior and improve benchmark performance~\citep{guo2025deepseek,jaech2024openai}.
However, these gains do not yet translate into robust long-horizon\footnote{Throughout, we use ``horizon'' in the sense of the reasoning horizon, i.e., the number of sequential
inference steps composed before the final answer, following the usage in prior work~\citep{motwani2025h1,lu2025rhorizon}.} reasoning, as the same model's performance can degrade sharply once tasks require many sequential reasoning steps, even when the corresponding subproblems are individually manageable~\citep{rameshkumar2025reasoning,motwani2025h1,lu2025rhorizon,zhou2025gsminfinite}. This has led to the suggestion that limitations in long-horizon reasoning may be fundamental to the autoregressive transformer architecture~\citep{saparov2025transformers,DBLP:conf/icml/BachmannN24}, raising a central question about the feasibility of extending a model's reasoning horizon through reinforcement learning:

\textit{As reasoning depth increases, can RL keep extending a model's reasoning horizon, or does training eventually become ineffective? If it can, how does the required training effort scale with depth?}

Answering this question requires a training environment with three properties simultaneously:
exact verifiability, fine-grained control over reasoning difficulty, and data available at scale. The domains currently driving RL progress, such as mathematics and coding, often provide reliable verification, but high-quality problems are expensive to curate and offer only limited control over reasoning horizon and task structure~\citep{hendrycks2021measuring,jain2024livecodebench}. Existing alternatives, including synthetic tasks~\citep{lan2025contextual,guo2025logicrl,liu2025saturn} and self-evolving pipelines~\citep{zhao2025absolute,huang2025rzero}, reduce the cost of data collection through automatic generation and verification, but often do not provide a sufficiently clean and interpretable set of control axes for analyzing scaling behavior.
Together, these limitations make it difficult to systematically characterize how reasoning complexity shapes RL scaling (Table~\ref{tab:comparison}).

\begin{table}[t]
\centering
\small
\caption{\textbf{Comparison of training data sources for RL post-training of LLM reasoning.} Here, $\times$ denotes no direct control, $\checkmark$ denotes direct support, and $\circ$ denotes coarse or indirect control. E.g., the logic task SAT can vary the number of variables and clauses, but only indirectly controls the reasoning horizon.}
\label{tab:comparison}
\setlength{\tabcolsep}{6pt}
\renewcommand{\arraystretch}{1}
\begin{tabular}{l cccc}
\toprule
\textbf{Data Source}
  & \textbf{Verifiable}
  & \textbf{Scalable}
  & \shortstack{\textbf{Controllable}\\\textbf{Horizon}}
  & \shortstack{\textbf{Controllable}\\\textbf{Expressiveness}} \\
\midrule
Math / Code \citep[e.g., ][]{hendrycks2021measuring}   & \cmark & \xmark & \xmark & \xmark \\
Game-based  \citep[e.g., ][]{wu2024werewolf}                     & \pmark & \cmark & \xmark & \xmark \\
Self-Evolving  \citep[e.g., ][]{zhao2025absolute}           & \cmark & \cmark & \xmark & \xmark \\
Logic Puzzles \citep[e.g., ][]{liu2025saturn}                                           & \cmark & \cmark & \pmark & \xmark \\
Benchmark Suites \citep[e.g., ][]{liu2025synlogic}              & \cmark & \cmark & \pmark & \xmark \\
\midrule
\textsc{ScaleLogic}                                                       & \cmark & \cmark & \cmark & \cmark \\
\bottomrule
\end{tabular}
\vspace{-1em}
\end{table}

\begin{figure*}[t]
    \centering
    \includegraphics[width=\textwidth]{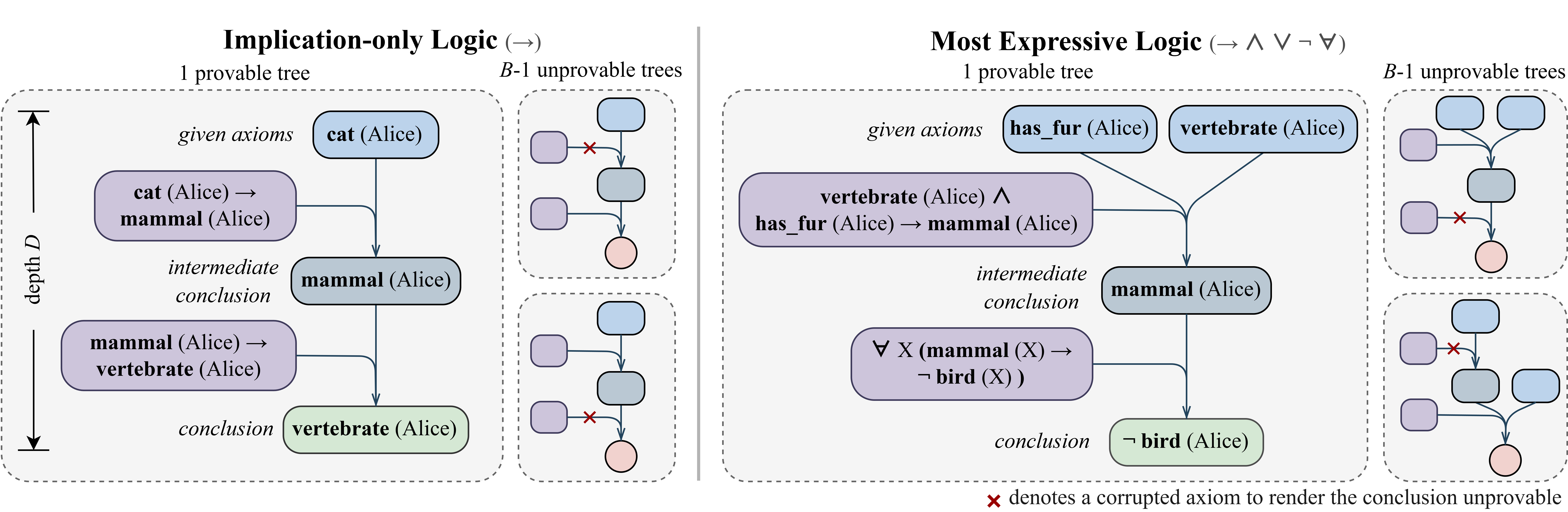}
    \caption{\textbf{Overview of \textsc{ScaleLogic}}. Each problem has $B$ candidate proof trees, exactly one of which has a provable conclusion; the others are made unprovable by corrupting one axiom. The depth $D$ controls proof depth. \textbf{Left:} Implication-only reasoning. \textbf{Right:} The most expressive logic setting (referred to as \textit{+ Quantification} in Section~\ref{sec:difficulty}) combines conjunction, disjunction, negation, and universal quantification.}
    \label{fig:overview}
    \vspace{-1em}
\end{figure*}

To address this gap, we propose \textsc{ScaleLogic}, a synthetic logical reasoning environment with explicit control over two key dimensions of difficulty: the depth of the required proof planning (i.e., the \emph{horizon}) and the logical expressiveness\footnote{In this work, we use \emph{expressiveness} to refer to the set of proofs that can be expressed using a progressively larger set of inference rules, each associated with a logical operator.} of the reasoning problems.
As illustrated in Figure~\ref{fig:overview}, in each example, the model is given a set of facts and a handful of candidate conclusions. The model is then asked to identify which candidate conclusion is logically derivable from those facts, yielding an exact and easily verifiable multiple-choice objective.
To find the correct multiple-choice option, the model must search for a proof of each candidate conclusion. Our environment requires it to expend comparable reasoning effort to determine the provability of each candidate, 
which substantially limits the model's ability to exploit simple heuristics or shortcuts during reasoning.
Beyond scaling the depth of the proof tree, our environment allows us to vary the expressive structure of the underlying formal system, including the logical operators and structural properties involved in each example. Combined with automatic low-cost generation and verification, this provides a controlled testbed for systematically studying RL scaling in long-horizon reasoning.

Experiments in this controlled environment show that over the observed depth range, the RL training steps $T$ required to reach $90\%$ validation accuracy follow a power law in the proof-tree depth $D$ ($T \propto D^{\gamma}$, $R^2 > 0.99$), where the scaling exponent $\gamma$ increases monotonically with expressiveness from $1.05$ to $2.60$. This indicates that more logically expressive settings require disproportionately more training to solve problems of the same depth. We also find that the power-law relationship holds across different RL methods and models, and these dynamics depend strongly on the training distribution: a carefully designed curriculum improves training efficiency and stability.

We also demonstrate that training a model in our synthetic reasoning environment improves performance on downstream real-world mathematics and general reasoning benchmarks, with the most expressive setting improving mean accuracy across eight benchmarks by up to $+10.66$ percentage points over the base model. Critically, training on highly expressive data yields a monotonically increasing downstream performance curve, while less expressive settings plateau early. This indicates that \emph{what} the model is trained on, not just \emph{how much} it is trained, shapes downstream transfer. Taken together, these results suggest that the expressiveness of the training data
influences both RL scaling on synthetic tasks and transferable gains in reasoning performance for real-world applications. 
More broadly, LLM shortcomings in long-horizon reasoning could be surmountable through improved training methodology and data.

In summary, we introduce a scalable synthetic reasoning framework for studying and improving long-horizon reasoning in RL post-training. Our main contributions are as follows:

\begin{itemize}
[leftmargin=1.5em,itemsep=1pt,parsep=0pt,topsep=2pt]
    \item We propose \textsc{ScaleLogic}, a controlled and scalable synthetic logical reasoning testbed for RL post-training, with exact verifiability, low-cost automatic generation, and explicit control over reasoning horizon and expressiveness.

    \item We show that RL training effort follows a power law in reasoning depth, with the scaling exponent $\gamma$ increasing monotonically with expressiveness. The power-law relationship also holds across multiple RL methods and models, while curriculum training improves scaling efficiency.


    \item We show that synthetic reasoning training improves downstream reasoning performance by up to $+10.66$ points, and these gains depend strongly on the expressiveness of the training environment, with more expressive settings yielding larger and more compute-efficient transfer.
\end{itemize}

\section{Related Work}

\paragraph{Long-horizon Reasoning Limitations.}
Recent work reveals that LLMs often exhibit sharp performance degradation as the reasoning horizon increases.
\citet{rameshkumar2025reasoning} show that reasoning models can perform well on graph-based reasoning tasks within a limited complexity regime, but their performance drops abruptly once the reasoning horizon is sufficiently large.
SeqBench~\citep{ramezanali2025seqbench} and GSM-Infinite~\citep{zhou2025gsminfinite} also report steep performance degradation with exponential and sigmoid-like decay patterns.
\citet{saparov2025transformers,DBLP:conf/icml/BachmannN24} found that teacher forcing alone is unable to teach long-horizon reasoning to transformers, suggesting these limitations may be more fundamental, perhaps architectural.
Further studies, such as R-Horizon~\citep{lu2025rhorizon} and h1~\citep{motwani2025h1}, compose individually solvable mathematical problems into multi-step dependency chains and investigate whether RL training can mitigate long-horizon failures.
Building on this line of work, we characterize how RL training behaves as reasoning structures are systematically scaled, and how these scaling dynamics shape downstream transfer. 

\paragraph{Scaling in LLMs.}

Early scaling-law studies found regular power laws relating pre-training performance to model scale, data volume, and training compute~\citep{kaplan2020scaling,henighan2020autoregressive,hoffmann2022chinchilla}. 
Later, test-time scaling emerged to improve reasoning by allocating additional computation during decoding~\citep{wei2022cot,wang2023selfconsistency,yao2023tree,muennighoff2025s1}.
Recent work has extended scaling-law analysis to RL post-training, showing regular scaling behavior with model scale, data, and compute~\citep{khatri2025scalerl,tan2025scaling}.
However, existing RL scaling studies primarily vary the volume of training data, while providing limited control over the reasoning complexity of individual problems.
In contrast, our work enables a cleaner analysis through explicit and interpretable control of reasoning complexity.

\paragraph{RL for LLM Reasoning.}

RL with verifiable rewards (RLVR) has become a promising paradigm for reasoning-oriented post-training~\citep{luong2024reft,guo2025deepseek}.
With practical optimizers such as GRPO~\citep{shao2024deepseekmath} and later variants~\citep{yu2025dapo,zheng2025gspo}, it has enabled large-scale RL training and long chain-of-thought reasoning~\citep{jaech2024openai,guo2025deepseek}.
However, most existing RL reasoning work centers on mathematics and programming, where high-quality training problems are limited, often depend on human-curated solutions or test cases, and offer only coarse difficulty control~\citep{liu2025saturn}.
As a result, RL performance is limited by the collected data, making sustained improvement difficult as training scales.
In contrast, \textsc{ScaleLogic} offers explicit complexity control, verifiable solutions, and unlimited low-cost data generation, enabling a cleaner and more scalable framework for reasoning-oriented RL.

\paragraph{Synthetic Data for Post-training.}


Synthetic data is a natural fit for RLVR, motivating recent work on automatically generated reasoning problems with verifiable rewards.
Early efforts studied task-specific synthetic settings for logical reasoning, such as Knights and Knaves~\citep{guo2025logicrl,lin2025zebralogic}.
More recent work shifted toward task families with difficulty control such as SAT~\citep{liu2025saturn}, graph reasoning tasks such as G1~\citep{guo2025g1}, and benchmark-style synthetic reasoning suites~\citep{liu2025synlogic, chen2025enigmata, stojanovski2025reasoninggym, he2026resyn}.
While these studies show the promise of synthetic data for post-training, their tasks are typically worst-case NP-hard search problems (e.g., SAT and Hamiltonian path in G1) with limited control over underlying expressiveness, making it difficult to isolate how task complexity affects RL training compute and downstream reasoning.
In contrast, \textsc{ScaleLogic} independently controls proof depth and expressiveness with each instance oracle-solvable in time linear in the proof size.

\input{method}

\input{experiment}

\section{Conclusion}

We introduced \textsc{ScaleLogic}, a synthetic logical reasoning environment with independent control over reasoning depth and expressiveness, and used it to study how RL post-training scales with task complexity. Training compute $T$ follows a power law in proof depth $D$ ($T \propto D^{\gamma}$, $R^2 > 0.99$) across settings, with the scaling exponent $\gamma$ increasing monotonically from $1.05$ to $2.60$ as the logic grows richer. Crucially, expressiveness influences not only training dynamics but also downstream transfer: more expressive settings yield larger and more compute-efficient gains on real-world reasoning tasks, indicating that downstream transfer is shaped by \emph{what} a model is trained on, not merely by \emph{how much} it is trained. Our results suggest that improved training methodology and data could be used to overcome LLM limitations in long-horizon reasoning.

Several directions remain open for future work. First, while our study provides a controlled empirical characterization of RL scaling in long-horizon reasoning, it remains important to test whether these scaling trends persist for substantially larger models and broader RL training regimes. Second, our expressiveness hierarchy covers several core logical operators, but richer fragments---such as equality, higher-order reasoning, non-monotonic reasoning, and more realistic multi-entity relational structures---could reveal new scaling regimes beyond those observed here. Finally, the observed power-law exponents call for a more formal theoretical explanation: understanding why different logical operators change the scaling exponent may provide a principled account of how structural properties of training data shape RL efficiency. 
\medskip
\bibliographystyle{iclr2027_conference}
\bibliography{ref}

\newpage
\appendix

\begin{center}
    {\bf\Large Appendix}
\end{center}

{
\startcontents[sections]
\printcontents[sections]{l}{1}{\setcounter{tocdepth}{2}}
}

\clearpage

\input{appendix}

\end{document}

%% file: method.tex
\section{Method}

We present \textsc{ScaleLogic}, a framework for generating synthetic logical reasoning problems with fine-grained control over task difficulty.
Section~\ref{sec:generation} describes how reasoning problems are constructed, Section~\ref{sec:difficulty} explains how the difficulty of the generated problems is systematically controlled through expressiveness, and Section~\ref{sec:rl_grpo} describes the reinforcement learning setup for post-training.

\subsection{Generation of Logical Reasoning Problems}\label{sec:generation}
Our training environment is built on a pipeline for generating synthetic logical reasoning problems.
We refer to a grounded predicate applied to a specific entity, possibly negated, such as ``Alice is a cat'' (written in logic as \logic{cat(Alice)}) or ``Alice is not a cat'' (\logic{$\neg$cat(Alice)}), as a \emph{literal}.
Each instance consists of a collection of axioms, which include literals such as ``Alice is a cat'' (\logic{cat(Alice)}) and rules such as ``If Alice is a cat, then Alice is a mammal''
(\logic{cat(Alice)}~$\rightarrow$~\logic{mammal(Alice)}), which together determine the set of conclusions that can be logically derived, such as ``Alice is a mammal'' (\logic{mammal(Alice)}).
The underlying task is to identify, among a set of candidate conclusions (each a single literal), which one is logically derivable from the given axioms.

Each instance is presented as a single-answer multiple-choice problem. To construct it, we first sample $B$ literals, each serving as the root of a proof tree.
Starting from each root, we recursively expand the tree by adding child literals to its leaves.
Each parent node, along with its set of children, defines a proof step: the children serve as the premises, while the parent is the conclusion.
The conclusion of one proof step may in turn serve as a premise for another step higher in the tree.
Thus, expanding a leaf amounts to generating its supporting premises, which become new leaves of the proof tree.
We repeat this process until a target proof depth~$D$ is reached, at which point the leaves are treated as the literal axioms from which the proof begins (see Algorithm~\ref{alg:generate_instance}).
This procedure generates proofs ``backwards'': it starts from the conclusion and progressively constructs the premises needed to derive it.
To avoid introducing alternative derivations, each expansion adds premise literals with fresh predicates that do not appear elsewhere in any proof trees.
This ensures that every node has a unique derivation from the axioms, corresponding to the subtree rooted at that node.\footnote{Similar to the generation procedure in \citet{opedal2025efficientreasoners} which also guarantees the uniqueness of generated proofs.}
Applying the procedure to the $B$ roots yields one proof tree for each candidate conclusion.

We keep one proof tree intact so that its root conclusion remains derivable.
For each of the remaining $B-1$ candidates, we uniformly sample one axiom from the proof and corrupt it in one of two ways:
(i)~\emph{removing} the axiom, or (ii)~\emph{flipping} the polarity of one literal within the axiom, e.g., changing ``Alice is a cat'' (\logic{cat(Alice)}) to ``Alice is not a cat'' (\logic{$\neg$cat(Alice)}). 
Note that option (ii) is available only when negation is included in the underlying formal system; it is disabled in less expressive settings that do not support negation (see Appendix~\ref{app:expressiveness}).
The uniform sampling process prevents the model from exploiting positional shortcuts.
Since our backward construction yields a unique proof for each candidate, corrupting a single axiom severs the proof path to the root and makes the corresponding conclusion non-derivable.
We additionally insert a small number of distracting rules to increase local ambiguity without changing the derivability of any candidate conclusion; their generation is detailed in Appendix~\ref{app:stage2}. 

This construction yields two interpretable structural variables, $B$ and $D$, that directly govern instance complexity. Increasing $B$ introduces more plausible candidate conclusions that the model must distinguish, while increasing $D$ requires reasoning over longer proof chains. At the same time, the formulation allows exact verification of the final answer without supervising the entire proof, making it a natural fit for reinforcement learning with verifiable rewards.

After construction, we convert the axioms and candidate conclusions into natural language via predefined templates, instantiated with randomly sampled entity names and fake predicate words (see Appendix~\ref{app:stage3}). An illustrative example is provided below and algorithms are given in Appendix~\ref{app:algorithm}.

\begin{examplebox}[Natural Language Example]
Suppose we have the following facts:

Alice has fur.
Alice is a vertebrate.
If Alice has fur and is a vertebrate, then Alice is a mammal.
If anyone is a mammal, then they are not a bird.
...

Here are some candidate statements:
(A) Alice is not a bird.
(B) Alice is cold-blooded.
(C) ...

Which one of the above statements can be logically derived from the given facts?

\textbf{Answer:} Alice is not a bird.
\end{examplebox}

\subsection{Control of Expressiveness}\label{sec:difficulty}


Beyond depth and candidate count, we vary the expressiveness of the underlying formal system while preserving the same task format.
We construct a hierarchy of five settings by progressively expanding the set of available inference rules, with each added logical operator enabling a broader range of proof structures; further details are provided in Appendix~\ref{app:expressiveness}.

\paragraph{Implication-only.}
We start with a simple logic that contains only the implication operator (i.e., ``if-then'', written \logic{$\to$}). Each axiom is either a grounded literal, such as ``Alice is a cat''
(\logic{cat(Alice)}), or a simple implication rule such as ``If Alice is a cat, then Alice is a mammal'' (\logic{cat(Alice)~$\rightarrow$~mammal(Alice)}).\footnote{This logic is also known as \emph{implicational propositional logic} or \emph{implicational propositional calculus}.}
A statement is considered valid if it can be derived from the grounded literals by repeatedly applying rules whose antecedents are satisfied.

\paragraph{+ Conjunction.}
We extend the simplest logic with the conjunction operator (i.e., ``and'', written \logic{$\land$}), which allows a rule to depend on multiple premises simultaneously.\footnote{We only permit conjunctions in the antecedent of if-then statements, since \logic{A~$\rightarrow$~B~$\land$~C} is equivalent to \logic{A~$\rightarrow$~B} and \logic{A~$\rightarrow$~C}. Similarly, we only permit disjunctions in the consequent.} Under this logic, a rule may take the form of a conjunction of grounded literals entailing a single conclusion, such as ``If Alice is a vertebrate and has fur, then Alice is a mammal''
(\logic{vertebrate(Alice)~$\land$~has\_fur(Alice)~$\rightarrow$~mammal(Alice)}).
During inference, the model must coordinate multiple satisfied supporting literals before applying each rule, rather than relying on single-premise inference.

\paragraph{+ Negation.}
We further extend the logic with the negation operator (i.e., ``not'', written \logic{$\neg$}), which allows premises and conclusions to involve negated literals. Under this logic, rules may condition on negated properties or derive negated literals as a consequence of other literals, such as
``If Alice is a mammal, then Alice is not a bird''
(\logic{mammal(Alice)~$\rightarrow$~$\neg$bird(Alice)}).
With negation available, each literal in a proof tree has a well-defined negated counterpart, so the model must track not only whether a predicate has been established but also its polarity.

\paragraph{+ Disjunction.}
We further extend the logic with the disjunction operator (i.e.,
``or'', written \logic{$\lor$}), which allows a rule to produce multiple possible consequents from the same premises. Under this logic, a rule may take the form of the antecedent entailing a disjunction of grounded literals, such as ``If Alice is a pet, then Alice is a cat or a dog''
(\logic{pet(Alice)~$\rightarrow$~cat(Alice)~$\lor$~dog(Alice)}).
With disjunction, the model must reason over multiple possible conclusions, determining which alternatives are eliminated and which support the target statement.

\paragraph{+ Quantification.}
The most expressive logic we consider extends the earlier logic from purely propositional reasoning towards first-order reasoning, through \emph{universal quantification} (i.e., ``for all'', written \logic{$\forall$}), which enables the definition of rules that apply to \emph{any} entity rather than a specific one. For example, ``Anyone who is a cat is a mammal''
(\logic{$\forall$X$($cat(X)~$\rightarrow$~mammal(X)$)$}).
At inference time, applying such quantified rules requires the model to instantiate them with a concrete entity in the current context, and to verify that the instantiated antecedents hold before deriving the consequents.


\subsection{Reinforcement Learning Framework}\label{sec:rl_grpo}

We use DAPO~\citep{yu2025dapo}, an extension of GRPO~\citep{shao2024deepseekmath}, as our default RL algorithm. For each prompt $q$, we sample $G$ completions $\{o_i\}_{i=1}^G$ and optimize the policy objective

\vspace{-0.5em}
\begin{equation*}
\mathcal{L}_{\mathrm{DAPO}}(\theta)
=
\mathbb{E}_{q,\{o_i\}_{i=1}^G \sim \pi_{\theta_{\mathrm{old}}}}
\left[
\frac{1}{\sum_{i=1}^G|o_i|}
\sum_{i=1}^G
\sum_{t=1}^{|o_i|}
\min
\left(
r_{i,t}(\theta)\,\hat{A}_i,\;
\mathrm{clip}\!\left(r_{i,t}(\theta),\,1-\epsilon_l,\,1+\epsilon_h\right)\hat{A}_i
\right)
\right],
\end{equation*}
\begin{equation*}
\text{where}\hspace{3em} r_{i,t}(\theta)=\frac{\pi_{\theta}(o_{i,t}\mid q,o_{i,<t})}{\pi_{\theta_{\mathrm{old}}}(o_{i,t}\mid q,o_{i,<t})},
\qquad
\hat{A}_i=\frac{R_i-\mathrm{mean}(\{R_j\}_{j=1}^G)}{\mathrm{std}(\{R_j\}_{j=1}^G)}.
\end{equation*}
\vspace{-0.5em}

Here, $r_{i,t}(\theta)$ is the token-level policy ratio, and $\hat{A}_i$ is the group-normalized advantage over completion rewards $\{R_i\}_{i=1}^G$. Following DAPO, we use dynamic sampling and asymmetric clipping with $(\epsilon_l,\epsilon_h)=(0.20,0.28)$.
For reward design, we adopt a simple and verifiable binary reward, following common practice in reasoning-oriented RL~\citep{guo2025deepseek}. Specifically, we require the model to place its final answer within \texttt{<answer>...</answer>}. During evaluation, the verifier extracts the predicted answer from this span and compares it with the ground-truth answer via exact match. If the output violates the required format or the extracted answer does not match the ground truth, we set $R_i = 0$; otherwise, $R_i = 1$.
Full prompt templates are provided in Appendix~\ref{app:template}.

%% file: experiment.tex
\begin{figure}[t]
    \centering
    \begin{subfigure}[b]{0.49\textwidth}
        \centering
        \includegraphics[width=\textwidth]{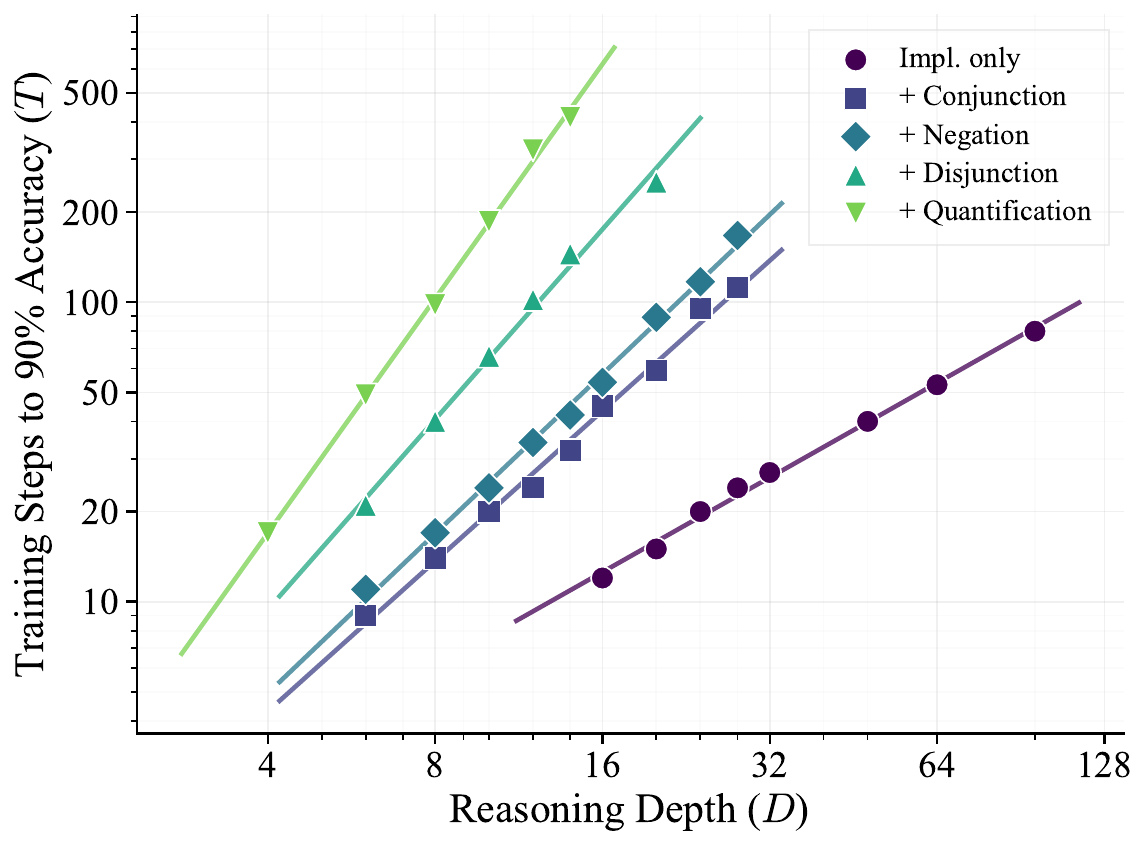}
        \vspace{-1em}
        \caption{Log-log scaling curves with power-law fits.}
        \vspace{-0.7em}
        \label{fig:scaling-loglog}
    \end{subfigure}
    \hfill
    \begin{subfigure}[b]{0.49\textwidth}
        \centering
        \includegraphics[width=\textwidth]{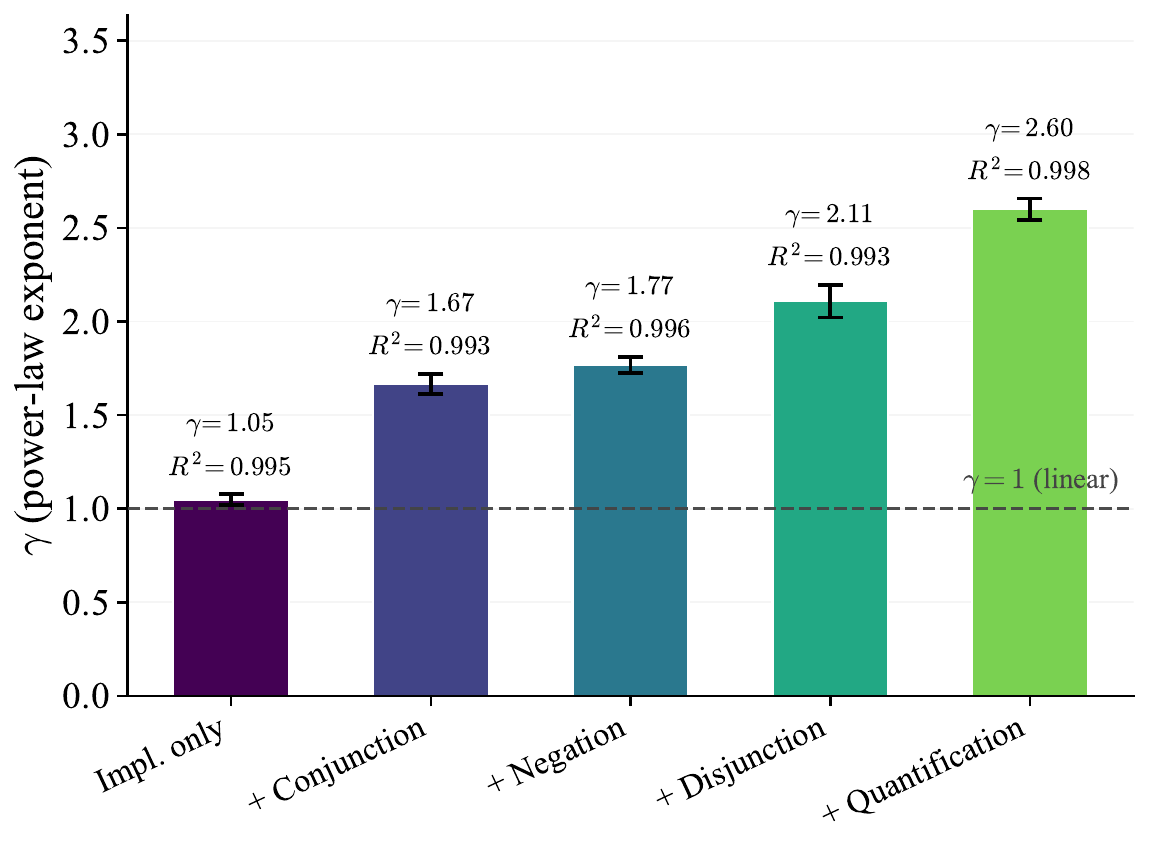}
        \vspace{-1em}
        \caption{Fitted exponent $\gamma$ vs. Expressiveness of the logic.}
        \vspace{-0.7em}
        \label{fig:gamma-bar}
    \end{subfigure}
    \caption{\textbf{Training cost exhibits power-law scaling with reasoning depth, with exponent $\gamma$ increasing systematically along the expressiveness hierarchy.} \textbf{(a)} Scatter points show the training steps $T$ required to reach the 90\% accuracy threshold as a function of reasoning depth $D$. Solid lines show power-law fits $T \propto D^{\gamma}$. \textbf{(b)} Fitted exponent $\gamma$ increases monotonically when expressiveness increases from \textit{Implication-only} ($\gamma\!=\!1.05$) to \textit{+~Quantification} ($\gamma\!=\!2.60$). Error bars denote $\pm 1$ standard error of the fitted exponent.}
    \label{fig:scaling-power-law}
    \vspace{-1em}
\end{figure}

\section{Experiments}

In this section, we aim to answer the following research questions.

\begin{itemize}[leftmargin=2.5em, labelsep=0.5em, itemsep=1pt, parsep=1pt, topsep=-4pt]
    \item[\textbf{RQ1}] \textbf{Scaling with complexity} (\S\ref{sec:scaling-depth}). How does the cost of training a model with RL to reach a target accuracy scale with reasoning depth and expressiveness?
    \item[\textbf{RQ2}] \textbf{Downstream transfer} (\S\ref{sec:downstream}). Does training on synthetic reasoning tasks improve performance on real-world benchmarks, and how does expressiveness affect transfer?
    \item[\textbf{RQ3}] \textbf{Training distribution} (\S\ref{sec:distribution}). How does the training distribution affect scaling efficiency?
    \item[\textbf{RQ4}] \textbf{Generalization across RL algorithms} (\S\ref{sec:cross-algorithm}). Is the observed scaling behavior algorithm-specific, or does it reflect a broader phenomenon across RL methods?
    \item[\textbf{RQ5}] \textbf{OOD generalization} (\S\ref{sec:ood}). Does training generalize to more difficult (unseen) depths?
\end{itemize}

\subsection{Experimental Setup}\label{sec:setup}

\paragraph{Models and implementation.}
We perform RL post-training on the non-thinking Qwen3-4B~\citep{yang2025qwen3} using the \texttt{verl} library~\citep{sheng2024hybridflow}. To assess generality across model scales and families, we replicate a subset of experiments on Qwen3-8B and Ministral3-3B~\citep{liu2026ministral} in Appendix~\ref{app:qwen3-8b}. Full implementation details and hyperparameters are provided in Appendix~\ref{app:implementation}.

\paragraph{Task configurations.}
We vary task difficulty along two axes: \emph{reasoning depth} and \emph{expressiveness}.
Unless otherwise specified, we fix the number of candidates to $B=4$ throughout, and study the effect of varying $B$ separately in Appendix~\ref{app:branch-scaling}.
For each target depth $D$, we construct a training set by uniformly sampling instances from depths $\{1,\dots,D\}$, choosing a dataset size (typically 100,000) sufficient for convergence within a single epoch, with each instance seen only once.
Evaluation is performed on a held-out validation set of 1{,}000 instances from the same configuration.
We study five levels of expressiveness, ranging from implication-only reasoning towards increasingly expressive first-order reasoning, as summarized in Table~\ref{tab:expressiveness-mapping}.
To confirm task difficulty, we evaluate six frontier LLMs on \textsc{ScaleLogic}; all models degrade substantially at large depths (see Appendix~\ref{app:strong_llm}).

\paragraph{Training protocol and compute metric.}
For each configuration, we train a separate model and evaluate it on the corresponding held-out validation set after every RL training step. Our primary measure of training compute, denoted by $T$, is the number of RL training steps required to reach a held-out \texttt{Pass@1} threshold of $\mu = 90\%$.
We use this metric throughout the main text and discuss additional compute-related quantities, including generated tokens and FLOPs, in Appendix~\ref{app:training_effort}.

\paragraph{Downstream benchmarks.}
To evaluate transfer beyond the synthetic environment, we test the trained models on a diverse set of downstream reasoning benchmarks, including AIME 2024 and 2025, AMC 2023, MATH-500 \citep{hendrycks2021measuring}, Minerva \citep{lewkowycz2022solving}, OlympiadBench text-only subset \citep{he2024olympiadbench}, GPQA-Diamond \citep{rein2023gpqa}, and a STEM subset of MMLU-Pro \citep{wang2024mmlu}.
Our main downstream metric is the mean \texttt{Avg@8} across the eight benchmarks. 
Detailed results for each benchmark are reported in Appendix~\ref{app:downstream}.

\begin{figure}[t]
  \centering
  \begin{subfigure}[b]{0.49\textwidth}
    \centering
    \includegraphics[width=\textwidth]{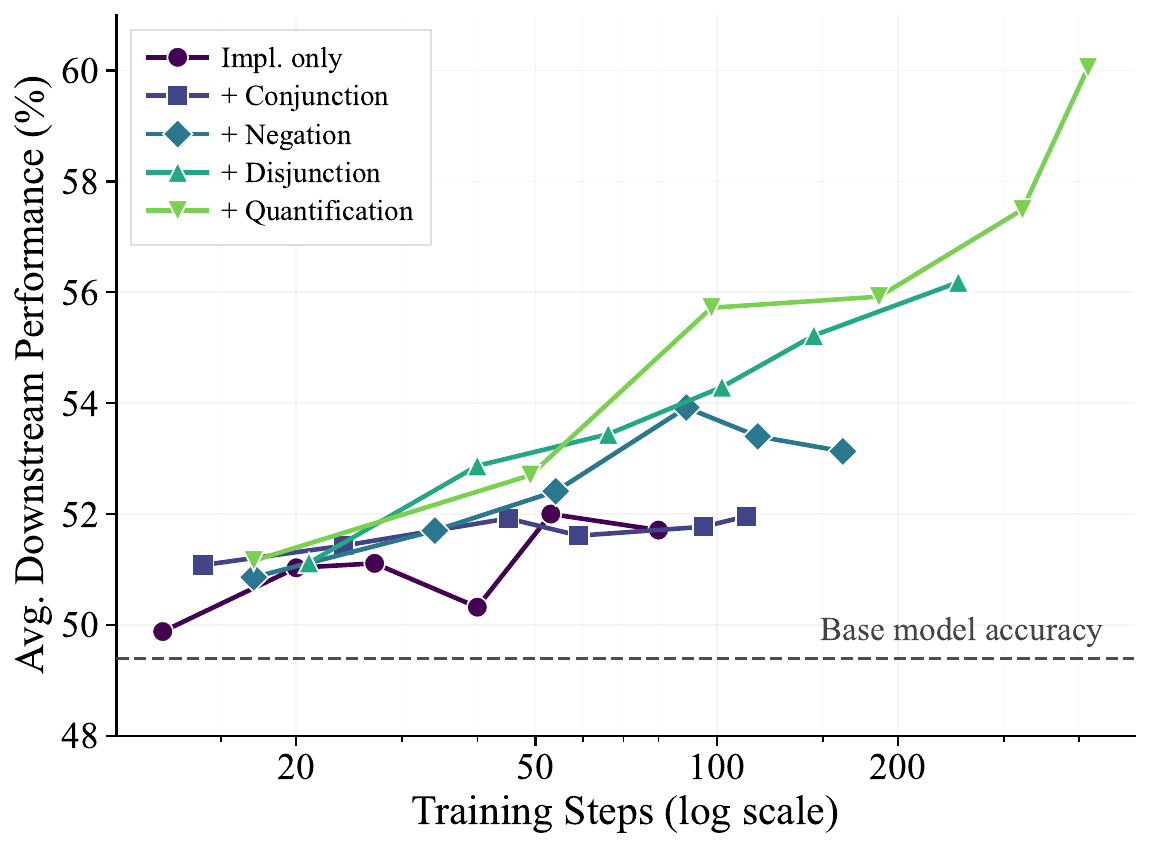}
    \vspace{-1em}
    \caption{Downstream performance vs.\ Training steps.}
    \vspace{-0.7em}
    \label{fig:downstream_a}
  \end{subfigure}
  \hfill
  \begin{subfigure}[b]{0.49\textwidth}
    \centering
    \includegraphics[width=\textwidth]{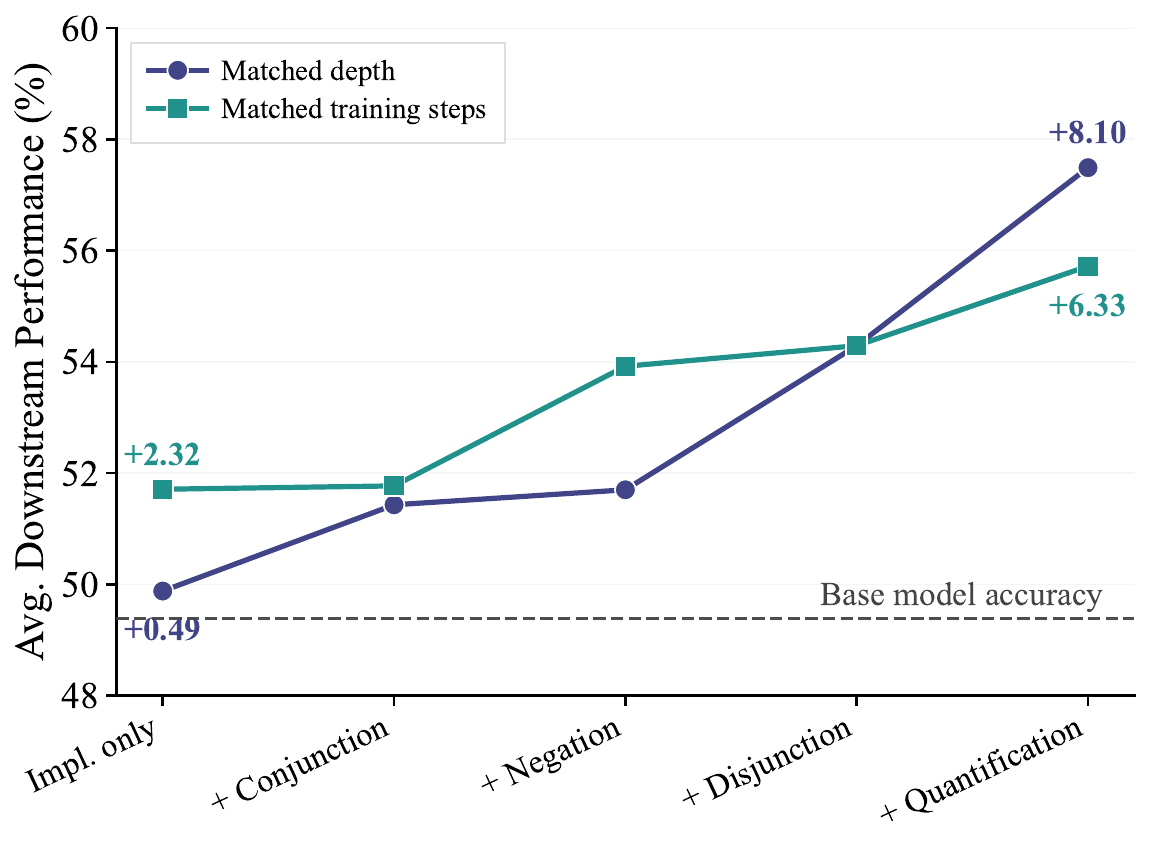}
    \vspace{-1em}
    \caption{Controlled comparisons.}
    \vspace{-0.7em}
    \label{fig:downstream_b}
  \end{subfigure}
  \caption{\textbf{Downstream transfer from synthetic reasoning training.}
  \textbf{(a)} All settings outperform the base model, with richer logical settings producing larger and more sustained gains. \textbf{(b)} Average downstream performance across logical settings under \emph{matched depth} ($D=12$) and \emph{matched training steps} ($\sim 100$ steps). Both controls exhibit the same monotone trend: More expressive training settings yield stronger downstream performance.}
  \label{fig:downstream}
  \vspace{-1em}
\end{figure}

\subsection{Power-Law Scaling with Depth and Expressiveness}\label{sec:scaling-depth}

We first study how training effort scales with reasoning depth across different levels of expressiveness.
As shown in Figure~\ref{fig:scaling-loglog}, over the observed depth range, training compute $T$ follows a power law with respect to the reasoning depth $D$, namely $T = a \cdot D^{\gamma}$, across all five expressiveness levels, with $R^2 > 0.99$ in every case.
We fit this relationship via ordinary least squares in log-log space. The power-law model consistently outperforms an exponential fit across all settings ($\Delta\text{AIC} \geq +13.9$; see Appendix~\ref{app:fit_comparison}),
suggesting that training cost grows polynomially rather than exponentially with reasoning depth in the studied regime.

While all settings follow a power-law form, the fitted exponent $\gamma$ increases monotonically with expressiveness (Figure~\ref{fig:gamma-bar}), from $1.05$ for \textit{Implication-only} to $2.60$ for \textit{+~Quantification}.
Thus, doubling reasoning depth increases training cost by roughly $2\times$ in the simplest setting, but about $6\times$ in the most expressive setting.
The near-linear exponent for \textit{Implication-only} suggests that, when the logical structure is sufficiently simple, each additional unit of depth imposes an approximately constant marginal learning cost: the model primarily learns to ``chain one more step.''
By contrast, richer logical operators introduce additional combinatorial structure, such as jointly verifying multiple premises under conjunction, and exhibit more rapidly compounding depth-related costs.
Notably, \textit{+~Conjunction} ($\gamma = 1.67 \pm 0.05$) and \textit{+~Negation} ($\gamma = 1.77 \pm 0.04$) exhibit the smallest increase in scaling exponent between adjacent levels ($\Delta\gamma = 0.10$). 
This is consistent with negation primarily requiring the model to track literal polarity rather than introducing additional combinatorial structure, particularly as our setting excludes proof by contradiction.
Appendix~\ref{app:Scaling_robustness} further shows that the monotone expressiveness ordering is robust to alternative analysis choices and potential confounders in task construction.

\begin{figure}[t]
    \centering
    \begin{subfigure}[b]{0.49\linewidth}
        \centering
        \includegraphics[width=\linewidth]{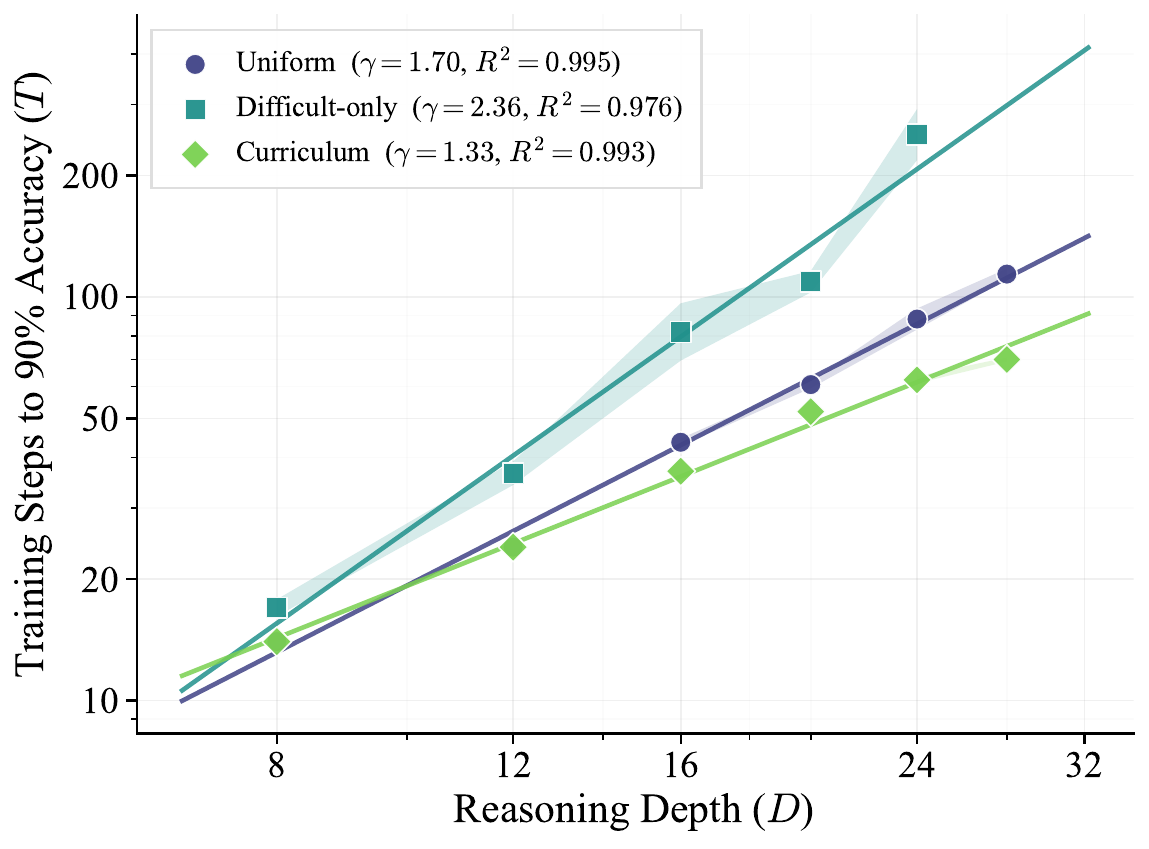}
        \vspace{-1em}
        \caption{Training distribution comparison.}
        \vspace{-0.7em}
        \label{fig:distribution_efficiency}
    \end{subfigure}
    \hfill
    \begin{subfigure}[b]{0.49\linewidth}
        \centering
        \includegraphics[width=\linewidth]{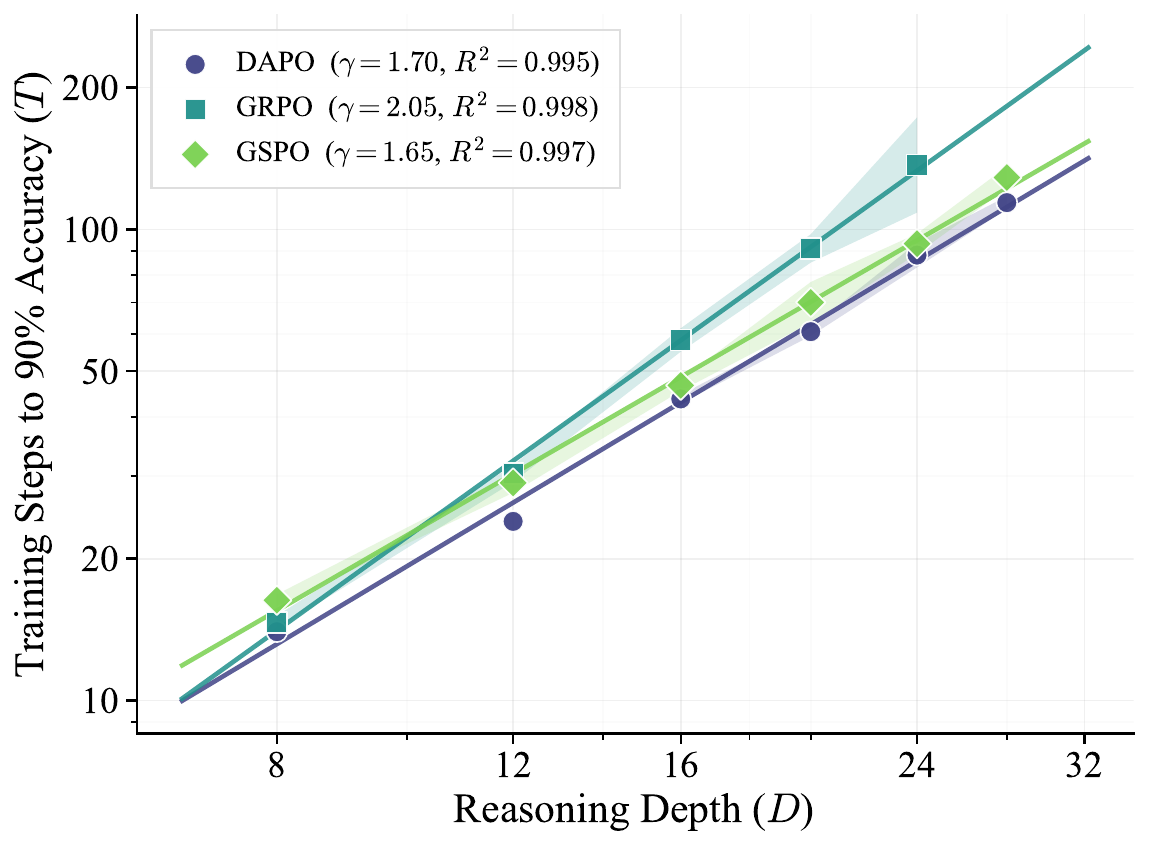}
        \vspace{-1em}
        \caption{Cross-algorithm comparison.}
        \vspace{-0.7em}
        \label{fig:algorithm}
    \end{subfigure}
    \caption{
\textbf{Effect of training distribution and RL algorithm on scaling efficiency in the \emph{+~Conjunction} setting.} Each curve aggregates three independent seeds;
shading denotes $\pm 1$ standard deviation in log-space across seeds.
\textbf{(a)} Curriculum training yields the lowest exponent ($\gamma=1.33$), whereas difficult-only training yields the highest ($\gamma=2.36$).
\textbf{(b)} All RL algorithms follow power-law scaling ($R^2>0.99$), with exponents from $1.65$ to $2.05$.
}
    \label{fig:distribution_algorithm}
    \vspace{-1em}
\end{figure}

\subsection{Downstream Transfer from Synthetic Reasoning Tasks}
\label{sec:downstream}

We next study whether learning on synthetic reasoning tasks transfers to real-world downstream benchmarks.
Figure~\ref{fig:downstream_a} reports average downstream performance as a function of RL training steps across all five expressiveness levels.
All settings improve over the base model (49.39\%), confirming that the reasoning skills learned in our synthetic environment are not confined to the training domain. 
However, the size and persistence of these gains differ markedly depending on the expressiveness of the training examples. Simpler settings (\textit{Implication-only} and \textit{+~Conjunction}) plateau early around 52\%, whereas the \textit{+~Quantification} setting continues to improve throughout training, reaching 60.05\% at 414 steps---an absolute gain of 10.66 percentage points.

To disentangle the contributions of expressiveness and training compute, Figure~\ref{fig:downstream_b} compares two controlled conditions across the five settings.
The \emph{matched depth} line holds the target reasoning depth at $D = 12$, isolating the effect of expressiveness at a comparable task horizon.\footnote{For the \textit{Implication-only} setting, we use $D=16$ because smaller depths are close to saturation (see Appendix~\ref{app:implementation}).}
The \emph{matched training steps} line selects the checkpoint closest to $100$ training steps for each setting, controlling for training budget.
Under matched depth, downstream gains increase from $+0.49$ points for \textit{Implication-only} to $+8.10$ points for \textit{+~Quantification}, indicating a strong association between expressiveness and downstream transfer.
Under matched training steps, gains follow the same monotone trend, rising from $+2.32$ to $+6.33$ points.
Together, these controls suggest that downstream reasoning transfer depends strongly on \emph{what} the model is trained on, not merely on \emph{how much} training it receives.
Additional analyses in Appendix~\ref{app:additional-result} show that neither SFT nor correctness-independent RL reproduces these downstream gains, while external logical-reasoning evaluations and comparison with an alternative synthetic framework further support the effectiveness and generality of \textsc{ScaleLogic}.

\subsection{Effect of Training Distribution on Scaling Efficiency}\label{sec:distribution}

We next study how the distribution of training difficulty affects the efficiency and stability of RL scaling. Under the \textit{+~Conjunction} setting, we compare three strategies for a target depth $D$: \textbf{uniform} sampling over depths $\{1, \dots, D\}$ as our default setting, \textbf{curriculum} sampling that gradually increases the depth during training, and \textbf{difficult-only} training that uses only depth-$D$ instances. We run each setting with three seeds per depth. The implementation details are provided in Appendix~\ref{app:curriculum}.

As shown in Figure~\ref{fig:distribution_efficiency}, the training distribution has a major effect on scaling efficiency. All three strategies maintain power-law scaling, but the fitted exponents differ substantially: curriculum training achieves the lowest exponent ($\gamma = 1.33$), followed by uniform training ($\gamma = 1.70$), while difficult-only training scales most steeply ($\gamma = 2.36$) and exhibits the largest variance.
Appendix~\ref{app:curriculum_quant} confirms the same benefit under \emph{+~Quantification}, lowering $\gamma$ from $2.60$ with uniform training to $2.30$ with curriculum training.
To better understand this effect, we analyze training dynamics in Appendix~\ref{app:curriculum_res}. Under curriculum training, long chain-of-thought behavior emerges at broadly similar stages across depths; under uniform or difficult-only training, however, its emergence is increasingly delayed and often irregular (e.g., note the large gap between depths 20 and 24).
These patterns suggest that curriculum training improves scaling efficiency by accelerating long-CoT emergence across depths, possibly because shallower instances provide a smoother bootstrap for learning useful reasoning patterns. We leave a thorough investigation of this mechanism to future work.

\begin{figure}[t]
    \centering
    \begin{subfigure}[t]{0.49\linewidth}
        \centering
        \includegraphics[width=\linewidth]{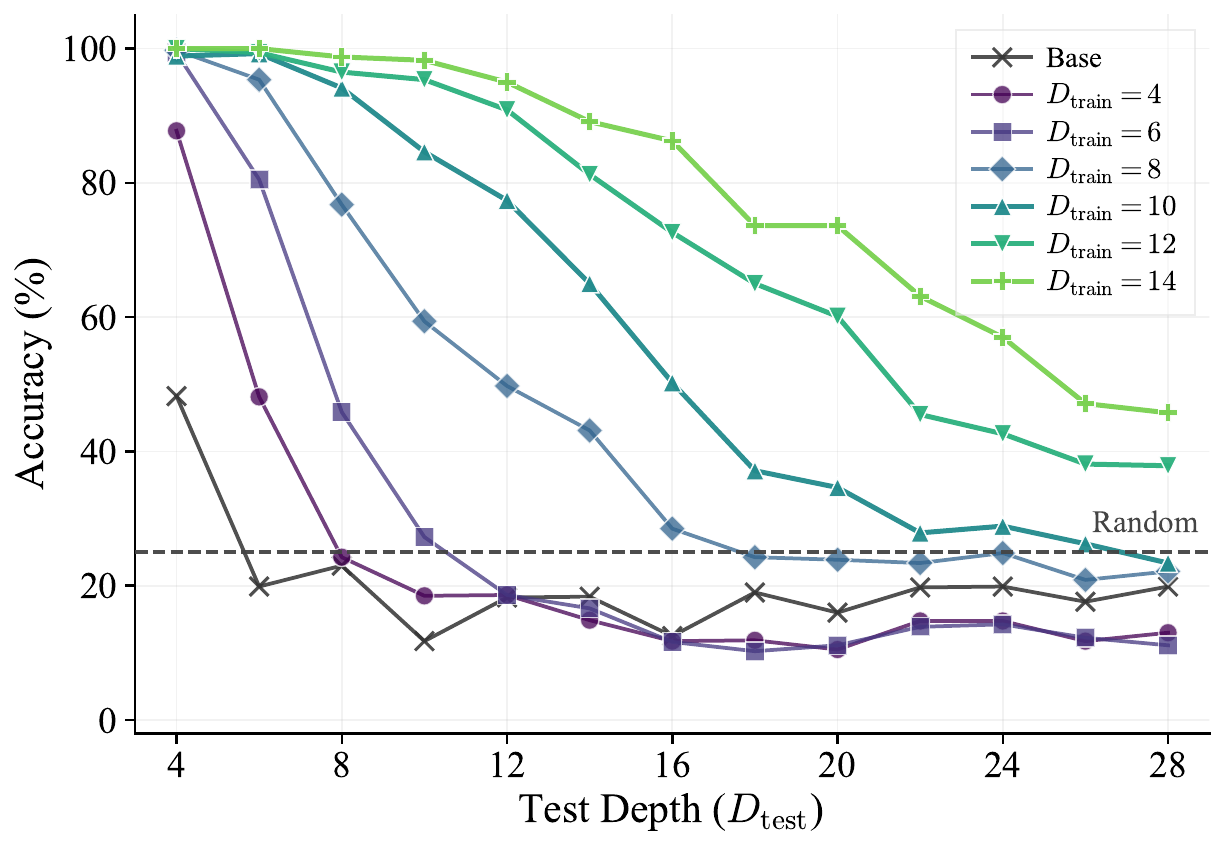}
        \vspace{-1em}
        \caption{Accuracy vs. Absolute test depth.}
        \vspace{-0.7em}
        \label{fig:ood_abs}
    \end{subfigure}
    \hfill
    \begin{subfigure}[t]{0.49\linewidth}
        \centering
        \includegraphics[width=\linewidth]{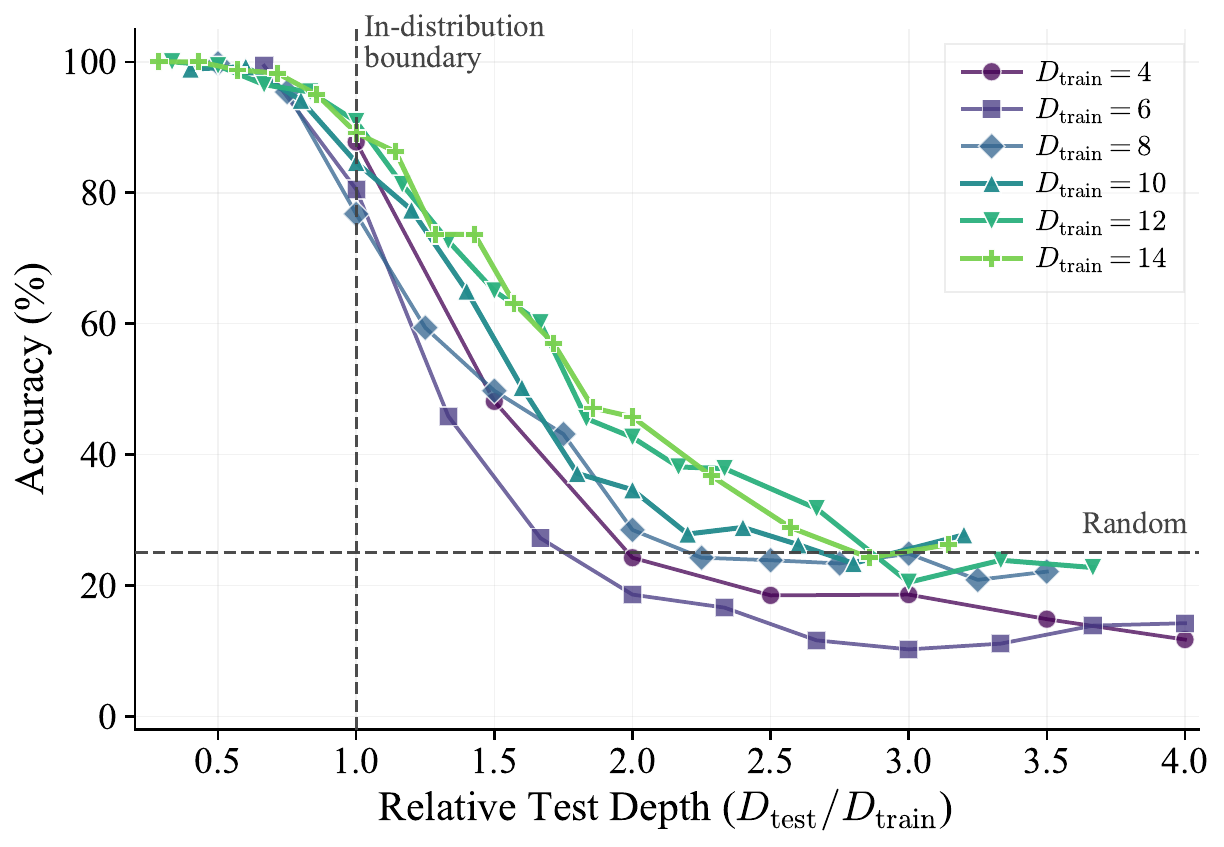}
        \vspace{-1em}
        \caption{Accuracy vs. Normalized test depth by train depth.}
        \vspace{-0.7em}
        \label{fig:ood_collapse}
    \end{subfigure}
    \caption{
    \textbf{Out-of-distribution generalization across reasoning depths.}
    \textbf{(a)} Increasing the training depth consistently extends the range of solvable test depths. \textbf{(b)} OOD generalization remains bounded: even the models trained at the largest depths fall to random at $D_{\mathrm{test}}/D_{\mathrm{train}}\!\approx\!3$.
    }
    \label{fig:ood}
    \vspace{-1em}
\end{figure}

\subsection{Cross-Algorithm Robustness}
\label{sec:cross-algorithm}

To assess whether the observed scaling behavior is specific to DAPO or reflects a broader property of RL training on these tasks, we repeat the scaling experiment in the \emph{+~Conjunction} setting with two additional algorithms: base GRPO~\citep{shao2024deepseekmath} (without DAPO extension) and GSPO~\citep{zheng2025gspo}, a recent sequence-level policy optimization variant.
We run each algorithm with three seeds per depth.
As shown in Figure~\ref{fig:algorithm}, all three algorithms exhibit clear power-law scaling ($R^2 > 0.99$), confirming that the relationship $T \propto D^{\gamma}$ is not an artifact of a particular optimization method. DAPO and GSPO have similar fitted exponents ($\gamma = 1.70$ and $1.65$, respectively), while GRPO exhibits a steeper scaling exponent ($\gamma=2.05$) and greater variance at larger depths, suggesting lower sample efficiency in the long-horizon regime.

\subsection{Out-of-Distribution Generalization}\label{sec:ood}

We examine whether training at a target reasoning depth generalizes to higher unseen depths by evaluating models trained with different depths under \textit{+~Quantification}.
Figure~\ref{fig:ood_abs} shows that larger training depths consistently extend the range over which models maintain strong performance: shallow-depth models degrade rapidly, while deeper-trained models preserve higher accuracy on harder evaluations.
This suggests that training expands the model's effective reasoning horizon rather than merely improving performance at the target depth.
However, this generalization remains bounded. Even the models trained at the largest depths eventually degrade at sufficiently large evaluation depths.
After normalizing test depth by training depth (Figure~\ref{fig:ood_collapse}), the two deepest models ($D_{\text{train}}\!=\!12,14$) collapse onto a frontier curve, with performance approaching random at roughly $3D_{\text{train}}$. Thus, deeper training extends the solvable range approximately linearly, but does not eliminate the horizon limit.


%% file: appendix.tex
\section{Limitations}
\label{sec:limitations}
The main experiments are conducted on Qwen3-4B, with a subset replicated on Qwen3-8B and Ministral3-3B (Appendix~\ref{app:qwen3-8b}) to assess cross-scale and cross-model consistency. However, it remains unclear whether the observed scaling laws generalize to substantially larger models or different architectures like MoE.
The power-law relationship $T = a \cdot D^{\gamma}$ is fitted using limited depth values for each expressiveness level. Due to computational constraints, we cannot further extend the depth range. Although the fits achieve $R^2 > 0.99$ and consistently outperform exponential alternatives (Table~\ref{tab:fit-comparison}), this limited range means that the power-law characterization should be interpreted as a high-fidelity empirical description of the observed regime, rather than a robust asymptotic law.
In addition, the main scaling curves are fit from single-seed runs; our multi-seed analysis at \emph{+~Conjunction} shows similar fitted exponents (Appendix~\ref{app:multi_seed}), though broader multi-seed validation is still needed.
Both the training-distribution comparison and the cross-algorithm analysis are conducted at selected expressiveness levels; whether these conclusions generalize across all expressiveness levels is not verified.
Finally, our analysis is empirical. While we observe a clean power-law dependence and a monotone relationship between expressiveness and the scaling exponent $\gamma$, we provide no theoretical derivation of these relationships. Establishing the theoretical foundation is left to future work.


\begin{algorithm}[t]
\caption{Backward Construction of a Proof Tree}
\label{alg:generate_instance}
\KwIn{target depth $D$, rule budget $E$, expressiveness flags $\Phi \subseteq \{\wedge, \neg, \vee, \forall\}$}
\KwOut{axiom-goal pair $(\mathcal{S}, goal)$}
$goal \leftarrow$ sample a fresh literal \tcp*{root conclusion of the proof tree}
$Q \leftarrow [(0, \; goal)]$ \tcp*{initial depth and root literal}
$\mathcal{S} \leftarrow [\,]$\;
$\mathcal{T} \leftarrow [\,]$ \tcp*{reusable quantified-rule templates}
\While{$Q \neq \emptyset$}{
    $(d, u) \leftarrow \textsc{Pop}(Q)$
    \tcp*{pop max-depth leaf}
    \If{$d \ge D$ \textbf{or} $\mathrm{rules}(S) \ge E$}{
        $\mathcal{S} \leftarrow \mathcal{S} \cup \{u\}$ \tcp*{add a literal axiom}
        \textbf{continue}\;
    }
    $\textit{scope} \leftarrow \textsc{ChooseEntityScope}(\forall \in \Phi)$ \tcp*{entity-specific or universal}
    \If{$\textit{scope} = \forall$ \textbf{and} $\mathcal{T} \neq \emptyset$ \textbf{and} $\textsc{Bernoulli}(p_\text{reuse})$}{
        $r \leftarrow$ instantiate a template from $\mathcal{T}$ with a fresh entity\;
        Bind one conclusion of $r$ to $u$, update $\mathcal{S}$; enqueue other literals of $r$ into $Q$\;
        Remove the template from $\mathcal{T}$ if it has been instantiated for all entities\;
        \textbf{continue}\;
    }
    $(n_p, n_c) \leftarrow \textsc{SampleArity}(\Phi)$
    \tcp*{$n_p \ge 2$ needs $\wedge \in \Phi$,\ $n_c \ge 2$ needs $\vee \in \Phi$}
    $P \leftarrow \textsc{BuildPremises}(d+1, n_p)$\;
    $C \leftarrow \textsc{BuildConclusions}(d, n_c, u)$
    \tcp*{$u$ should be one of the conclusions}
    enqueue newly introduced literals from $P$ and $C$ into $Q$\;
    $\mathcal{S} \leftarrow \mathcal{S} \cup \{P \to C\}$
    \tcp*{adds a rule axiom}
    \If{$\textit{scope} = \forall$}{
        $\mathcal{T} \leftarrow \mathcal{T} \cup \{P \to C\}$\;
    }
}
\If{$\neg \in \Phi$}{
    \ForEach{literal $u$ in $\mathcal{S}$}{
        \If{\textsc{Bernoulli}($p_{\neg}$)}{
            negate all occurrences of $u$ in $\mathcal{S}$
            \tcp*{random polarity flip for $\neg$}
        }
    }
}
\Return{$(\mathcal{S}, goal)$}\;
\end{algorithm}

\begin{algorithm}[t]
\caption{Multiple-Choice Instance Assembly}
\label{alg:merge_branches}
\KwIn{$B$ axiom-goal pairs $\{(\mathcal{S}_i, goal_i)\}_{i=1}^{B}$, maximum depth $D_{\max}$, distractor budget $M$, expressiveness flags $\Phi$}
\KwOut{combined axiom set $\mathcal{S}$, candidate conclusions $\mathcal{C}$}
$\mathcal{S} \leftarrow \mathcal{S}_1$\; $\mathcal{C} \leftarrow \{goal_1\}$ \tcp*{goal~1 is fixed as provable}
\For{$i=2$ \KwTo $B$}{
    $a^\star \leftarrow$ a randomly selected axiom from $\mathcal{S}_i$\;
    \uIf{$\neg \notin \Phi$ \textbf{or} $\textsc{Bernoulli}(p_\text{remove})$}{
        remove $a^\star$ from $\mathcal{S}_i$\;
    }
    \Else{
        flip the polarity of one literal in $a^\star$
        \tcp*{available only when $\neg \in \Phi$}
    }
    $\mathcal{S} \leftarrow \mathcal{S} \cup \mathcal{S}_i$\;
    $\mathcal{C} \leftarrow \mathcal{C} \cup \{goal_i\}$\;
}
$m \leftarrow \textsc{Uniform}\{0,\ldots,M\}$\;
$k \leftarrow 0$\;
\While{$k < m$ \textbf{and} $\mathrm{rules}(S) < BD_\mathrm{max}$}{
    randomly choose to introduce fresh distractor literals for $P_d$, $C_d$, or both\;
    $\mathcal{S} \leftarrow \mathcal{S} \cup \{P_d \to C_d\}$;
    $k \leftarrow k+1$
    \tcp*{add distractor rules}
}
\Return{$(\mathcal{S}, \mathcal{C})$}\;
\end{algorithm}

\section{Task Generation Procedure}
\label{app:algorithm}

This appendix provides the full generation procedure for \textsc{ScaleLogic}. We decompose the pipeline into three stages: backward construction of candidate proof trees, assembly of a multiple-choice instance with exactly one provable candidate, and conversion from symbolic formulas to natural language prompts. Throughout the pipeline, the expressiveness flags $\Phi \subseteq \{\land, \neg, \lor, \forall\}$ determine which logical operators are available, corresponding to the settings described in Section~\ref{sec:difficulty}. The soundness of the resulting construction is audited in Appendix~\ref{app:validity}.

\subsection{Backward Construction of Proof Trees}\label{app:stage1}

Algorithm~\ref{alg:generate_instance} constructs a single candidate proof tree.
The generator initializes the work queue as $Q=[(0,goal)]$, where $goal$ is a randomly sampled root literal at depth $0$.
During construction, $Q$ stores the current open leaves of the partially generated proof tree, each paired with its depth.
In practice, $Q$ is processed in a depth-first order: \texttt{POP} always selects a leaf literal at the largest current depth.
Once a popped literal has depth $d \ge D$ or the rule budget $E$ is exhausted ($E \geq D$), it is added to the axiom set as a given literal and is no longer expanded.
This depth-first expansion guarantees that the proof tree is grown until it reaches the target proof depth $D$.

For each popped leaf literal $u$ with $d<D$, the generator assigns a rule whose conclusion contains $u$.
The expressiveness flags $\Phi$ restrict the admissible rule forms: when conjunction or disjunction is enabled, we allow rules with multiple premises or multiple conclusions, respectively, with the arity set to 2 in our experiments.
All newly introduced literals in the premise set $P$ and conclusion set $C$, except for the already fixed literal $u$, are assigned fresh predicates to avoid ambiguity, and are treated as new leaf literals added to $Q$.
For details on how disjunctive conclusions are resolved to preserve a unique derivation, see Appendix~\ref{app:expressiveness}.
When universal quantification is enabled, the generator may reuse an existing universal rule template from the template set $\mathcal{T}$ with probability
$p_{\mathrm{reuse}}=\frac{|\mathcal{T}|}{|\mathcal{T}|+1}$.
It instantiates the template with an entity for which the template has not yet been used, and binds one of the instantiated conclusions to the current leaf literal $u$.
All remaining literals introduced by the instantiated rule are then handled in the same way as in ordinary rule expansion.
In our experiments, each instance contains at most two entities, so a reused universal template is removed from $\mathcal{T}$ once it has been instantiated for all available entities.
After the proof tree is built, if negation is enabled, each literal is independently polarity-flipped with probability $p_{\neg}=0.5$, and the flip is applied consistently to all of its occurrences in the axiom set.
This final polarity randomization changes the surface polarity distribution while preserving the derivability of the constructed proof.

\subsection{Multi-Choice Instance Assembly}\label{app:stage2}
Algorithm~\ref{alg:merge_branches} assembles a $B$-choice problem from $B$ independently generated axiom-goal pairs $\{(S_i,goal_i)\}_{i=1}^B$.
Without loss of generality, the first proof tree is kept intact, so $goal_1$ remains provable and serves as the unique correct candidate (the candidate order is randomized later in Appendix~\ref{app:stage3}).
For each remaining proof tree $i=2,\ldots,B$, the generator selects an axiom $a^\star$ uniformly at random from $S_i$ and corrupts it.
If negation is not enabled, the selected axiom is removed.
If negation is enabled, the axiom is either removed with probability $p_{\mathrm{remove}}=0.5$ or corrupted by uniformly sampling one literal in the axiom and flipping its polarity.
Because Algorithm~\ref{alg:generate_instance} constructs each candidate with a unique derivation, corrupting one selected axiom breaks the only derivation of $goal_i$, making it unprovable.
The corrupted axiom set is then merged into the global axiom set $S$, and $goal_i$ is added to the candidate list.

After merging all $B$ proof trees, the generator optionally adds distractor rules to increase local ambiguity.
In our implementation, the total number of rule applications is capped at $BD_\mathrm{max}$, where $D_\mathrm{\max}$ denotes the maximum depth of the corresponding training configuration.
We sample a distractor count $m \sim \mathrm{Uniform}\{0,\ldots,M\}$ with $M=5$, and then add at most $m$ distractor rules without exceeding the $BD_\mathrm{max}$ edge cap.
Each distractor follows the rule forms permitted by $\Phi$, but is constructed so that at least one side of the rule, either the premise set $P_d$, the conclusion set $C_d$, or both, contains only fresh predicates not used elsewhere in the instance.
If only one side is fresh, the remaining side is sampled from the literals already present in the current instance.
For example, a distractor might state ``If Alice is a cat, then Alice is small'' (\logic{cat(Alice) $\to$ small(Alice)}) where “small” is a fresh predicate absent from all proof trees.
Consequently, each distractor can interact with the existing derivation graph on at most one side: it may be triggered by existing literals but only derive fresh literals, or it may require fresh literals and therefore cannot be triggered by the existing axioms.
This increases local ambiguity without changing the unique provable answer.

\begin{algorithm}[t]
\caption{Natural Language Conversion}
\label{alg:to_nl}
\KwIn{axiom set $\mathcal{S}$, candidate conclusions $\mathcal{C}$}
\KwOut{prompt $x$, answer $y$}

$\phi_e \leftarrow$ random bijection from entity IDs to names \tcp*{e.g., $e_0 \mapsto$ ``Alice''}
$\phi_p \leftarrow$ random bijection from literal IDs to predicates \tcp*{e.g., $u_3 \mapsto$ ``furry''}

$F \leftarrow [\,]$ \tcp*{natural-language facts}
\ForEach{$s \in \textsc{Shuffle}(\mathcal{S})$}{
    \uIf{$s$ is a given literal}{
        $F.\textsc{append}(\textsc{ToText}(s, \phi_e, \phi_p))$ \tcp*{e.g., ``Alice is furry.''}
    }
    \Else{
        $f \leftarrow \textsc{RuleToText}(s, \phi_e, \phi_p)$\;
        \If{$s$ involves a universal entity}{
            $f \leftarrow \textsc{Quantify}(f)$ \tcp*{e.g., ``If anyone is X, they are Y.''}
        }
        $F.\textsc{append}(f)$\;
    }
}

$C_{\text{nl}} \leftarrow [\,]$\tcp*{natural-language conclusions}
\ForEach{$c \in \textsc{Shuffle}(\mathcal{C})$}{
    $C_{\text{nl}}.\textsc{append}(\textsc{ToText}(c, \phi_e, \phi_p))$\;
}

$y \leftarrow$ the unique provable candidate in $C_{\text{nl}}$\;
$x \leftarrow \textsc{FormatPrompt}(F, C_{\text{nl}})$\;
\Return{$(x, y)$}\;
\end{algorithm}

\subsection{Natural Language Conversion}\label{app:stage3}

Algorithm~\ref{alg:to_nl} converts the symbolic instance into a natural-language multiple-choice prompt.
For each instance, we first sample a bijection from internal entity IDs to a fixed pool of $26$ names, shown below.

\begin{center}
\setlength{\fboxsep}{6pt}
\fbox{
\begin{minipage}{0.92\linewidth}
\small
\textbf{Name pool.}
Alice, Bob, Carol, David, Emma, Frank, Grace, Henry, Irene, Jack,
Kate, Leo, Mona, Nick, Olivia, Paul, Quincy, Rachel, Sam, Tina,
Ulysses, Victoria, Wendy, Xavier, Yvonne, and Zach.
\end{minipage}
}
\end{center}

We also sample a separate bijection from predicate IDs to random $5$-letter predicate strings.
These mappings are resampled independently for every instance, preventing the model from relying on fixed entity identities or real-world predicate semantics.

Each symbolic axiom is then rendered with predefined templates: literal axioms become factual statements (e.g., Alice is abcde), grounded rules become if-then statements (e.g., If Alice is abcde, Alice is bcdef), and universal rules are expressed with quantified templates such as ``If anyone is abcde, then they are bcdef''.
The rendered facts are shuffled before being placed in the prompt, and the candidate conclusions are also shuffled to remove ordering cues.
The final answer $y$ is the natural-language rendering of the unique provable candidate, while the prompt $x$ consists of the rendered facts with the shuffled candidate list using the templates in Appendix~\ref{app:template}.

\section{Detailed Description of Expressiveness}\label{app:expressiveness}

This section provides the full description of the five expressiveness levels summarized in the main text (Section~\ref{sec:difficulty}). Table~\ref{tab:expressiveness-mapping} summarizes these levels and their features.

\begin{table}[h]
\centering
\caption{The five levels of expressiveness and their features.}
\label{tab:expressiveness-mapping}
\begin{tabular}{lccccc}
\toprule
\textbf{Setting} & $\rightarrow$ & $\land$ & $\neg$ & $\lor$ & $\forall$ \\
\midrule
Implication-only & \checkmark & -- & -- & -- & -- \\
+ Conjunction    & \checkmark & \checkmark & -- & -- & -- \\
+ Negation       & \checkmark & \checkmark & \checkmark & -- & -- \\
+ Disjunction    & \checkmark & \checkmark & \checkmark & \checkmark & -- \\
+ Quantification & \checkmark & \checkmark & \checkmark & \checkmark & \checkmark \\
\bottomrule
\end{tabular}
\end{table}

\subsection{Implication-only}
In the implication-only setting, each axiom is either a grounded literal or a simple implication rule.
Starting from the literal axioms, reasoning proceeds by repeatedly applying rules whose antecedents are satisfied, thereby deriving new literals. A statement is considered valid if it can be derived from the axioms through a sequence of such rule applications.
Performing deductive reasoning under this logic is equivalent to path-finding in simple directed graphs.

\subsection{+~Conjunction}
In this setting, a rule may take the form of a conjunction of grounded literals entailing a single conclusion.
Accordingly, at inference time, the model must jointly track and verify multiple supporting literals at each reasoning step.
Compared to implication-only reasoning, this shifts the task from single-premise inference to coordinating multiple supporting literals before each rule application.
Performing deductive reasoning under this logic is equivalent to path-finding in simple directed hypergraphs where each hyperedge may have more than one source node.

\subsection{+~Negation}
In this setting, rules may condition on negated properties or derive negated consequents.
Compared to the conjunction setting, negation further requires the model to maintain polarity information throughout the proof, introducing a qualitatively different form of reasoning.\footnote{We note that our proofs do not contain hypothetical derivations, such as proofs by contradiction.}
As a byproduct, negation provides a natural mechanism for constructing non-target candidate conclusions: flipping the polarity of an intermediate literal within a proof-tree axiom breaks the derivation and renders the root conclusion unprovable. For example, \logic{mammal(Alice) $\to$ $\neg$bird(Alice)} can be corrupted into either \logic{$\neg$mammal(Alice) $\to$ $\neg$bird(Alice)} or \logic{mammal(Alice) $\to$ bird(Alice)}, depending on whether the sampled literal is the premise or the conclusion (cf.\ Section~\ref{sec:generation}; algorithm is provided in Appendix~\ref{app:stage2}).

\subsection{+~Disjunction}
In this setting, a rule may take the form of the antecedents entailing a disjunction of grounded literals.
Applying such a rule yields a set of candidate conclusions rather than a single deterministic outcome.
A consequence of this is that the simple proof tree analogy described in Section~\ref{sec:generation} must be extended: With disjunction, each proof step may have more than one literal as the conclusion. As such, each edge in the proof tree must be generalized as a hyperedge, which can have more than one parent node and more than one child node.
This creates an ambiguity that is absent in ordinary proof trees: a disjunctive rule establishes a disjunction, but not any individual disjunct by itself.
For example, suppose the current root literal is \logic{mammal(Alice)}, and the backward construction introduces the rule \logic{cat(Alice) $\rightarrow$ mammal(Alice)}, making \logic{cat(Alice)} a literal that must be supported.
If we then try to support this literal using \logic{pet(Alice) $\rightarrow$ cat(Alice) $\lor$ dog(Alice)} together with \logic{pet(Alice)}, we can derive \logic{cat(Alice) $\lor$ dog(Alice)}, but not \logic{cat(Alice)} itself.
Therefore, the original root literal \logic{mammal(Alice)} is not guaranteed to be derivable unless the alternative disjunct \logic{dog(Alice)} is resolved.

To keep the proof well-defined during our ``backwards'' proof generation procedure, we resolve each alternative disjunct in one of two standard ways: (i)~it is ruled out by additional axioms that contradict it (e.g., a separate axiom states \logic{$\neg$dog(Alice)}), or (ii)~it eventually leads to a common downstream conclusion shared with the chosen disjunct (e.g., add a rule \logic{dog(Alice) $\rightarrow$ mammal(Alice)} so that both \logic{cat(Alice)} and \logic{dog(Alice)} lead to \logic{mammal(Alice)}).
At inference time, the model must therefore examine which disjuncts are eliminated and which converge on a shared conclusion, shifting the task from committing to a single outcome at each step to reasoning over a set of possible conclusions.

\subsection{+~Quantification}
In this setting, universal quantification enables defining rules that apply to \emph{any} entity rather than to a specific one.
We also consider the number of entities in the problem as an additional axis of complexity. In many of our experiments, we consider the simplest case where all generated literals apply to a single entity (e.g., \logic{Alice}). We also consider a multi-entity setting, where axioms may involve several distinct entities within the same problem (e.g., \logic{Alice} and \logic{Bob}); by default, we use two entities. The use of multiple entities is particularly important when rules are universally quantified: without multiple entities, quantified rules reduce to entity-specific ones and offer no additional reasoning challenge.
With multiple entities present, we additionally enable \emph{predicate and rule reuse}: the same predicate may apply to different entities, and a single quantified rule may be instantiated multiple times within the same proof, creating compositional overlap that requires the model to disambiguate shared structure rather than treating each occurrence independently.
Together, these extensions move the environment from basic propositional inference towards the more expressive regime of first-order reasoning.

\section{Validity of the Construction}\label{app:validity}

We provide two complementary checks to ensure that our synthetic problems are free of surface-form shortcuts and are logically sound.

\subsection{Shortcut Controls}\label{app:shortcut_prevention}

We control surface-form shortcuts through the generation and rendering procedure described in Appendix~\ref{app:algorithm}.
Predicates are sampled as fresh random strings and entity-name mappings are resampled independently for each instance, preventing stable lexical or world-knowledge cues.
For unprovable candidates, the corrupted axiom is selected uniformly from the corresponding proof tree, preventing the position of the corrupted axiom from revealing the answer.
When negation is enabled, polarity-flip corruptions are also randomized, preventing polarity statistics from serving as a shortcut.
Finally, both the axiom order and candidate order are shuffled before rendering, preventing ordering cues in the prompt or answer options.
These controls are designed to ensure that the correct answer is determined by logical derivability rather than by surface statistics.

\subsection{Logical Soundness via Z3 Verification}\label{app:z3}

Although our backward-construction procedure is designed to produce correct labels, implementation imperfections may still introduce unwanted label noise.
To further verify the correctness of the labels, we audit a random subset of generated problems using the Z3 SMT solver~\citep{demoura2008z3}. For each sampled problem, we encode the axiom set as a first-order theory in Z3 and test entailment for every candidate literal. An instance passes the audit only if Z3 confirms that the marked-provable candidate is entailed and all $B-1$ marked-unprovable candidates are not entailed.
We sample $N\!=\!1000$ problems from every configuration used in our experiments, and all sampled instances pass the audit, confirming that the labels used during RL training are logically reliable.

\begin{table}[t]
\centering
\caption{Default RL Training Hyperparameters}
\label{tab:train_hparams}
\begin{tabular}{l c}
\toprule
\textbf{Parameter} & \textbf{Value} \\
\midrule
Generate batch size        & 384 prompts\\
Training batch size        & 256 prompts\\
Mini batch size            & 64 prompts\\
Samples per prompt         & 8 \\
Max response length        & 8192 \\
Clip ratio                 & (0.20, 0.28) \\
Temperature                & 1.0 \\
Top p                      & 1.0 \\

\midrule

Optimizer                  & AdamW \\
KL coefficient     & 0 \\
Actor learning rate        & $1 \times 10^{-6}$ \\
LR warmup steps    & 10 \\
LR scheduler       & constant \\

\bottomrule
\end{tabular}
\end{table}

\section{Full Implementation Details}
\label{app:implementation}

\subsection{Experimental Details}

\paragraph{Training.}
We conduct all RL post-training with the \texttt{verl} library~\citep{sheng2024hybridflow} on the non-thinking versions of Qwen3-4B and Qwen3-8B~\citep{yang2025qwen3} and Ministral3-3B-Instruct-2512~\citep{liu2026ministral}. Most runs are performed on 8$\times$B200 180G GPUs (see Figure~\ref{fig:effort-fits}). After completing the main experiments, we additionally ran a small number of configurations on 8$\times$H100 GPUs using the same training recipe to increase coverage in overlapping depth ranges in the scaling analysis of Figure~\ref{fig:scaling-power-law}. These additional runs are used only for the scaling analyses.
When selecting depth values, we exclude configurations for which the base model is already near saturation, defined as $\geq 85\%$ validation accuracy before RL, as they provide little measurable learning range.
Unless otherwise specified, all experiments share the training recipe described in Section~\ref{sec:rl_grpo}, with hyperparameters summarized in Table~\ref{tab:train_hparams}. Full training dynamics are provided in Appendix~\ref{app:dynamics}.

\paragraph{Evaluation.}
For all downstream benchmarks, we follow the official decoding parameters recommended by Qwen~\citep{yang2025qwen3}: temperature $T = 0.7$, top-$p = 0.8$, and top-$k = 20$. The maximum response length is set to 8192 tokens, matching the training budget. We record \texttt{Avg@8} and \texttt{Pass@8} for each benchmark and report the mean accuracy across benchmarks (i.e., the mean of per-benchmark \texttt{Avg@8} scores) as our primary metric in Section~\ref{sec:downstream}.

For the out-of-distribution generalization experiments in Section~\ref{sec:ood}, where evaluation depths can substantially exceed the training depth, we increase the maximum response length to 32{,}768 tokens. We find this is necessary because solving deeper OOD problems requires substantially longer reasoning than seen during training, and a tighter budget would truncate completions before they reach an answer, conflating reasoning capability with generation-length limits. All other decoding settings remain unchanged; we also report \texttt{Avg@8} as the main metric.

\subsection{Curriculum Implementation}
\label{app:curriculum}

We implement the curriculum strategy used in Section~\ref{sec:distribution} following the threshold-triggered scheme of \citet{stojanovski2025reasoninggym}.
At each training step, tasks are sampled uniformly from proof depths $\{1, \dots, D_{\mathrm{cur}}\}$, where $D_{\mathrm{cur}}$ is the current curriculum depth; instances beyond $D_{\mathrm{cur}}$ are filtered out. Whenever the rolling training accuracy reaches
$70\%$, we increase $D_{\mathrm{cur}}$ by a fixed step size
$\Delta$, until it reaches the maximum training depth $D_{\max}$. We use the curriculum hyperparameters in the following table. The initial $D_{\mathrm{cur}}$ is chosen as the smallest depth at which the base model is not yet saturated; shallower depths are already solved at or near the threshold and would trigger immediate curriculum advancement. We focus on demonstrating the effectiveness of curriculum-based training; designing more efficient curricula and selecting their hyperparameters more systematically is left to future work.

\begin{center}
\begin{tabular}{lcc}
\toprule
\textbf{Setting} & $D_{\mathrm{cur}}$ & $\Delta$ \\
\midrule
+~Conjunction    & $8$ & $4$ \\
+~Quantification & $4$ & $2$ \\
\bottomrule
\end{tabular}
\end{center}

For comparison, the default uniform setting is trained on depths uniformly distributed over ${1, \dots, D_{\max}}$ throughout, while \emph{difficult-only} training is trained exclusively on depth $D_{\max}$. All three strategies use identical training hyperparameters (Table~\ref{tab:train_hparams}).

\section{Prompt Templates}\label{app:template}

We provide the prompt templates used for both training and downstream evaluation, ensuring consistency and reproducibility across tasks.

\begin{examplebox}[Prompt Template for Logical Reasoning Task]
Suppose we have the following facts:
\newline

\{List of shuffled axioms and rules\}
\newline

Here are some candidate statements:
\newline

\{List of shuffled candidate statements\}
\newline

Exactly one of the above statements can be logically derived from the given facts.

Identify that statement and return it within <answer> </answer> tags. Example: <answer> Alice is abcde </answer>.

\end{examplebox}

\begin{examplebox}[Prompt Template for Math Benchmarks]
\{Problem\}
\newline

Please reason step by step, and put your final answer within \textbackslash boxed\{\}.

\end{examplebox}

\begin{examplebox}[Prompt Template for GPQA-Diamond and MMLU-Pro]
\{Problem\}
\newline

Please reason step by step, and put your final answer within \textbackslash boxed\{\}. Example: \textbackslash boxed\{A\}.

\end{examplebox}

\section{Scaling Robustness}
\label{app:Scaling_robustness}

We conduct eight complementary checks to verify that our main scaling claims are robust to key design and analysis choices.

\subsection{Power-Law vs.\ Exponential Fit Comparison}\label{app:fit_comparison}

To validate the power-law scaling relationship reported in Section~\ref{sec:scaling-depth}, we compare power-law and exponential fits for each expressiveness setting. Both models are fit via ordinary least squares in log space with two free parameters each, enabling direct comparison through the Akaike Information Criterion (AIC).

\begin{table}[h]
\centering
\caption{Power-law ($T = a \cdot D^{\gamma}$) vs.\ exponential ($T = a \cdot e^{bD}$) fit comparison across expressiveness levels. Both models use two free parameters and are fit via OLS in log space. $\Delta$AIC = AIC$_{\text{exp}}$ $-$ AIC$_{\text{pow}}$; positive values favor the power-law model.}
\label{tab:fit-comparison}
\begin{tabular}{lcccccc}
\toprule
\multirow{2}{*}{\textbf{Setting}} & \multicolumn{2}{c}{\textbf{Power-law}} & \multicolumn{2}{c}{\textbf{Exponential}} & \multirow{2}{*}{$\Delta$AIC} \\
\cmidrule(lr){2-3} \cmidrule(lr){4-5}
 & $\gamma \pm \text{SE}$ & $R^2$ & $b \pm \text{SE}$ & $R^2$ & \\
\midrule
Implication-only     & $1.05 \pm 0.03$ & $0.995$ & $0.024 \pm 0.003$ & $0.899$ & $+23.4$ \\
+ Conjunction    & $1.67 \pm 0.05$ & $0.993$ & $0.113 \pm 0.008$ & $0.967$ & $+13.9$  \\
+ Negation    & $1.77 \pm 0.04$ & $0.996$ & $0.120 \pm 0.008$ & $0.972$ & $+17.5$ \\
+ Disjunction       & $2.11 \pm 0.09$ & $0.993$ & $0.174 \pm 0.024$ & $0.929$ & $+14.1$ \\
+ Quantification     & $2.60 \pm 0.06$ & $0.998$ & $0.318 \pm 0.031$ & $0.964$ & $+17.6$ \\
\bottomrule
\end{tabular}
\end{table}

\begin{figure}[t]
    \centering
    \includegraphics[width=\textwidth]{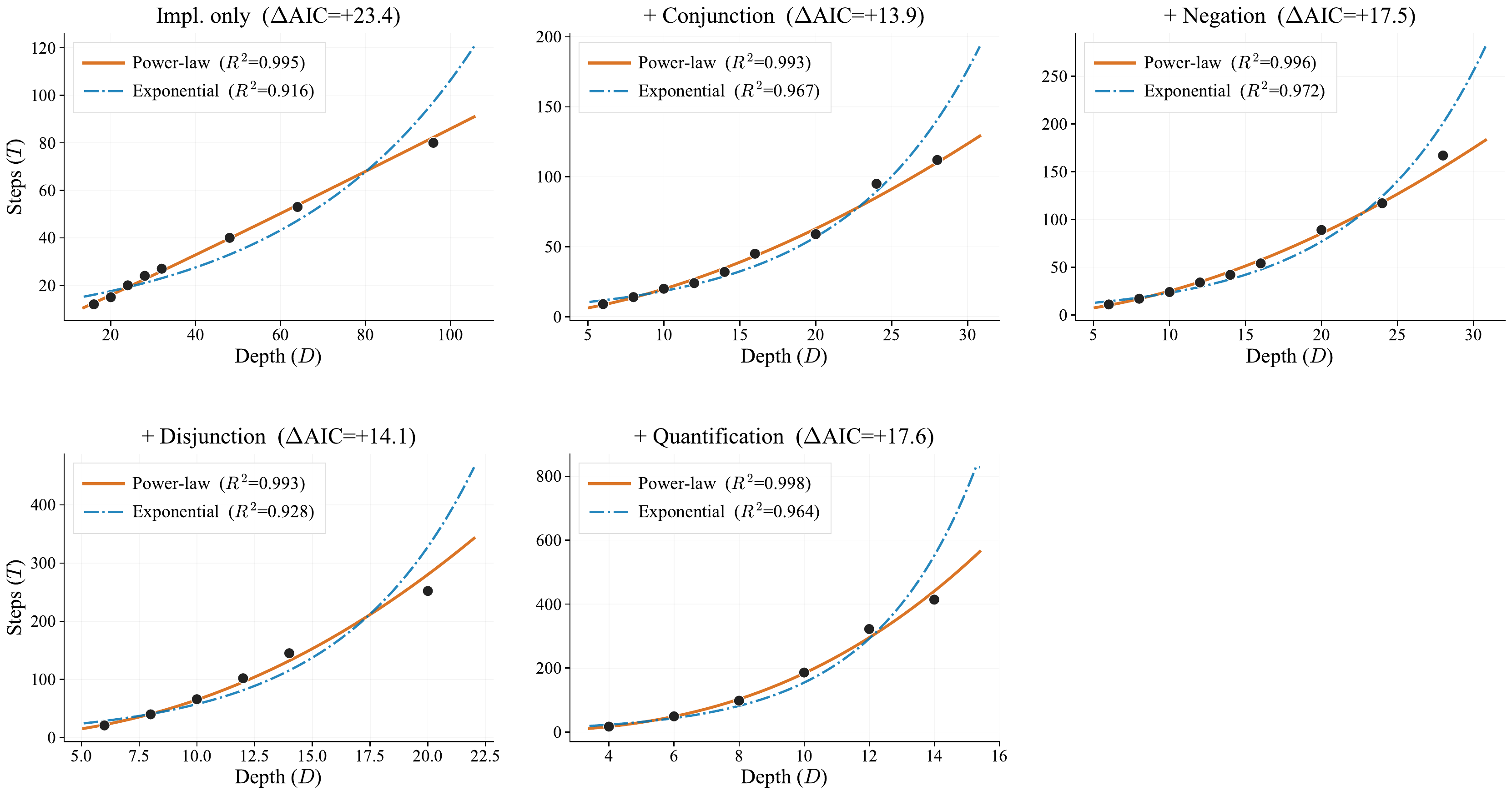}
    \caption{Power-law (orange) vs.\ exponential (blue) fits for each expressiveness setting. $\Delta$AIC values consistently favor the power-law model.}
    \label{fig:fit-comparison}
\end{figure}

As shown in Table~\ref{tab:fit-comparison} and Figure~\ref{fig:fit-comparison}, the power-law model achieves higher $R^2$ and lower AIC than the exponential model across all five settings ($\Delta$AIC $\geq +13.9$). Figure~\ref{fig:fit-comparison} visualizes both fits against the observed data, showing that the exponential model systematically overshoots at large depths while the power-law fit remains accurate throughout.

\subsection{Sensitivity to the Accuracy Threshold}\label{app:threshold}

\begin{figure}[h]
\centering
\begin{subfigure}{0.49\textwidth}
\centering
\includegraphics[width=\linewidth]{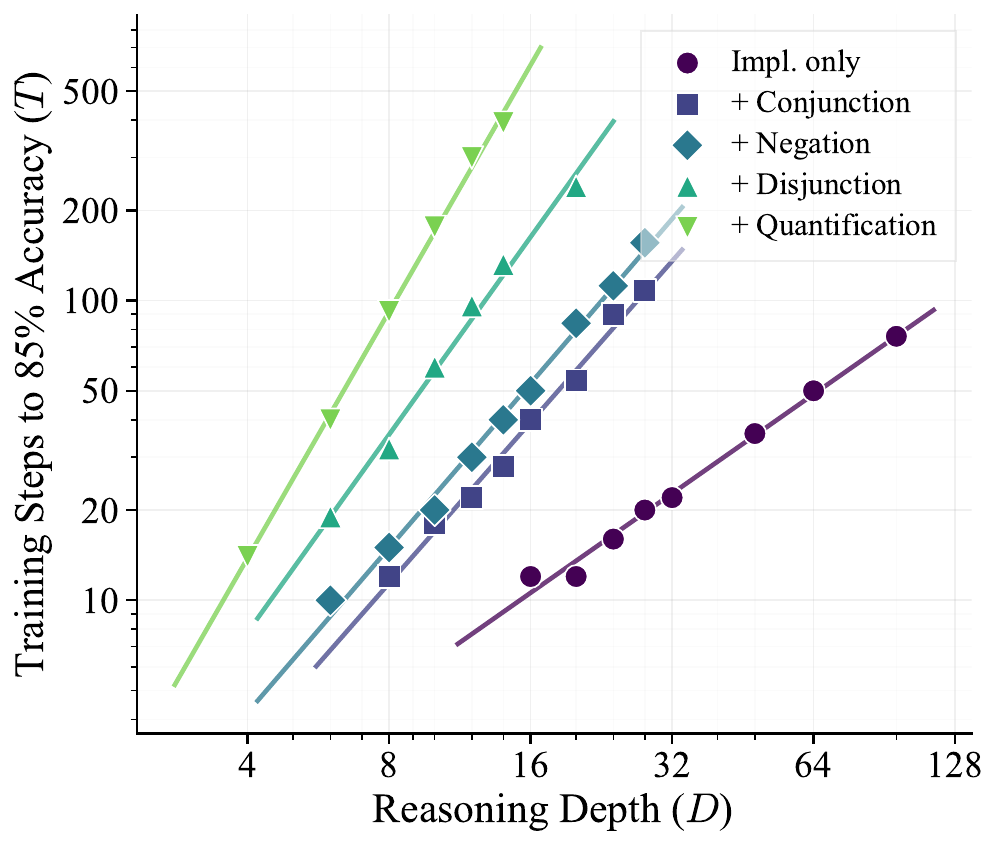}
\caption{Log-log scaling curves at $85\%$ threshold.}
\end{subfigure}
\hfill
\begin{subfigure}{0.49\textwidth}
\centering
\includegraphics[width=\linewidth]{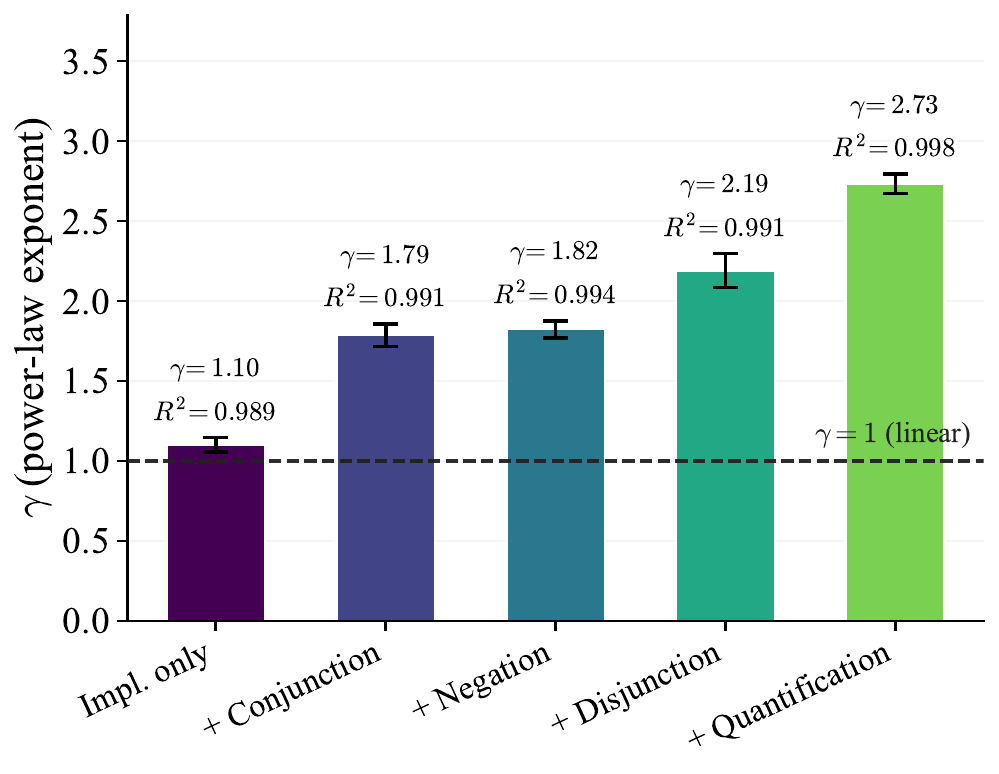}
\caption{Fitted $\gamma$ vs.\ expressiveness at $85\%$  threshold.}
\end{subfigure}
\caption{Power-law scaling under $85\%$ accuracy threshold (cf.\ main-text Figure~\ref{fig:scaling-power-law} for $90\%$). The qualitative pattern is unchanged: all five settings follow a power law ($R^{2} \geq 0.989$), and $\gamma$ increases monotonically with expressiveness.}
\label{fig:scaling_85}
\end{figure}

To verify that our scaling claims are not artifacts of the specific $90\%$ accuracy threshold used in the main text, we refit $T \propto D^{\gamma}$ using a lower threshold of $85\%$ (Figure~\ref{fig:scaling_85}). As shown in Table~\ref{tab:threshold_comparison}, the fitted exponents shift only modestly across all five settings.

All five settings continue to admit a clean power-law fit ($R^{2} \geq 0.989$), and $\gamma$ remains monotonically increasing in expressiveness ($1.10 \to 2.73$). The relative ordering of the five exponents is preserved exactly. Our main conclusions are therefore not sensitive to the specific accuracy threshold used.

\begin{table}[t]
\centering
\caption{Power-law fits at the main-text accuracy threshold ($\mu=90\%$) versus a relaxed threshold ($\mu=85\%$). All five expressiveness settings preserve a clean power-law fit under both thresholds, and the relative ordering of the fitted exponents is preserved exactly.}
\label{tab:threshold_comparison}
\begin{tabular}{lcccc}
\toprule
& \multicolumn{2}{c}{$\mu = 90\%$} & \multicolumn{2}{c}{$\mu = 85\%$} \\
\cmidrule(lr){2-3} \cmidrule(lr){4-5}
Setting & $\gamma \pm \text{SE}$ & $R^{2}$ & $\gamma \pm \text{SE}$ & $R^{2}$ \\
\midrule
Implication-only   & $1.05 \pm 0.03$ & $0.995$ & $1.10 \pm 0.05$ & $0.989$ \\
+ Conjunction      & $1.67 \pm 0.05$ & $0.993$ & $1.79 \pm 0.07$ & $0.991$ \\
+ Negation         & $1.77 \pm 0.04$ & $0.996$ & $1.82 \pm 0.05$ & $0.994$ \\
+ Disjunction      & $2.11 \pm 0.09$ & $0.993$ & $2.19 \pm 0.11$ & $0.991$ \\
+ Quantification   & $2.60 \pm 0.06$ & $0.998$ & $2.73 \pm 0.06$ & $0.998$ \\
\bottomrule
\end{tabular}
\end{table}

\subsection{Different Metrics of Training Compute}\label{app:training_effort}

The main text uses RL training steps as the measure of training effort $T$. To check that our scaling claims do not depend on this choice, we refit $T \propto D^{\gamma}$ under four alternative compute measures: total generated tokens across all rollouts including rejected samples $T_{\text{gen-tok}}$; prompt and response tokens in the kept batch $T_{\text{upd-tok}}$; training FLOPs by the Kaplan rule~\citep{kaplan2020scaling}, $T_{\text{FLOPs}} = 2N\,T_{\text{gen-tok}} + 6N\,T_{\text{upd-tok}}$ with $N$ the model's parameter count; and wall-clock GPU-hours on $8\!\times\!\text{B200}$.

\begin{table}[h]
\centering
\small
\setlength{\tabcolsep}{4pt}
\renewcommand{\arraystretch}{1.1}
\caption{Power-law fits $T(D)=a\cdot D^{\gamma}$ for four compute-effort measures.  $T_{\text{gen-tok}}$ counts generated tokens over all rollouts (including rejected samples); $T_{\text{upd-tok}}$ counts the prompt and response tokens of the kept batch only; $T_{\text{FLOPs}}=2N\,T_{\text{gen-tok}}+6N\,T_{\text{upd-tok}}$ follows the Kaplan rule; GPU-hrs is wall-clock $\times\,8$ B200 GPUs. Because wall-clock time is not comparable across hardware, the GPU-hrs column is fit only on the runs executed on this machine, a subset of the depth points used in the token- and FLOP-based columns.}
\label{tab:effort-fits}
\vspace{0.5em}
\begin{tabular}{l cc cc cc cc}
\toprule
\multirow{2}{*}{\textbf{Setting}} & \multicolumn{2}{c}{$T_{\text{gen-tok}}$}
        & \multicolumn{2}{c}{$T_{\text{upd-tok}}$}
        & \multicolumn{2}{c}{$T_{\text{FLOPs}}$}
        & \multicolumn{2}{c}{GPU-hrs} \\
\cmidrule(lr){2-3} \cmidrule(lr){4-5} \cmidrule(lr){6-7} \cmidrule(lr){8-9}
 & $\gamma \pm \text{SE}$ & $R^2$ & $\gamma \pm \text{SE}$ & $R^2$ & $\gamma \pm \text{SE}$ & $R^2$ & $\gamma \pm \text{SE}$ & $R^2$ \\
\midrule
Implication-only & $1.05\!\pm\!0.03$ & $0.995$ & $1.55\!\pm\!0.04$ & $0.997$ & $1.44\!\pm\!0.03$ & $0.995$ & $1.38\!\pm\!0.05$ & $0.994$ \\
+ Conjunction & $2.06\!\pm\!0.11$ & $0.981$ & $2.35\!\pm\!0.07$ & $0.994$ & $2.26\!\pm\!0.08$ & $0.991$ & $2.40\!\pm\!0.13$ & $0.989$ \\
+ Negation    & $2.12\!\pm\!0.05$ & $0.996$ & $2.32\!\pm\!0.02$ & $0.999$ & $2.26\!\pm\!0.03$ & $0.999$ & $2.33\!\pm\!0.06$ & $0.998$ \\
+ Disjunction & $2.42\!\pm\!0.15$ & $0.982$ & $2.72\!\pm\!0.11$ & $0.992$ & $2.62\!\pm\!0.12$ & $0.990$ & $2.61\!\pm\!0.15$ & $0.984$ \\
+ Quantification & $2.97\!\pm\!0.17$ & $0.987$ & $3.19\!\pm\!0.13$ & $0.994$ & $3.12\!\pm\!0.14$ & $0.992$ & $3.09\!\pm\!0.16$ & $0.990$ \\
\bottomrule
\end{tabular}
\end{table}

\begin{figure}[t]
\centering
\includegraphics[width=0.95\textwidth]{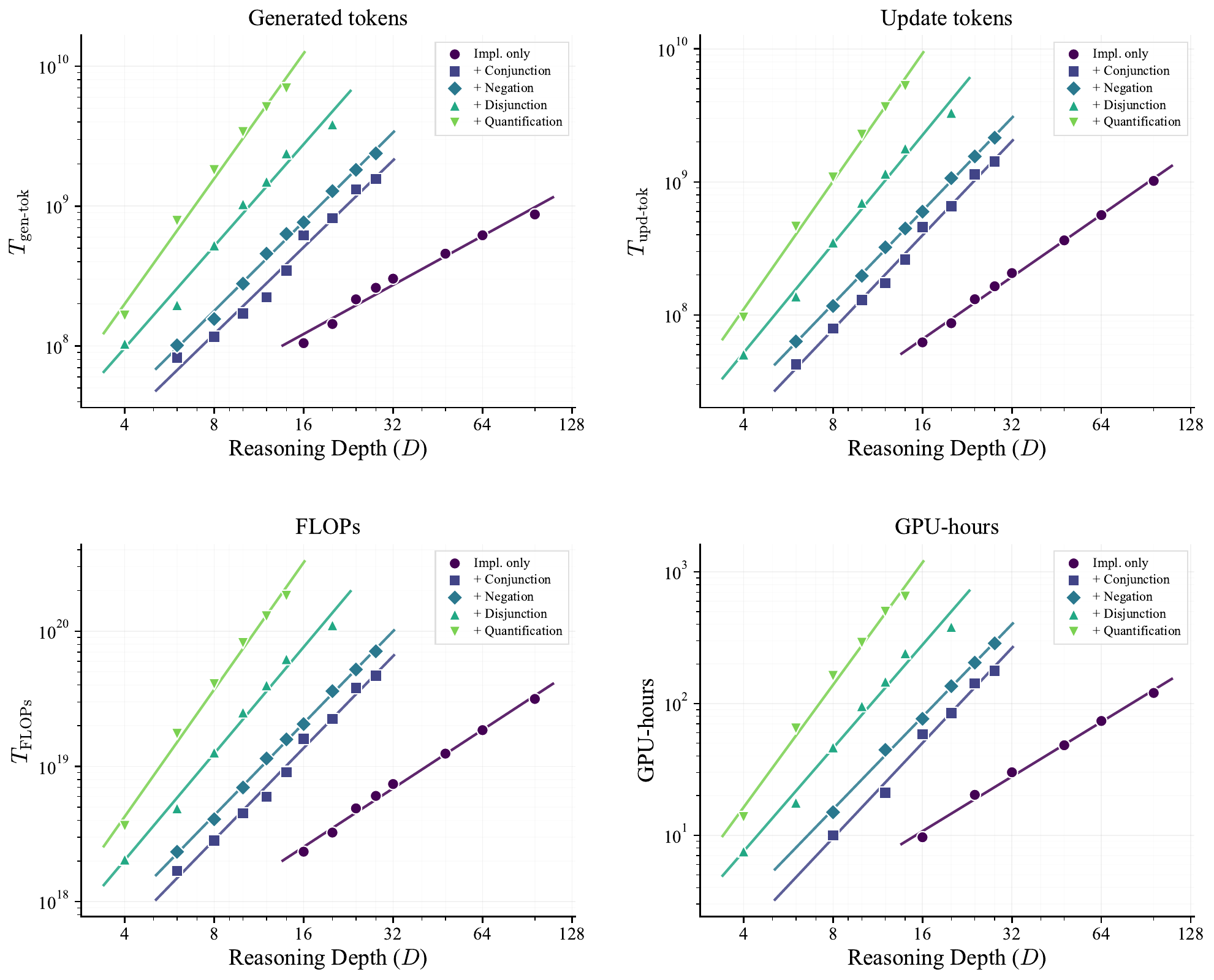}
\caption{Power-law fits $T(D) = a \cdot D^{\gamma}$ under four alternative compute measures, complementing Table~\ref{tab:effort-fits}. \textbf{Top-left:} generated tokens $T_{\text{gen-tok}}$. \textbf{Top-right:} kept-batch prompt and response tokens $T_{\text{upd-tok}}$. \textbf{Bottom-left:} training FLOPs $T_{\text{FLOPs}} = 2N\,T_{\text{gen-tok}} + 6N\,T_{\text{upd-tok}}$. \textbf{Bottom-right:} wall-clock GPU-hours on $8\!\times\!\text{B200}$ (fit on the runs from this machine only; see Table~\ref{tab:effort-fits}). All four measures preserve the power-law form ($R^{2} \geq 0.97$).}
\label{fig:effort-fits}
\end{figure}

Table~\ref{tab:effort-fits} and Figure~\ref{fig:effort-fits} report the fits. The power-law form holds in every cell ($R^{2} \geq 0.97$), and the overall ranking, \emph{Implication-only} $<$ \emph{+~Conjunction}, \emph{+~Negation} $<$ \emph{+~Disjunction} $<$ \emph{+~Quantification}, is preserved across all four measures. Exponents shift upward relative to step counts (e.g., \emph{+~Quantification} shifts from $\gamma = 2.60$ to $\gamma = 3.12$ under FLOPs), reflecting that token- and FLOP-based measures aggregate per-instance length on top of the number of gradient updates: prompt length grows with depth and expressiveness, and response length grows further during training. \emph{+~Conjunction} and \emph{+~Negation} remain within each other's standard-error ranges across all metrics, consistent with the main-text finding that the two impose comparable training burdens.

We retain training steps as our primary measure because they isolate the convergence rate from confounds in per-instance length. Token- and FLOP-based measures aggregate two such confounds: prompts grow structurally with reasoning depth and expressiveness (more axioms, longer rules, quantified templates), and within-training response length grows with both training progress and task difficulty (Figure~\ref{fig:all_trajectories}). Conflating these with training effort would mix structural and dynamical properties of the training environment into the scaling measurement.

\subsection{Variability across Random Seeds}\label{app:multi_seed}

The main scaling experiment in Section~\ref{sec:scaling-depth} uses a single random seed per configuration due to computational constraints. To verify that this does not bias the fitted exponent, we leverage the multi-seed runs from the training-distribution and cross-algorithm experiments (Sections~\ref{sec:distribution} and \ref{sec:cross-algorithm}), which use three seeds per depth on the \textit{+~Conjunction} setting. Pooling all three seeds and refitting yields $\gamma = 1.70$ ($R^{2} = 0.995$), within $0.02$ of the single-seed fit on the same depth points ($\gamma = 1.72$, $R^{2} = 0.991$). The marginally higher $R^{2}$ under pooling likely reflects per-depth variance being averaged out before fitting. This confirms that our single-seed scaling estimates are reliable.

\subsection{Multi-Entity Reasoning as a Potential Confounder}\label{app:multi_entity}

The most expressive setting (\textit{+~Quantification}) introduces multiple design elements simultaneously (Appendix~\ref{app:expressiveness}). A natural concern is that the elevated $\gamma$ at \textit{+~Quantification} could be partially driven by the multi-entity dimension rather than by quantification itself.

To isolate the multi-entity contribution, we run an ablation that varies only the number of entities while keeping $\forall$ and reuse disabled (i.e., starting from the \textit{+~Disjunction} setting). As shown in Table~\ref{tab:multi_entity_ablation}, training steps to $90\%$ accuracy are essentially identical between the single-entity ($1$ person) and multi-entity ($2$ persons) variants across $D \in \{6, 8, 10\}$, with absolute differences of at most $2$ steps and no systematic trend with depth.

\begin{table}[h]
\centering
\caption{Training steps to $90\%$ accuracy under the \textit{+~Disjunction} setting with single vs.\ multiple entities. Differences are within $\pm 2$ steps and show no systematic trend with depth, indicating that the multi-entity dimension by itself does not measurably increase training difficulty.}
\label{tab:multi_entity_ablation}
\begin{tabular}{ccc}
\toprule
$D$ & Single entity ($1$ person) & Multiple entities ($2$ persons) \\
\midrule
$6$  & $21$ & $21$ \\
$8$  & $39$ & $40$ \\
$10$ & $68$ & $66$ \\
\bottomrule
\end{tabular}
\end{table}

We conclude that the multi-entity design by itself contributes negligibly to training difficulty in this depth range, and that the elevated $\gamma$ at \emph{+~Quantification} therefore reflects the additional difficulty imposed by universal quantification and rule reuse rather than by the multi-entity dimension.

\subsection{Prompt Length as a Potential Confounder}\label{app:prompt_length}

Richer expressiveness settings tend to produce longer prompts because their rules contain more structural elements, such as multiple premises, multiple conclusions, and quantified templates. This raises a potential confound: the exponent ordering in Section~\ref{sec:scaling-depth} might reflect increasing prompt length rather than the reasoning structure induced by the logical operators. To examine this possibility, we reparameterize training cost in terms of the mean tokenized prompt length $L$ of each training set and fit $T \propto L^{\beta}$ separately for each expressiveness setting. If the observed ordering were primarily an artifact of prompt length, this reparameterization should substantially reduce the systematic separation across expressiveness levels.

\begin{figure}[h]
\centering
\includegraphics[width=0.55\textwidth]{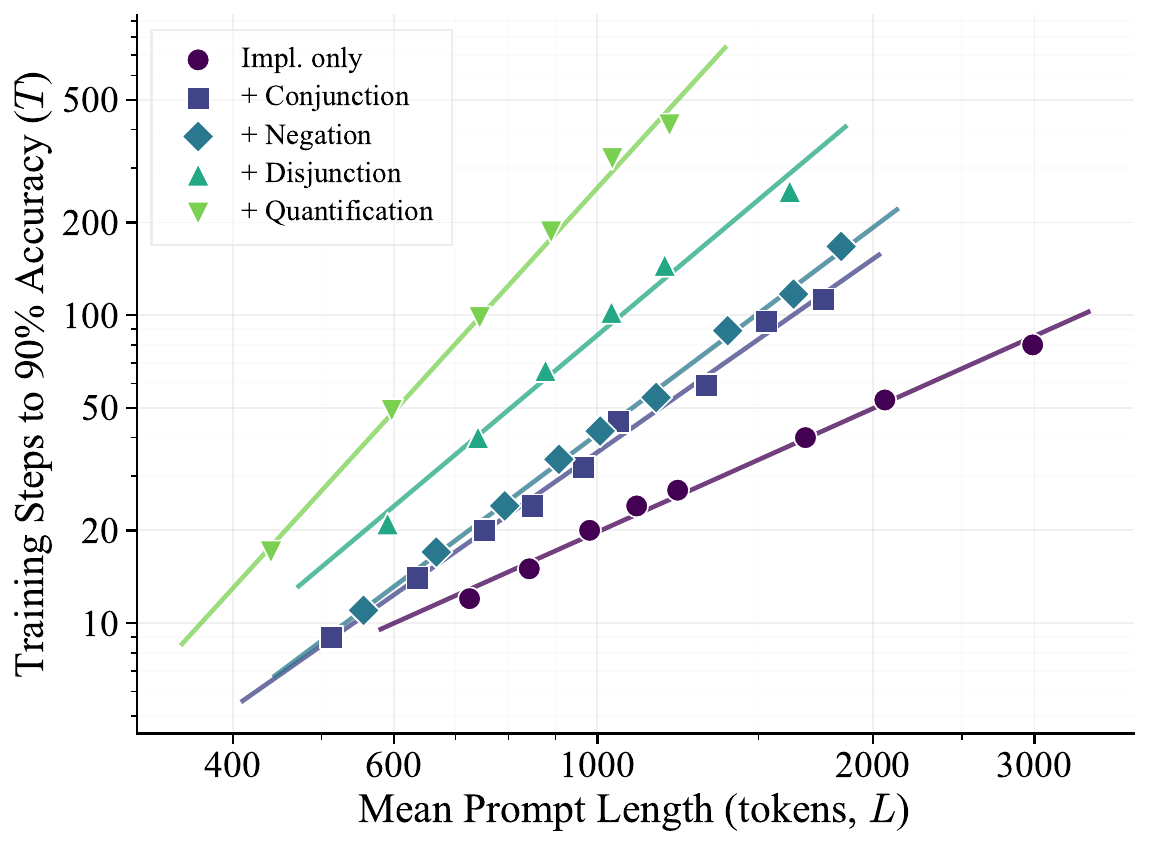}
\caption{
Training cost as a function of mean prompt length $L$ on log--log axes.
The fitted exponent $\beta$ increases monotonically with expressiveness (Table~\ref{tab:promptlen-fits}).}
\label{fig:scaling-promptlen}
\end{figure}

\begin{table}[h]
\centering
\caption{Power-law fits of training steps against mean prompt length, $T \propto L^{\beta}$, for each expressiveness setting.}
\label{tab:promptlen-fits}
\begin{tabular}{lcc}
\toprule
Setting & $\beta \pm \text{SE}$ & $R^{2}$ \\
\midrule
Implication-only & $1.33 \pm 0.04$ & $0.994$ \\
+ Conjunction    & $2.09 \pm 0.06$ & $0.994$ \\
+ Negation       & $2.23 \pm 0.03$ & $0.998$ \\
+ Disjunction    & $2.50 \pm 0.14$ & $0.988$ \\
+ Quantification & $3.27 \pm 0.10$ & $0.996$ \\
\bottomrule
\end{tabular}
\end{table}

Figure~\ref{fig:scaling-promptlen} and Table~\ref{tab:promptlen-fits} show that the separation persists after reparameterizing by prompt length. More directly, prompt length does not induce a consistent ordering of training difficulty across settings: \textit{Implication-only} at $D=96$ has an approximately $3{,}000$-token prompt but reaches the target accuracy in $80$ steps, whereas \textit{+~Quantification} at $D=8$ needs only $744$ tokens yet requires $98$ steps. This separation is not driven by disjoint prompt-length regimes: most runs across settings overlap within approximately 500 to 3,000 tokens, while their training costs remain systematically different.

Consistent with this observation, the fitted length exponent $\beta$ remains monotonically ordered across expressiveness levels, increasing from 1.33 for  \textit{Implication-only} to 3.27 for \textit{+~Quantification}, in lockstep with the depth exponents $\gamma$. The monotone $\beta$ ordering also replicates on Qwen3-8B ($\beta = 1.27, 1.86, 1.93, 2.24, 3.18$). Together, these results indicate that the observed exponent ordering is not explained by prompt length.

\subsection{Controlling for Surface-Form Differences}\label{app:template_control}

The five expressiveness settings share the same prompt scaffolding, question format, and entity set (Appendices~\ref{app:algorithm} and~\ref{app:template}); their main surface-level difference is the appearance of additional operator words as the rule language becomes richer: \emph{and} for conjunction, \emph{not} for negation, \emph{or} for disjunction, and \emph{for any} for quantification. These operator words cannot simply be removed from the higher-expressiveness settings without changing the underlying formal system. We therefore construct the reverse control: enriching the surface form of the \textit{Implication-only} setting while preserving its original semantics and proof structure. Specifically, each \texttt{Implication-only} rule is randomly rendered using one of two ``dressed'' templates that incorporate the surface form of the richer expressiveness settings:

\begin{itemize}[leftmargin=2em, itemsep=1pt, parsep=1pt, topsep=2pt]
    \item[] \textbf{plain:} ``If [entity] is [attribute$_1$], then [entity] is [attribute$_2$].''
    \item[] \textbf{dressed 1:} ``If [entity] is [attribute$_1$] and True, then [entity] is [attribute$_2$] or False.''
    \item[] \textbf{dressed 2:} ``For any X, if X is [attribute$_1$] and True, then X is [attribute$_2$] or False.''
\end{itemize}

Here, ``and True'' and ``or False'' use logical constants that require minimal verification and introduce no additional proof node or inference step. This differs from the genuine \textit{+~Conjunction} setting, where each conjunct must itself be established from the facts or preceding rules. Likewise, under the single-entity construction, the ``For any X'' wrapper is semantically vacuous and does not introduce rule reuse. We do not add \emph{not}, because we found no natural way to introduce negation into the surface form while preserving the original semantics.

This control yields a clear prediction: if the additional operator words themselves were responsible for the increased training difficulty, the dressed \textit{Implication-only} curve should shift toward the richer settings. Table~\ref{tab:template-control} shows the contrary. Across the depth range $D \in \{16, 24, 32, 48, 64\}$, dressing introduces only a small training overhead, with no systematic increase as depth grows. The fitted exponent remains within the uncertainty of the plain \textit{Implication-only} estimate, with $\gamma_{\text{dressed}} = 0.98 \pm 0.06$ versus $\gamma_{\text{plain}} = 1.06 \pm 0.04$, and stays far below $\gamma_{\textit{+Conjunction}} = 1.67 \pm 0.05$. At $D = 24$, \textit{+~Conjunction} requires $95$ updates, $3.8\times$ more training than dressed \textit{Implication-only}, despite the latter using the additional surface markers \emph{or}, \emph{and}, and \emph{for any}. Following the analysis of Appendix~\ref{app:prompt_length}, we also fit training cost as a function of prompt length, $T \propto L^{\beta}$. The exponent decreases from $\beta = 1.39 \pm 0.05$ for plain \textit{Implication-only} to $1.19 \pm 0.08$ for the dressed variant, remaining well below the \textit{+~Conjunction} reference of $2.09 \pm 0.06$; this decrease is expected because dressing inflates prompt length with semantically vacuous template tokens without adding proof-relevant information. Thus, adding richer operator words without their semantic complexity does not reproduce the higher training cost. The exponent hierarchy therefore appears to track the proof-search consequences of the operators, rather than their surface-form realization alone.

\begin{table}[h]
\centering
\small
\caption{Surface-form control: \textit{Implication-only} semantics rendered in the full-template syntax of the richer expressiveness levels. $^{*}$~marks reference values for \textit{+~Conjunction}, fitted on its full depth range.}
\label{tab:template-control}
\begin{tabular}{llcccccll}
\toprule
& & \multicolumn{5}{c}{$D$} & & \\
\cmidrule(lr){3-7}
\multicolumn{1}{c}{Setting} & & 16 & 24 & 32 & 48 & 64 & \multicolumn{1}{c}{Exponent $\pm$ SE} & \multicolumn{1}{c}{$R^{2}$} \\
\midrule
\multirow{2}{*}{Impl.\ only (plain)} & $T$ & 12 & 20 & 27 & 40 & 53 & $\gamma = 1.06 \pm 0.04$ & $0.996$ \\
 & $L$ & 725 & 981 & 1224 & 1687 & 2060 & $\beta = 1.39 \pm 0.05$ & $0.996$ \\
\midrule
\multirow{2}{*}{Impl.\ only (dressed)} & $T$ & 15 & 25 & 29 & 44 & 61 & $\gamma = 0.98 \pm 0.06$ & $0.990$ \\
 & $L$ & 901 & 1239 & 1600 & 2233 & 2778 & $\beta = 1.19 \pm 0.08$ & $0.986$ \\
\midrule
\multirow{2}{*}{+ Conjunction (reference)} & $T$ & 45 & 95 & --- & --- & --- & $\gamma = 1.67 \pm 0.05^{*}$ & $0.993$ \\
 & $L$ & 1055 & 1528 & --- & --- & --- & $\beta = 2.09 \pm 0.06^{*}$ & $0.994$ \\
\bottomrule
\end{tabular}
\end{table}

\subsection{Operator-Swap Ablation}\label{app:operator_swap}

The main hierarchy introduces operators in a fixed cumulative order (conjunction $\rightarrow$ negation $\rightarrow$ disjunction $\rightarrow$ quantification). We further test whether the exponent ordering is an artifact of the particular cumulative path.
We construct an additional \textit{+Conjunction+Disjunction} setting that introduces disjunction directly after conjunction, without negation, and train it over the depth range $D \in \{6, 8, 10, 12, 14\}$ using the same training recipe. Table~\ref{tab:operator-swap} compares this setting with the corresponding configurations from the main hierarchy, with all exponents refit on the same depth range.

\begin{table}[h]
\centering
\caption{Operator-swap ablation on the shared depth range $D \in \{6,8,10,12,14\}$. \mbox{+Conj}, \mbox{+Conj+Neg}, and \mbox{+Conj+Neg+Disj} correspond to the main hierarchy's \textit{+~Conjunction}, \textit{+~Negation}, and \textit{+~Disjunction} settings.}
\label{tab:operator-swap}
\begin{tabular}{lcccccll}
\toprule
& \multicolumn{5}{c}{$T$ at depth $D$} & & \\
\cmidrule(lr){2-6}
\multicolumn{1}{c}{Setting} & 6 & 8 & 10 & 12 & 14 &
\multicolumn{1}{c}{$\gamma \pm \text{SE}$} &
\multicolumn{1}{c}{$R^{2}$} \\
\midrule
+Conj           & 9  & 14 & 20 & 24  & 32  & $1.47 \pm 0.06$ & $0.996$ \\
+Conj+Neg       & 11 & 17 & 24 & 34  & 42  & $1.60 \pm 0.04$ & $0.998$ \\
+Conj+Disj      & 11 & 18 & 27 & 43  & 56  & $1.95 \pm 0.08$ & $0.995$ \\
+Conj+Neg+Disj  & 21 & 40 & 66 & 102 & 145 & $2.28 \pm 0.02$ & $1.000$ \\
\bottomrule
\end{tabular}
\end{table}

The alternative path $\textit{+Conj} \rightarrow \textit{+Conj+Disj} \rightarrow \textit{+Conj+Neg+Disj}$ also exhibits a strictly increasing exponent, $1.47 \rightarrow 1.95 \rightarrow 2.28$. Equivalently, adding disjunction increases the exponent both without negation ($1.47 \rightarrow 1.95$) and with negation ($1.60 \rightarrow 2.28$), while adding negation increases it both without disjunction ($1.47 \rightarrow 1.60$) and with disjunction ($1.95 \rightarrow 2.28$).
Thus, this operator-swap ablation shows that the monotone scaling trend is not an artifact of the particular order in the main cumulative hierarchy for negation and disjunction.

\begin{figure}[t]
\centering
\begin{subfigure}{0.49\textwidth}
\centering
\includegraphics[width=\linewidth]{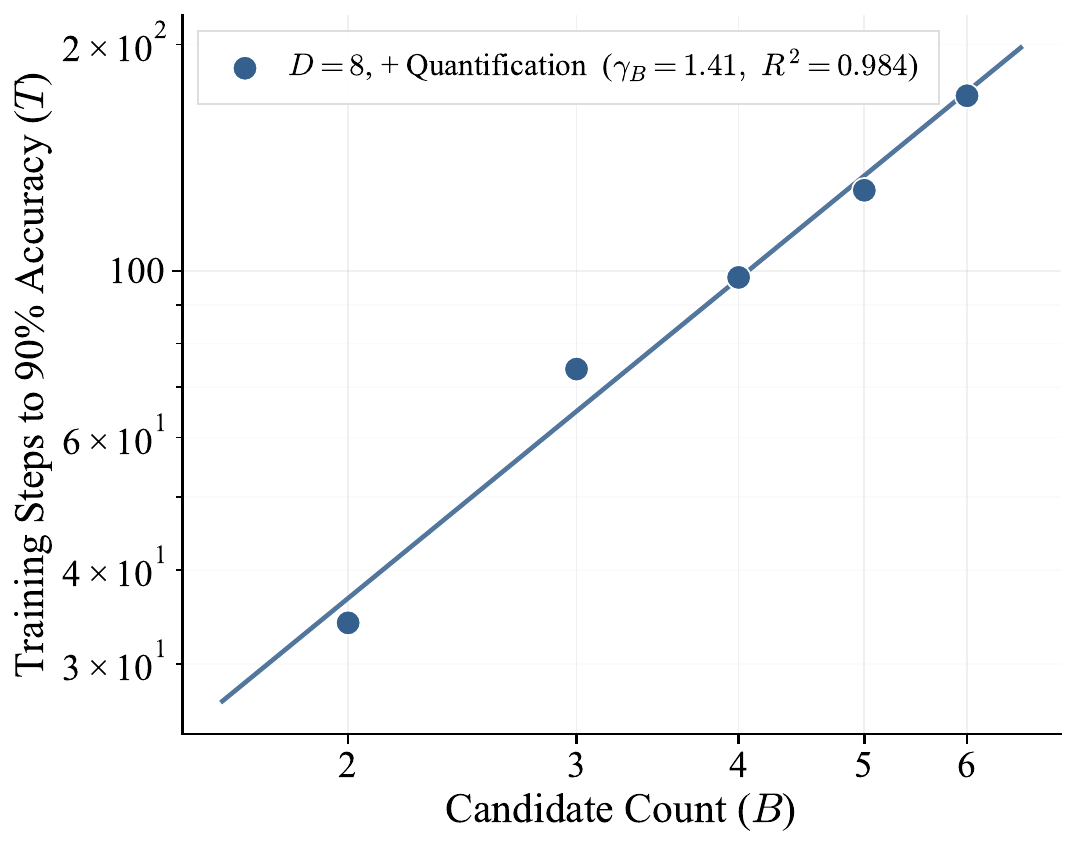}
\caption{Log-log scaling curves with power-law fits.}
\label{fig:branch-scaling_a}
\end{subfigure}
\hfill
\begin{subfigure}{0.49\textwidth}
\centering
\includegraphics[width=\linewidth]{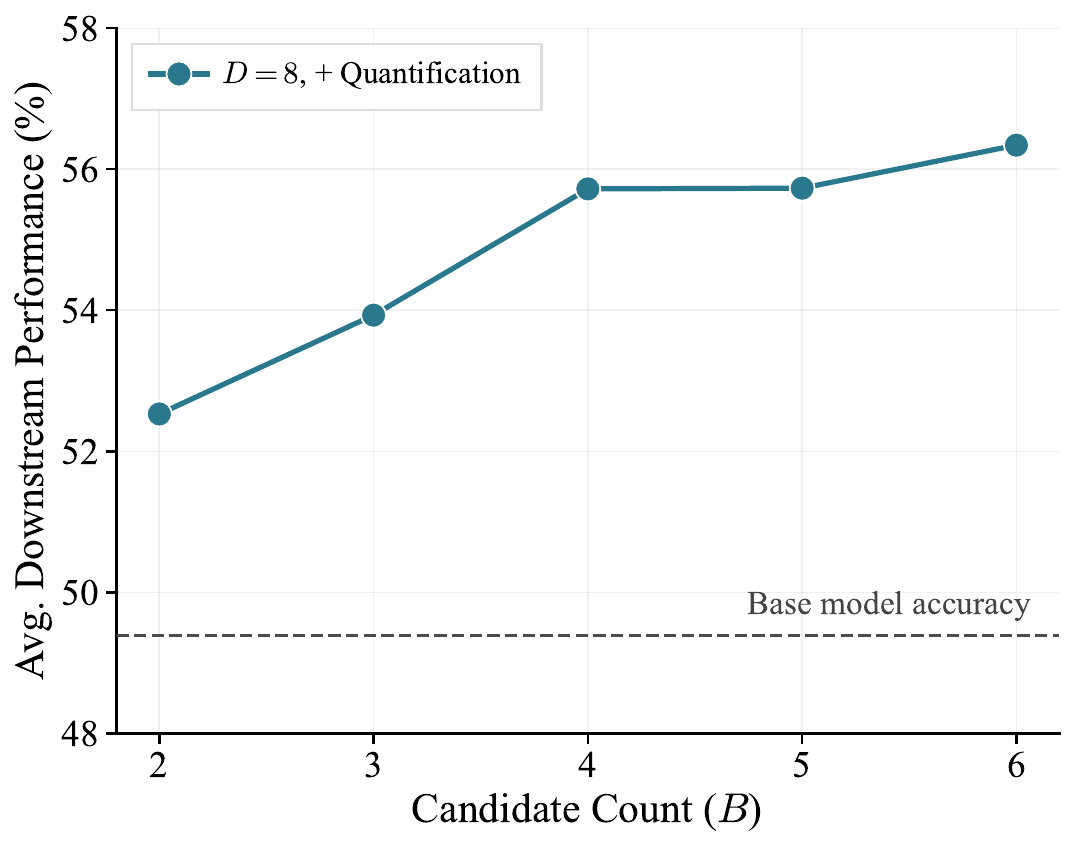}
\caption{Performance over candidate count.}
\label{fig:branch-scaling_b}
\end{subfigure}
\caption{Effect of candidate count $B$ at fixed depth $D = 8$ under the \emph{+~Quantification} setting. \textbf{(a)} Training steps to $90\%$ accuracy follow a power law in $B$ ($\gamma_B = 1.41$, $R^{2} = 0.984$, $\Delta\text{AIC} = +7.0$ vs.\ exponential). \textbf{(b)} Average downstream performance across the eight reasoning benchmarks of Section~\ref{sec:downstream}, plotted against candidate count $B$. Gains saturate quickly: performance rises from $52.5\%$ at $B\!=\!2$ to $55.7\%$ at $B\!=\!4$ ($+3.2$ pp), but only an additional $+0.6$ pp from $B\!=\!4$ to $B\!=\!6$.}
\label{fig:branch-scaling}
\end{figure}

\section{Additional Experimental Results}\label{app:additional-result}

\subsection{Candidate Count Scaling}
\label{app:branch-scaling}

We examine how the candidate count $B$ (i.e., the number of candidate conclusions per problem) affects both training compute and downstream transfer at fixed depth $D=8$ under the \emph{+~Quantification} setting.

As with depth scaling (Section~\ref{sec:scaling-depth}), we fit $T = a \cdot B^{\gamma_B}$ via OLS in log-log space. As shown in Figure~\ref{fig:branch-scaling_a}, training compute follows a power law in $B$ with exponent $\gamma_B = 1.41 \pm 0.10$ and $R^{2} = 0.984$. The power-law model again outperforms an exponential fit ($\Delta\text{AIC} = +7.0$).
The sub-quadratic exponent indicates that increasing the number of candidate branches is less costly than increasing reasoning depth under the same expressiveness setting (where $\gamma = 2.60$), consistent with the intuition that deeper proofs require compositionally harder reasoning while additional branches primarily expand the search space.

Larger $B$ also yields stronger downstream transfer, though with diminishing returns. As shown in Figure~\ref{fig:branch-scaling_b}, mean downstream accuracy (averaged over the eight benchmarks of Section~\ref{sec:downstream}) rises from $52.5\%$ at $B\!=\!2$ to $55.7\%$ at $B\!=\!4$ ($+3.2$ pp), then by only an additional $+0.6$ pp from $B\!=\!4$ to $B\!=\!6$. This contrasts with the depth axis under the same \emph{+~Quantification} setting, where deeper training continues to yield downstream gains throughout our tested range (up to 414 training steps), suggesting that $B$ is a less capacious axis of difficulty than depth. A small number of unprovable candidates is plausibly sufficient to force the model to evaluate each branch, after which additional candidates raise training cost without meaningfully extending the reasoning skills required. We accordingly fix $B=4$ as the default candidate count in our main experiments.

\subsection{Detailed Downstream Results}~\label{app:downstream}

Table~\ref{tab:downstream_results} reports the full downstream results for each benchmark and training setting. In Section~\ref{sec:downstream}, we present the corresponding averaged results across benchmarks for clarity.

\begin{table*}[t]
\centering
\caption{Detailed downstream results for all benchmarks and training settings. Each entry reports \texttt{Avg@8 / Pass@8} in percentage. The last column reports the mean accuracy across the eight benchmarks (i.e., the mean of per-benchmark \texttt{Avg@8} scores).}
\label{tab:downstream_results}
\resizebox{\textwidth}{!}{
\begin{tabular}{rccccccccc}
\toprule
\multirow{2}{*}{$D$} & \multicolumn{6}{c}{Math} & \multicolumn{2}{c}{Science} & \multirow{2}{*}{Avg} \\
\cmidrule(lr){2-7} \cmidrule(lr){8-9}
& Math500 & AMC23 & AIME24 & AIME25 & Minerva & Olympiad & GPQA-Diamond & MMLU-Pro & \\
\midrule

\multicolumn{10}{l}{\textbf{Base model}} \\
-- & 83.93 / 93.60 & 65.00 / 87.50 & 21.67 / 40.00 & 20.00 / 40.00 & 40.21 / 50.74 & 53.32 / 71.36 & 42.55 / 76.26 & 68.45 / 78.60 & 49.39 \\
\midrule

\multicolumn{10}{l}{\textbf{Implication-only}} \\
16 & 84.28 / 94.00 & 64.69 / 87.50 & 22.92 / 43.33 & 22.50 / 43.33 & 40.40 / 49.63 & 53.43 / 71.07 & 42.93 / 76.26 & 67.85 / 87.20 & 49.88 \\
24 & 84.85 / 94.60 & 68.44 / 90.00 & 24.58 / 46.67 & 21.67 / 46.67 & 40.85 / 50.00 & 54.32 / 73.15 & 43.62 / 75.25 & 69.90 / 88.60 & 51.03 \\
32 & 84.95 / 95.60 & 68.44 / 90.00 & 24.17 / 50.00 & 22.50 / 46.67 & 40.63 / 50.00 & 54.78 / 71.81 & 44.32 / 72.73 & 69.05 / 87.80 & 51.11 \\
48 & 84.73 / 94.60 & 67.19 / 87.50 & 21.67 / 40.00 & 20.42 / 43.33 & 40.81 / 51.10 & 55.51 / 72.70 & 43.18 / 77.27 & 69.05 / 88.40 & 50.32 \\
64 & 86.20 / 96.00 & 70.31 / 97.50 & 28.75 / 43.33 & 21.25 / 46.67 & 41.27 / 50.37 & 55.14 / 73.00 & 42.55 / 77.78 & 70.55 / 89.40 & 52.00 \\
96 & 85.53 / 95.20 & 70.31 / 97.50 & 23.75 / 53.33 & 24.17 / 50.00 & 40.40 / 51.84 & 55.17 / 73.00 & 44.32 / 77.78 & 70.03 / 88.80 & 51.71 \\
\midrule

\multicolumn{10}{l}{\textbf{+ Conjunction}} \\
8  & 84.30 / 95.00 & 70.63 / 95.00 & 26.25 / 46.67 & 22.50 / 50.00 & 39.75 / 50.74 & 53.75 / 71.51 & 42.36 / 75.25 & 69.10 / 87.80 & 51.08 \\
12 & 85.53 / 95.40 & 69.69 / 95.00 & 30.00 / 60.00 & 20.00 / 40.00 & 40.40 / 49.63 & 54.67 / 73.74 & 41.60 / 77.27 & 69.55 / 88.80 & 51.43 \\
16 & 85.60 / 95.00 & 71.88 / 95.00 & 24.17 / 43.33 & 24.17 / 56.67 & 40.53 / 51.10 & 55.42 / 73.15 & 43.06 / 78.79 & 70.53 / 88.00 & 51.92 \\
20 & 85.45 / 95.20 & 70.94 / 92.50 & 26.67 / 53.33 & 21.25 / 46.67 & 39.66 / 49.63 & 55.25 / 74.18 & 44.44 / 79.29 & 69.20 / 87.80 & 51.61 \\
24 & 85.83 / 94.80 & 72.81 / 97.50 & 24.17 / 40.00 & 19.58 / 36.67 & 40.76 / 50.37 & 55.84 / 71.81 & 45.52 / 77.27 & 69.68 / 87.60 & 51.77 \\
28 & 86.15 / 95.60 & 70.94 / 97.50 & 27.92 / 53.33 & 22.08 / 43.33 & 40.63 / 50.37 & 55.40 / 74.33 & 42.42 / 76.26 & 70.15 / 88.20 & 51.96 \\
\midrule

\multicolumn{10}{l}{\textbf{+ Negation}} \\
8  & 84.98 / 95.20 & 71.88 / 95.00 & 22.92 / 50.00 & 20.42 / 43.33 & 39.52 / 50.00 & 54.56 / 73.44 & 43.06 / 76.77 & 69.55 / 89.20 & 50.86 \\
12 & 85.83 / 96.00 & 71.56 / 95.00 & 26.25 / 53.33 & 20.42 / 43.33 & 40.53 / 50.74 & 55.77 / 73.44 & 44.00 / 76.26 & 69.25 / 87.40 & 51.70 \\
16 & 86.13 / 95.60 & 72.19 / 90.00 & 29.17 / 50.00 & 20.42 / 36.67 & 40.72 / 52.21 & 56.64 / 73.89 & 43.94 / 75.76 & 70.10 / 89.00 & 52.41 \\
20 & 86.85 / 95.40 & 74.38 / 97.50 & 30.00 / 53.33 & 22.92 / 50.00 & 41.73 / 52.21 & 56.81 / 73.59 & 46.72 / 80.30 & 71.93 / 89.60 & 53.92 \\
24 & 87.43 / 96.40 & 72.81 / 92.50 & 30.00 / 53.33 & 22.50 / 50.00 & 40.72 / 51.47 & 57.07 / 74.63 & 44.07 / 75.76 & 72.63 / 89.60 & 53.40 \\
28 & 87.38 / 97.00 & 73.75 / 92.50 & 27.08 / 46.67 & 20.83 / 43.33 & 40.95 / 51.10 & 57.62 / 76.41 & 45.39 / 78.28 & 72.08 / 87.60 & 53.13 \\
\midrule

\multicolumn{10}{l}{\textbf{+ Disjunction}} \\
6  & 85.48 / 95.40 & 70.31 / 90.00 & 24.58 / 46.67 & 22.92 / 46.67 & 39.80 / 50.37 & 54.95 / 73.15 & 42.30 / 77.27 & 68.63 / 87.80 & 51.12 \\
8  & 86.75 / 96.20 & 70.00 / 95.00 & 25.83 / 50.00 & 25.42 / 53.33 & 41.41 / 50.00 & 56.88 / 73.15 & 46.15 / 78.79 & 70.53 / 88.60 & 52.87 \\
10 & 87.40 / 96.40 & 71.56 / 95.00 & 27.50 / 50.00 & 25.42 / 50.00 & 40.81 / 51.10 & 57.60 / 74.33 & 45.08 / 79.29 & 72.18 / 88.40 & 53.44 \\
12 & 87.20 / 95.60 & 74.38 / 95.00 & 30.00 / 50.00 & 27.50 / 53.33 & 41.27 / 52.21 & 57.86 / 73.89 & 44.07 / 77.78 & 72.00 / 89.60 & 54.29 \\
14 & 87.78 / 96.00 & 76.56 / 95.00 & 33.33 / 63.33 & 25.83 / 50.00 & 41.27 / 51.84 & 58.88 / 76.41 & 45.27 / 80.30 & 72.85 / 89.60 & 55.22 \\
20 & 88.63 / 96.60 & 76.25 / 95.00 & 35.00 / 63.33 & 27.92 / 56.67 & 41.82 / 51.84 & 59.20 / 75.82 & 46.53 / 81.31 & 74.08 / 89.40 & 56.18 \\
\midrule

\multicolumn{10}{l}{\textbf{+ Quantification}} \\
4  & 85.43 / 95.40 & 69.38 / 92.50 & 24.58 / 46.67 & 23.33 / 50.00 & 40.17 / 50.37 & 55.06 / 73.15 & 41.86 / 77.78 & 69.47 / 88.20 & 51.16 \\
6  & 86.75 / 96.80 & 70.00 / 90.00 & 28.33 / 50.00 & 21.67 / 40.00 & 40.67 / 51.47 & 56.82 / 74.78 & 45.79 / 75.25 & 71.53 / 88.60 & 52.69 \\
8  & 88.15 / 96.40 & 78.12 / 97.50 & 35.00 / 56.67 & 25.00 / 53.33 & 41.68 / 51.84 & 58.90 / 75.22 & 46.34 / 76.77 & 72.53 / 89.80 & 55.72 \\
10 & 88.60 / 97.20 & 75.94 / 97.50 & 34.58 / 60.00 & 25.83 / 53.33 & 42.33 / 51.84 & 60.05 / 76.11 & 45.27 / 76.77 & 74.73 / 89.80 & 55.92 \\
12 & 88.70 / 96.60 & 80.90 / 97.50 & 35.00 / 60.00 & 27.92 / 46.67 & 42.42 / 51.84 & 61.13 / 77.45 & 49.12 / 75.76 & 74.73 / 89.20 & 57.49 \\
14 & 91.48 / 98.00 & 80.63 / 97.50 & 43.75 / 66.67 & 31.25 / 53.33 & 43.70 / 52.57 & 64.39 / 79.08 & 48.17 / 72.22 & 77.00 / 89.80 & 60.05 \\

\bottomrule
\end{tabular}
}
\end{table*}

\subsection{Curriculum at the Most Expressive Setting}\label{app:curriculum_quant}

The training-distribution comparison in Section~\ref{sec:distribution} is conducted under the \emph{+~Conjunction} setting. To verify that the benefit of curriculum training is not specific to this setting, we replicate the uniform vs.\ curriculum comparison at the most expressive \emph{+~Quantification} setting.

\begin{figure}[h]
\centering
\includegraphics[width=0.55\textwidth]{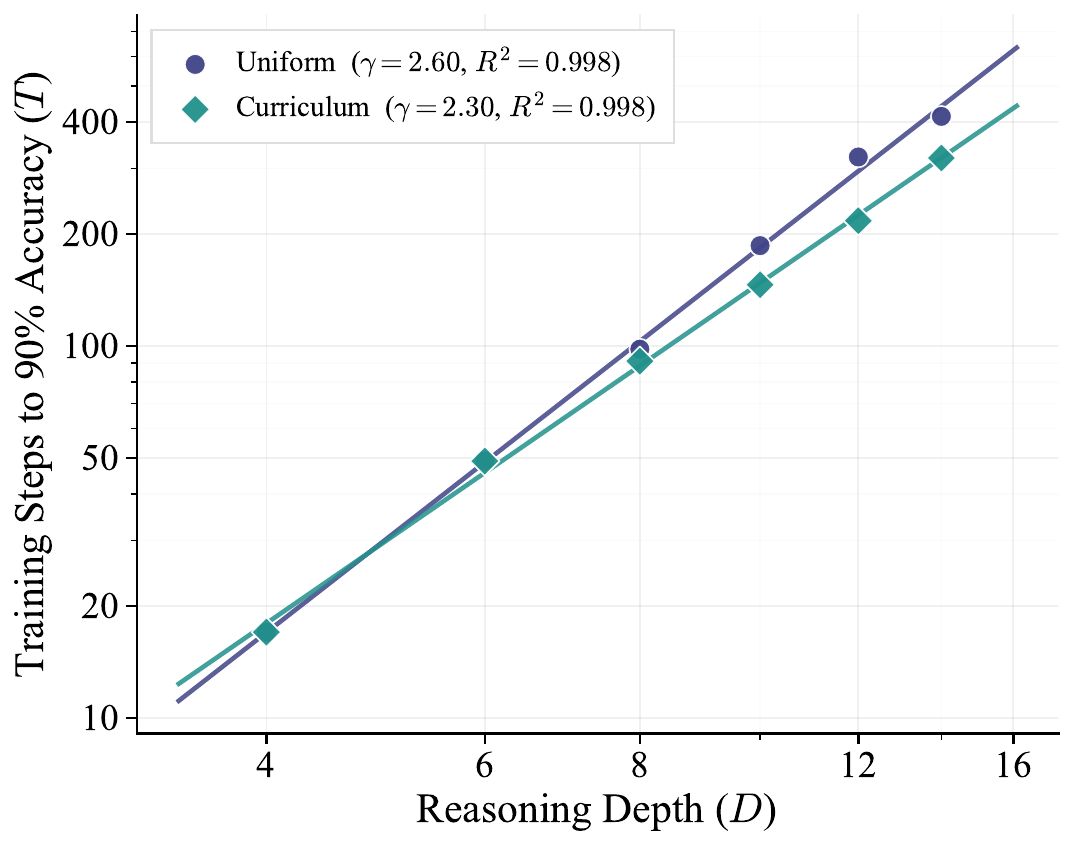}
\caption{Uniform vs.\ curriculum training under the most expressive \emph{+~Quantification} setting. Curriculum training reduces the scaling exponent from $\gamma = 2.60$ to $\gamma = 2.30$ ($R^{2} = 0.998$ in both cases), confirming that the benefit observed at \emph{+~Conjunction} (Section~\ref{sec:distribution}) extends to the most expressive logic in our hierarchy.}
\label{fig:distribution-full}
\end{figure}

As shown in Figure~\ref{fig:distribution-full}, curriculum training continues to lower the scaling exponent at \emph{+~Quantification}, from $\gamma = 2.60$ to $\gamma = 2.30$, with $R^{2} = 0.998$ in both cases. The reduction in exponent ($\Delta\gamma = -0.30$) is comparable to that observed at \emph{+~Conjunction} ($\Delta\gamma = -0.37$, from $1.70$ to $1.33$), and the qualitative effect is preserved: curriculum training yields sample-efficiency gains that compound with proof depth. We conclude that the benefit of curriculum training generalizes across the expressiveness hierarchy.

\subsection{Data Distribution Shapes Long-CoT Emergence}
\label{app:curriculum_res}

The main text shows that the training distribution substantially affects scaling efficiency in the \emph{+~Conjunction} setting. Here, we provide the full training trajectories behind this comparison and analyze how the three data distributions shape models' behavior. We focus on response length, actor entropy, and validation accuracy, which together reveal when longer reasoning traces emerge and how this transition relates to the observed scaling behavior.

\begin{figure}[t]
    \centering

    \begin{subfigure}[b]{\linewidth}
        \centering
        \includegraphics[width=\linewidth]{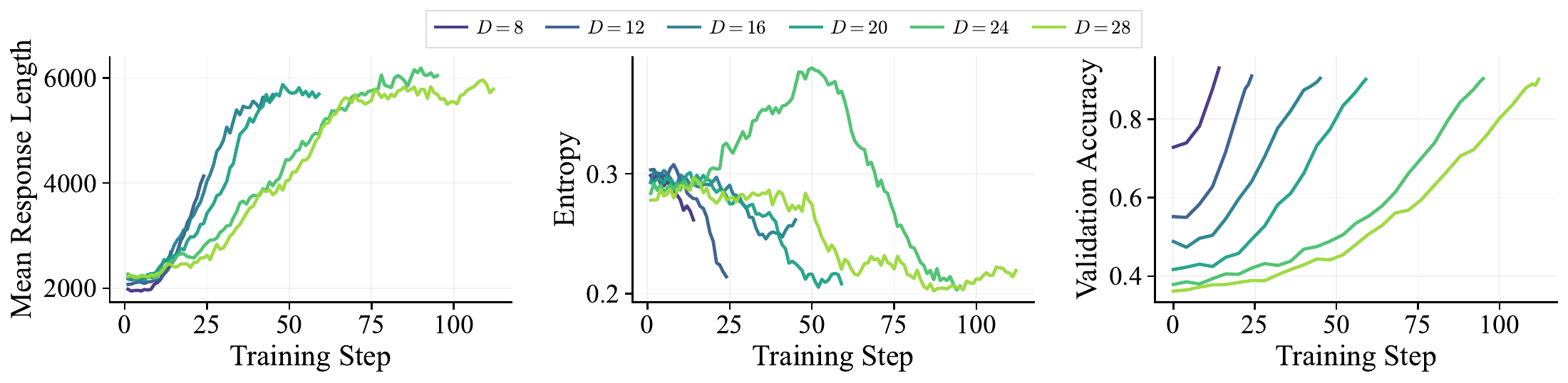}
        \caption{Uniform.}
    \end{subfigure}

    \vspace{0.4em}

    \begin{subfigure}[b]{\linewidth}
        \centering
        \includegraphics[width=\linewidth]{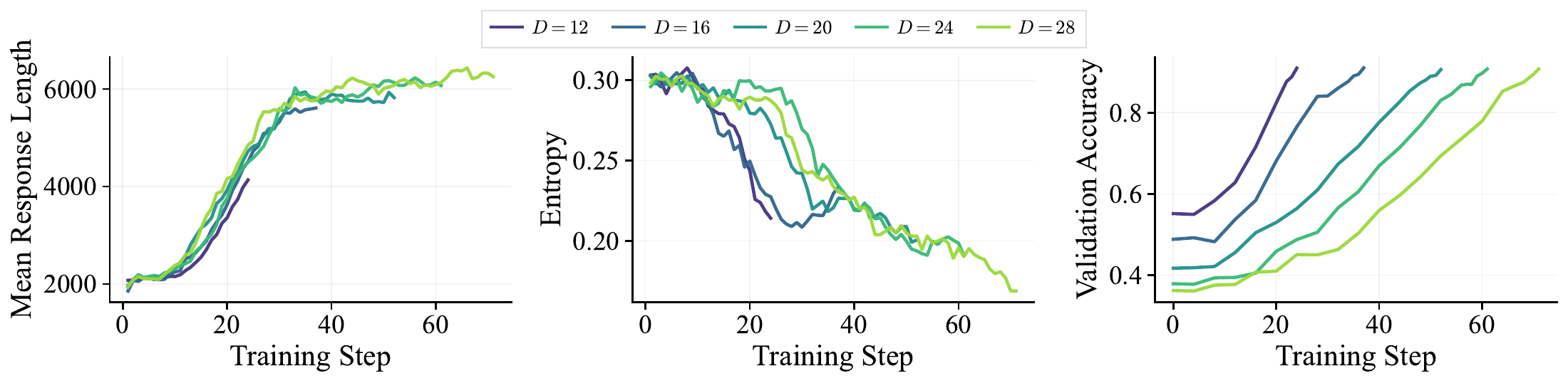}
        \caption{Curriculum.}
    \end{subfigure}

    \vspace{0.4em}

    \begin{subfigure}[b]{\linewidth}
        \centering
        \includegraphics[width=\linewidth]{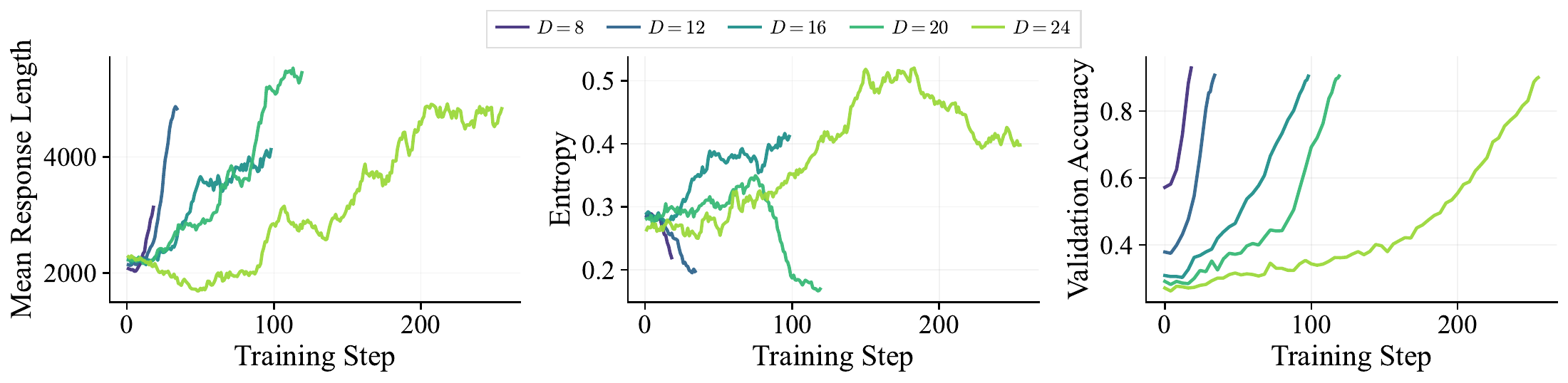}
        \caption{Difficult-only.}
    \end{subfigure}

    \caption{Training trajectories on \emph{+~Conjunction} under three data distributions.
    Each row: response length (left), entropy (middle), validation accuracy (right);
    legend (training depth $D$) is shared across the three sub-panels. Note that axis ranges may differ across rows to improve readability.}
    \label{fig:curriculum_response_length}
\end{figure}

Figure~\ref{fig:curriculum_response_length} shows that curriculum training substantially eliminates the delay in long-CoT emergence across depths. Under the curriculum regime, runs at different target depths enter the long-CoT regime within a relatively narrow range of training steps; this length expansion is accompanied by a sharper entropy drop and earlier improvement in validation accuracy. By contrast, under the uniform distribution, long-CoT emergence becomes increasingly delayed as the target depth grows, with a particularly large gap between $D=20$ and $D=24$; the corresponding entropy decrease and accuracy rise are likewise shifted to much later training steps. The difficult-only regime exhibits a similar but more pronounced pattern, especially at larger depths, where response length remains short and entropy stays elevated for a longer warm-up period before accuracy begins to improve.

Together, these dynamics suggest that the distribution effect in the main text is closely tied to when long-CoT behavior emerges. Curriculum training appears to provide a smoother path into longer reasoning: the policy first learns useful reasoning patterns on shallower instances and can then extend them as deeper instances are introduced. By contrast, uniform and difficult-only training tend to remain longer in a short-response, high-entropy regime before transitioning to long-CoT behavior. This delayed transition is consistent with their larger scaling exponents, whereas the earlier and more synchronized transition under curriculum training helps explain its improved scaling efficiency.

\subsection{Cross-Scale and Cross-Model Replication}\label{app:qwen3-8b}

\begin{figure}[t]
    \centering
    \begin{subfigure}[b]{0.49\textwidth}
        \centering
        \includegraphics[width=\textwidth]{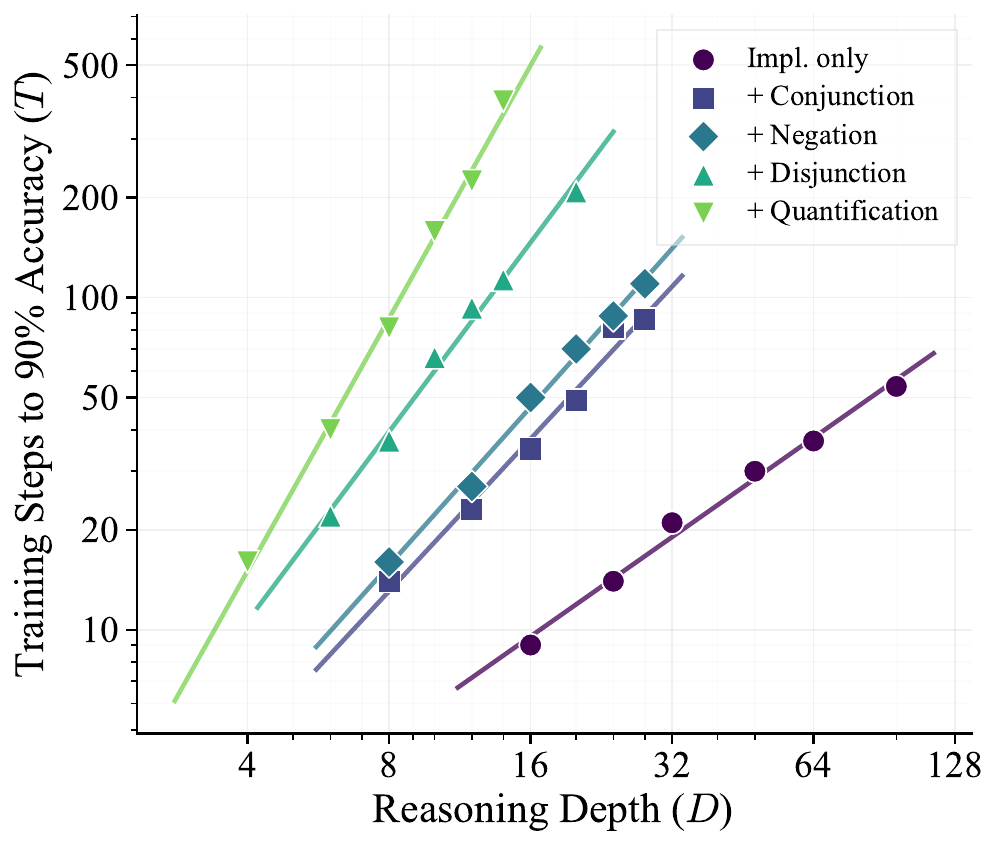}
        \caption{Log-log scaling curves with power-law fits.}
        \label{fig:scaling-loglog-8B}
    \end{subfigure}
    \hfill
    \begin{subfigure}[b]{0.49\textwidth}
        \centering
        \includegraphics[width=\textwidth]{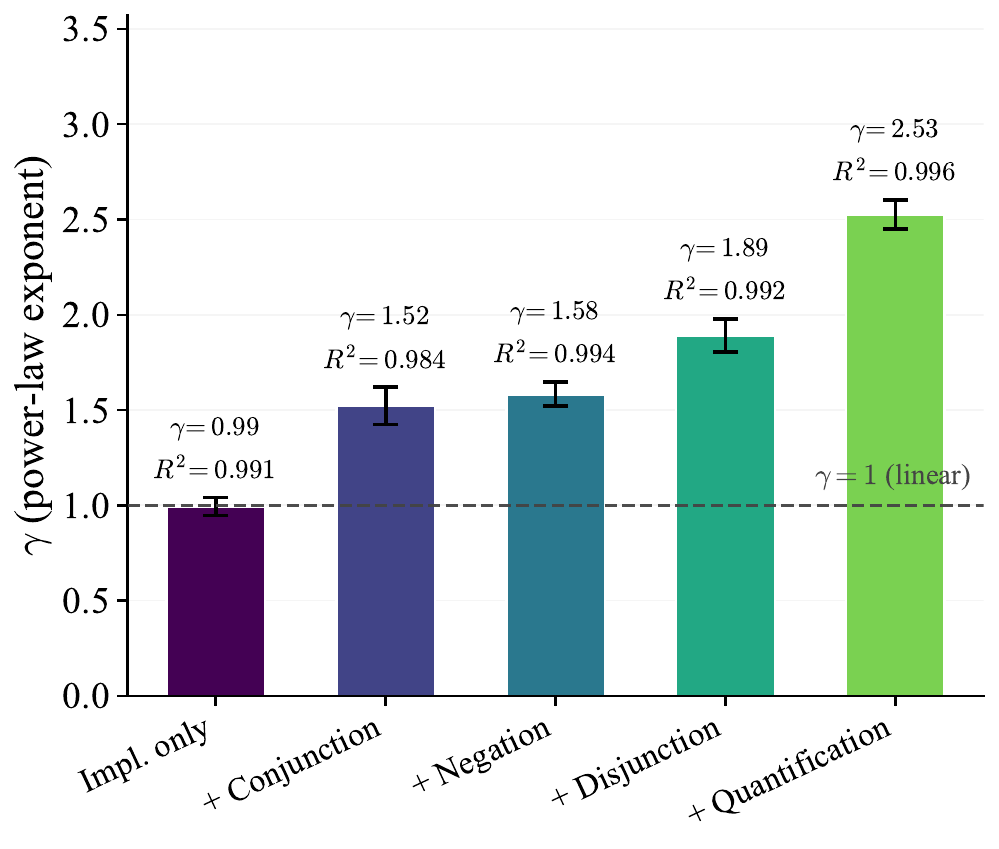}
        \caption{Fitted exponent $\gamma$ vs. expressiveness of the logic.}
        \label{fig:gamma-bar-8B}
    \end{subfigure}
    \caption{Cross-scale replication on Qwen3-8B. \textbf{(a)} Training steps to convergence vs.\ reasoning depth across five expressiveness levels on log-log axes. Solid lines show power-law fits $T \propto D^{\gamma}$. \textbf{(b)} Fitted $\gamma$ increases monotonically when expressiveness increases from \textit{Implication-only} ($\gamma\!=\!0.99$) to the \textit{+~Quantification} setting ($\gamma\!=\!2.53$), mirroring the 4B picture in Figure~\ref{fig:scaling-power-law} but with systematically smaller exponents. Error bars denote $\pm 1$ standard error of the fitted exponent.}
    \label{fig:scaling-power-law-8B}
\end{figure}

\paragraph{Cross-scale replication.} To assess whether the scaling behavior reported in the main text generalizes beyond the 4B scale, we replicate the central scaling experiment of Section~\ref{sec:scaling-depth} on Qwen3-8B. Figure~\ref{fig:scaling-power-law-8B} shows the log-log scaling curves and the fitted exponents.
The scaling behavior at 8B closely mirrors that at 4B. All five expressiveness settings continue to follow a clean power law in proof depth, with $R^{2} \geq 0.98$. The fitted exponent $\gamma$ remains monotonically increasing with expressiveness, ranging from $0.99$ for \textit{Implication-only} to $2.53$ for \textit{+~Quantification}, and the small gap between \textit{+~Conjunction} ($\gamma = 1.52$) and \textit{+~Negation} ($\gamma = 1.58$) is also preserved.
Compared with the 4B model, the 8B exponents are systematically smaller across all settings (e.g., $1.52$ vs.\ $1.67$ under \textit{+~Conjunction}), while the relative ordering remains unchanged. This suggests that the qualitative scaling pattern is preserved across model scales.

As an additional downstream check, the Qwen3-8B model trained at the deepest \emph{+~Quantification} setting ($D=14$) improves the eight-benchmark average from $50.93\%$ to $60.53\%$, a $+9.60$ percentage-point gain, suggesting a transfer trend similar to that observed for Qwen3-4B.

\begin{figure}[t]
    \centering
    \includegraphics[width=0.55\linewidth]{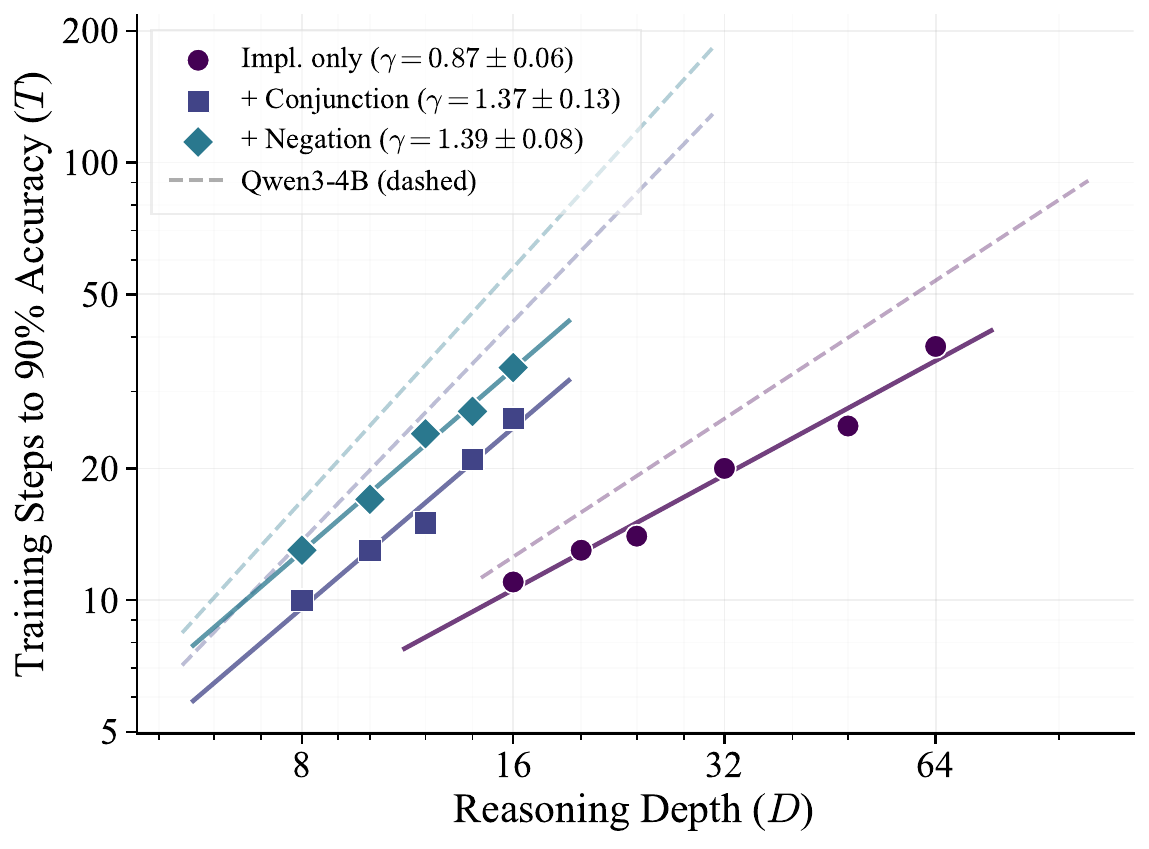}
    \caption{
    Partial cross-model replication on Ministral3-3B-Instruct-2512.
    Solid lines show power-law fits for three representative expressiveness settings; dashed lines show the corresponding Qwen3-4B fits for reference.
    The fitted exponent increases monotonically from \emph{Implication-only} ($\gamma = 0.87$) to \emph{+~Conjunction} ($\gamma = 1.37$) and \emph{+~Negation} ($\gamma = 1.39$), preserving the qualitative ordering across model families.
    }
    \label{fig:ministral}
\end{figure}

\paragraph{Cross-model replication.} We further conduct a partial replication on the non-Qwen-family Ministral3-3B-Instruct-2512 to examine whether the observed exponent ordering is specific to the Qwen family. Due to compute constraints, we evaluate three representative settings: \emph{Implication-only}, \emph{+~Conjunction}, and \emph{+~Negation}. As shown in Figure~\ref{fig:ministral}, all three settings exhibit clear power-law scaling, and the fitted exponent again increases monotonically as the training setting becomes more expressive, from $0.87\pm0.06$ to $1.37\pm0.13$ and $1.39\pm0.08$, respectively. This provides complementary evidence that the qualitative exponent ordering is not unique to the Qwen family. 

Together, the Qwen3-8B and Ministral3-3B results provide complementary evidence that the observed exponent hierarchy persists across both model scale and model family.

\subsection{Matched-Budget Alternative-Training Controls}\label{app:alt_training}

The downstream gains in Section~\ref{sec:downstream} could in principle arise from factors other than learning to solve the underlying reasoning task. For example, they might arise from mere exposure to the synthetic reasoning data, or from generic RL-induced parameter drift that requires no correctness signal.
We test these two controls under a matched training budget: supervised fine-tuning (SFT) on gold proof traces, and RL with correctness-independent rewards.
Both controls use the \textit{+~Quantification}, $D=8$ setting, where the RL run reaches the target accuracy in $T=98$ steps.

\paragraph{Setup.}
(i)~\emph{SFT control.} We train Qwen3-4B on the same prompts using Z3-verified gold proof traces, obtained via breadth-first search and rendered in natural language. Each response contains the complete derivation and ends with the same answer format used in RL. We use a learning rate of $10^{-5}$ with cosine decay and match the RL mini-batch size.
(ii)~\emph{Correctness-independent RL controls.} We use the same training data, model, and DAPO recipe as in the original RL run, replacing only the correctness reward with a format-only or random reward. The format-only variant assigns reward for agreement with the required answer format regardless of whether the selected answer is correct, whereas the random-reward variant assigns rewards independently of answer correctness.
(iii)~\emph{Evaluation.} All checkpoints are evaluated with the protocol of Section~\ref{sec:setup}: Avg@8 on the eight downstream benchmarks, together with accuracy on the held-out \textsc{ScaleLogic} test set.
(iv)~\emph{Budget matching.} An RL checkpoint at $T$ steps performs gradient updates over $T \times 256$ prompts, each with $8$ rollouts, corresponding to $T \times 256 \times 8$ training sequences (approximately $200$k sequences at $T = 98$). We match the controls by the total number of training sequences: for SFT, the $100$k gold traces correspond to two epochs; for the correctness-independent RL controls, which use the same RL recipe, we compare models at the same training step $T$.

\begin{table}[t]
\centering
\small
\setlength{\tabcolsep}{3pt}
\caption{Matched-budget alternative-training controls (\textit{+~Quantification}, $D{=}8$; Avg@8). $\Delta$ is the change in the eight-benchmark average relative to the base model; ``In-dist'' is the held-out \textsc{ScaleLogic} test set.}
\label{tab:alt-training}
\resizebox{\textwidth}{!}{
\begin{tabular}{@{}lrrrrrrrrrrr@{}}
\toprule
Model & Math500 & AMC23 & AIME24 & AIME25 & Minerva & Olympiad & GPQA-D & MMLU-Pro & Avg & $\Delta$ & In-dist \\
\midrule
Qwen3-4B (base)            & 83.93 & 65.00 & 21.67 & 20.00 & 40.21 & 53.32 & 42.55 & 68.45 & 49.39 & ---        & 34.35 \\
SFT (1 ep.) & 60.22 & 29.69 &  2.08 &  1.67 & 30.28 & 25.09 & 10.35 & 40.92 & 25.04 & $-$24.35 & 89.84 \\
SFT (2 ep.) & 57.65 & 28.75 &  2.50 &  0.42 & 28.58 & 24.04 &  9.22 & 41.50 & 24.08 & $-$25.31 & \textbf{97.30} \\
Format-only RL       & 83.70 & 64.38 & 21.25 & 22.50 & 39.94 & 51.93 & 44.07 & 67.10 & 49.36 & $-$0.03  & 37.43 \\
Random-reward RL       & 83.45 & 60.94 & 20.83 & 17.50 & 38.92 & 52.24 & 43.62 & 66.97 & 48.06 & $-$1.33  & 32.17 \\
RL (ours)       & \textbf{88.15} & \textbf{78.12} & \textbf{35.00} & \textbf{25.00} & \textbf{41.68} & \textbf{58.90} & \textbf{46.34} & \textbf{72.53} & \textbf{55.72} & \textbf{$+$6.33}  & 90.30 \\
\bottomrule
\end{tabular}
}
\end{table}

\paragraph{Results.}
As shown in Table~\ref{tab:alt-training}, neither alternative-training control reproduces the downstream transfer of our model.

\emph{SFT successfully learns the in-distribution task but fails to transfer.} At the matched budget, SFT reaches 97.30\% on the held-out \textsc{ScaleLogic} test set, exceeding the 90.30\% achieved by RL. Nevertheless, its average downstream performance falls from 49.39 for the base model to 24.08 (-25.31~pp). This collapse is already present after one epoch: despite reaching 89.84\% in-distribution accuracy, SFT scores only 25.04 downstream (-24.35~pp). The lack of transfer therefore cannot be attributed to insufficient SFT convergence or prolonged SFT training. This result indicates that SFT induces severe task specialization and forgetting of general reasoning capabilities as observed in prior work~\citep{lobo2025impact}.

\emph{Correctness-independent RL also fails to transfer.} With the same data, model, optimizer, and training budget as our RL recipe, format-only and random rewards yield downstream averages of 49.36 (-0.03~pp) and 48.06 (-1.33~pp), respectively; both show essentially no improvement over the base model. Their in-distribution accuracies also remain near the base level ($37.43\%$ and $32.17\%$, versus $34.35\%$). Moreover, both controls shorten the model's responses: the average length decreases from $1.7$k tokens initially to $1.2$k under the random reward and $0.9$k under the format-only reward. In contrast, correctness-guided RL increases the response length to $5.6$k tokens by the end of training, consistent with the acquisition of more extensive reasoning.

Together, these controls show that, among the matched alternatives we test, only correctness-guided RL reproduces the downstream gains.

\subsection{Proof-Faithfulness Audit}\label{app:faithfulness}

Our training objective uses only a final-answer exact-match reward (Section~3.3), without supervising intermediate reasoning steps. Strong outcome-level performance therefore does not by itself establish whether the generated reasoning is logically valid or whether the final answer is supported by the preceding trace. We thus conduct a direct proof-faithfulness audit.

For the \textit{+~Quantification}, $D{=}8$ checkpoint, we sample four rollouts on each of $500$ fresh instances and audit every rollout with a correct final answer. A conservative rule-based parser extracts atomic assertions of the form ``[entity] is (not) [attribute]'', excluding hypothetical statements (markers such as ``suppose'' or ``might''); a symbolic theorem prover then checks every extracted assertion for entailment against the ground truth. As a calibration reference, we apply the same audit to the SFT model of Appendix~\ref{app:alt_training}, which is trained directly on gold proof traces.

\begin{table}[t]
\centering
\caption{Proof-faithfulness audit (\textit{+~Quantification}, $D{=}8$): 500 instances $\times$ 4 rollouts per model.
The audit is conditioned on rollouts with correct final answers. A symbolic theorem prover checks every extracted asserted step against the ground truth.
}
\label{tab:faithfulness}
\begin{tabular}{@{}lrr@{}}
\toprule
& RL & SFT \\
\midrule
Correct rollouts audited                       & 1768 / 2000 & 1941 / 2000 \\
Asserted steps per rollout (mean)              & 50.5    & 6.6     \\
Asserted steps that are entailed truths        & 94.6\%      & 98.2\%      \\
\midrule
Final answer explicitly derived   & 96.5\% & 97.7\%  \\
Final answer supported by prior assertions & 89.4\%      & 97.7\%      \\
\bottomrule
\end{tabular}
\end{table}

As shown in Table~\ref{tab:faithfulness}, across 1,768 correct RL rollouts, the model produces approximately 89k asserted steps (50.5 per rollout), of which $94.6\%$ are logically entailed truths. Moreover, $96.5\%$ of correct rollouts explicitly derive the selected final answer, and in $89.4\%$ the final answer is entailed by previously established assertions. The remaining $10.6\%$ that fail our strict grounding criterion should be viewed as a conservative upper bound, since valid elimination arguments and reasoning with skipped intermediate steps can also be counted as ungrounded by this validation methodology. The SFT model, trained directly on gold proof traces, achieves $98.2\%$ entailed assertions and $97.7\%$ on both explicit final-answer derivation and grounding, as expected under direct reasoning supervision. These results show that, despite receiving only an outcome reward, the RL-trained model produces predominantly valid and answer-supporting reasoning traces.

\subsection{Evaluation on External Logical-Reasoning Benchmarks}\label{app:external_logic}

To evaluate whether the logical reasoning improvements induced by training on \textsc{ScaleLogic} transfer beyond the synthetic training distribution, we evaluate the \textit{+~Quantification}, $D=8$ model on two established logical-reasoning benchmarks: the harder Logical Deduction-5/7 subsets of BBH~\citep{suzgun2023challenging}, and 500 randomly sampled ZebraLogic~\citep{lin2025zebralogic} test examples in multiple-choice mode. All checkpoints are evaluated using the same protocol as in Section~\ref{sec:setup}, and we report Avg@8.

\begin{table}[t]
\centering
\caption{Performance on external logical-reasoning benchmarks (Avg@8).
SFT and correctness-independent RL are the matched-budget controls from Appendix~\ref{app:alt_training}.}
\label{tab:external-logic}
\begin{tabular}{@{}lrr@{}}
\toprule
Model & BBH & ZebraLogic \\
\midrule
Qwen3-4B (base)      & 84.55 & 64.70 \\
SFT                  & 35.85 & 4.12  \\
Format-only RL & 85.15 & 62.48 \\
Random-reward RL & 84.02 & 61.52 \\
RL (ours)            & \textbf{93.35} & \textbf{75.10} \\
\bottomrule
\end{tabular}
\end{table}

As shown in Table~\ref{tab:external-logic}, our model substantially outperforms the base model on both benchmarks, improving Avg@8 from 84.55 to 93.35 on BBH and from 64.70 to 75.10 on ZebraLogic. In contrast, SFT on the same data causes severe catastrophic forgetting, reducing performance to 35.85 and 4.12, respectively, while the matched-budget correctness-independent RL controls (Appendix~\ref{app:alt_training}) produce only small and inconsistent changes relative to the base model (85.15/62.48 with the format reward and 84.02/61.52 with the random reward). Overall, this demonstrates the good
generalization capability of our model on out-of-distribution logical reasoning tasks, providing additional evidence that the gains extend beyond the \textsc{ScaleLogic} training distribution.

\subsection{Comparison with an Alternative Synthetic-Data Baseline}\label{app:data_baseline}

We further compare \textsc{ScaleLogic} against SATURN~\citep{liu2025saturn}, a synthetic-data framework for Boolean satisfiability (SAT) reasoning. SATURN trains language models on synthetic 3-SAT instances with verifiable rewards, where task difficulty is controlled by the number of variables and clauses. To enable a controlled comparison, we use the same Qwen3-4B model and DAPO training recipe as in our main experiments. Starting from $(k,l)=(5,15)$, the configuration used in the original work, we keep the clause-to-variable ratio fixed at $l/k=3$ and progressively increase the problem size to $(8,24)$, $(10,30)$, and $(12,36)$.

\begin{table}[t]
\centering
\footnotesize
\setlength{\tabcolsep}{3pt}
\caption{Downstream comparison between \textsc{ScaleLogic} and SATURN 3-SAT training (Avg@8). $T$ is the number of training steps to reach $90\%$ validation accuracy on the corresponding task.}
\label{tab:sat-baseline}
\resizebox{\textwidth}{!}{
\begin{tabular}{@{}lc rrrrrr rr c@{}}
\toprule
Setting & $T$ & Math500 & AMC23 & AIME24 & AIME25 & Minerva & Olympiad & GPQA-D & MMLU-Pro & Avg \\
\midrule
Qwen3-4B (Base) & --- & 83.93 & 65.00 & 21.67 & 20.00 & 40.21 & 53.32 & 42.55 & 68.45 & 49.39 \\
\midrule
3-SAT ($k{=}5$, $l{=}15$)  & 19  & 85.58 & 67.19 & 24.17 & 20.00 & 40.95 & 55.29 & 42.23 & 68.85 & 50.53 \\
3-SAT ($k{=}8$, $l{=}24$)  & 52  & 86.33 & 75.31 & 28.33 & 22.50 & 41.22 & 57.53 & 41.67 & 69.60 & 52.81 \\
3-SAT ($k{=}10$, $l{=}30$)    & 66  & 86.67 & 76.25 & 30.00 & 24.17 & 41.50 & 58.10 & 45.27 & 71.28 & 54.15 \\
3-SAT ($k{=}12$, $l{=}36$)    & 105 & 87.10 & 77.20 & 30.80 & 22.90 & 40.80 & 58.40 & 43.80 & 71.80 & 54.10 \\
\midrule
Ours (+~Quantification, $D{=}8$)  & 98  & \textbf{88.15} & \textbf{78.12} & \textbf{35.00} & \textbf{25.00} & \textbf{41.68} & \textbf{58.90} & \textbf{46.34} & \textbf{72.53} & \textbf{55.72} \\
\bottomrule
\end{tabular}
}
\end{table}

As shown in Table~\ref{tab:sat-baseline}, increasing the SAT problem size initially improves downstream transfer, but the gains plateau at $(k, l) = (10, 30)$, where the average score reaches 54.15; further scaling to $(12, 36)$ slightly decreases the score to 54.10. Moreover, the $(12, 36)$ setting requires more training updates than our \textit{+~Quantification}, $D = 8$ setting to reach $90\%$ validation accuracy (105 vs.\ 98), while still yielding lower downstream performance (54.10 vs.\ 55.72). Compared with the rapid saturation observed for SAT, our framework continues to benefit from increased task difficulty: raising the depth from $D = 8$ to $D = 14$ further improves the average score from 55.72 to 60.05 (Section~\ref{sec:downstream}). This controlled comparison suggests that, under the matched model and RL recipe, \textsc{ScaleLogic} provides stronger downstream transfer and supports more effective scaling with task difficulty.

\section{Training Dynamics of the Main Experiments}\label{app:dynamics}

To accompany the training-step scaling results reported in the main text, we visualize the training trajectories of the five main experimental settings. For each setting, we plot three quantities as a function of the training step: (i) the mean response length, (ii) the actor entropy, and (iii) the validation accuracy (the same quantity used to define the training compute $T$ in the main paper).

\begin{figure}[p]
    \centering

    \begin{subfigure}[b]{\linewidth}
        \centering
        \includegraphics[width=\linewidth]{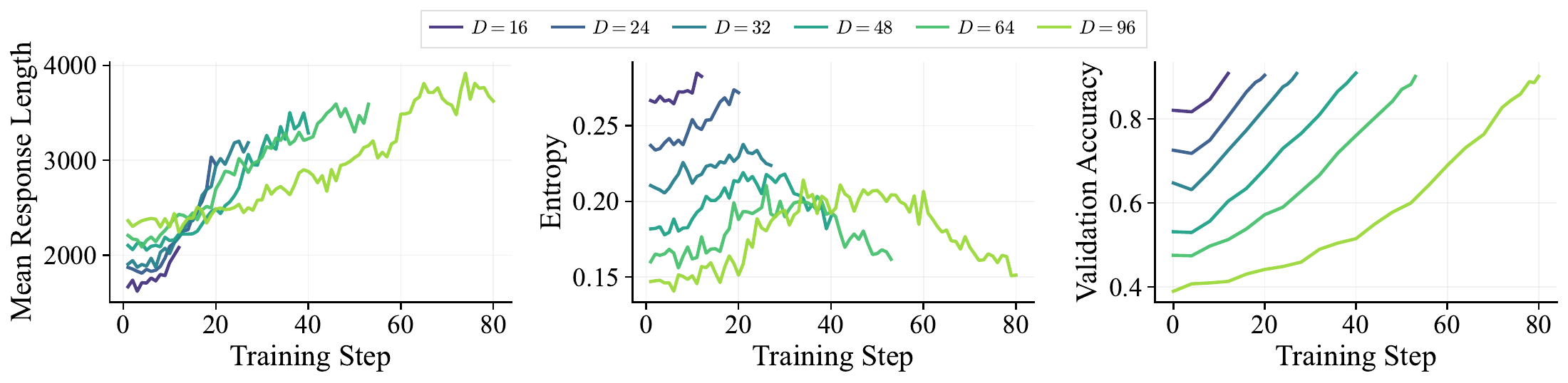}
        \caption{Impl.\ only}
        \label{fig:impl_only_trajectories}
    \end{subfigure}

    \vspace{0.4em}

    \begin{subfigure}[b]{\linewidth}
        \centering
        \includegraphics[width=\linewidth]{{figures/fig_Conjunction_panel}.pdf}
        \caption{+ Conjunction}
        \label{fig:conjunction_trajectories}
    \end{subfigure}

    \vspace{0.4em}

    \begin{subfigure}[b]{\linewidth}
        \centering
        \includegraphics[width=\linewidth]{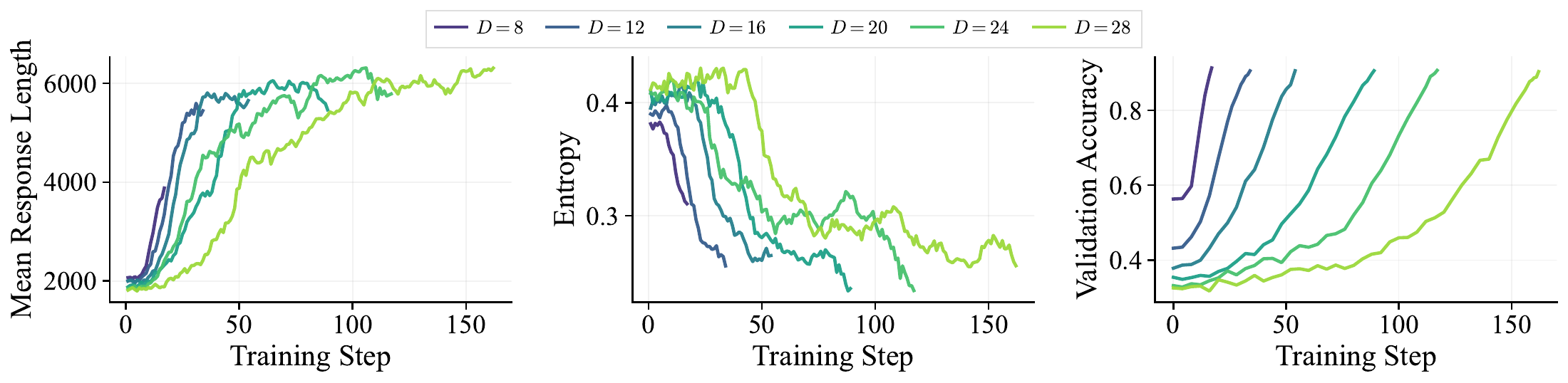}
        \caption{+ Negation}
        \label{fig:negation_trajectories}
    \end{subfigure}

    \vspace{0.4em}

    \begin{subfigure}[b]{\linewidth}
        \centering
        \includegraphics[width=\linewidth]{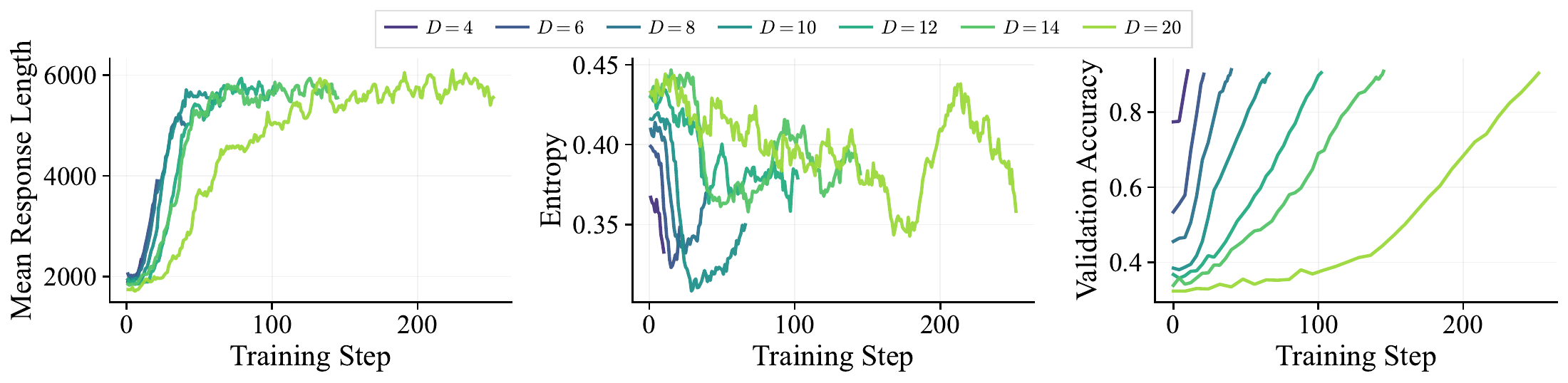}
        \caption{+ Disjunction}
        \label{fig:disjunction_trajectories}
    \end{subfigure}

    \vspace{0.4em}

    \begin{subfigure}[b]{\linewidth}
        \centering
        \includegraphics[width=\linewidth]{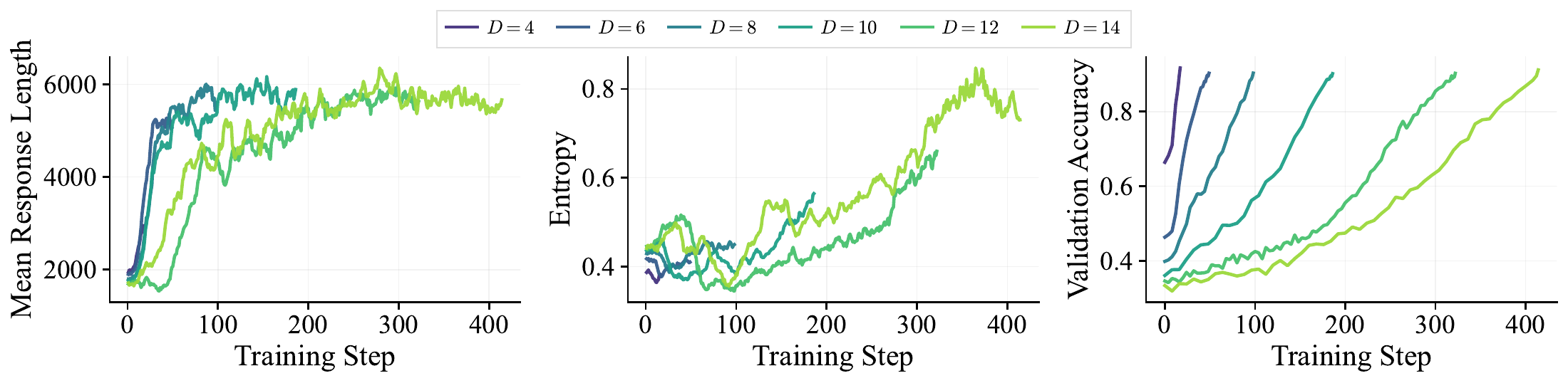}
        \caption{+ Quantification}
        \label{fig:full_trajectories}
    \end{subfigure}

    \caption{%
        Training trajectories for all five settings.
        Each row: response length (left), entropy (middle), validation accuracy (right);
        legend (training depth $D$) is shared across the three panels. Note that axis ranges may differ across rows to improve readability.
    }
    \label{fig:all_trajectories}
\end{figure}

Across the five settings, Figure~\ref{fig:all_trajectories} reveals a consistent set of training dynamics. First, validation accuracy eventually reaches the $0.9$ threshold for all trained depths, but deeper problems require systematically more training steps to reach this threshold. These data points are exactly the quantities used to construct the scaling curves in the main text. Second, response length tends to grow during training, especially at larger depths and in more expressive settings, indicating that RL often induces longer reasoning traces as the task requires deeper proof search. Third, actor entropy generally decreases as training progresses, suggesting that the policy gradually concentrates on more reliable reasoning patterns; on harder configurations, this decrease is sometimes preceded by a longer exploratory phase. Together, these trajectories show that the measured scaling behavior is not an artifact of isolated convergence points, but reflects broader changes in accuracy, reasoning length, and policy uncertainty throughout training.

\section{Evaluation of Frontier LLMs on \textsc{ScaleLogic}}\label{app:strong_llm}

\begin{figure}[t]
    \centering
    \includegraphics[width=0.7\linewidth]{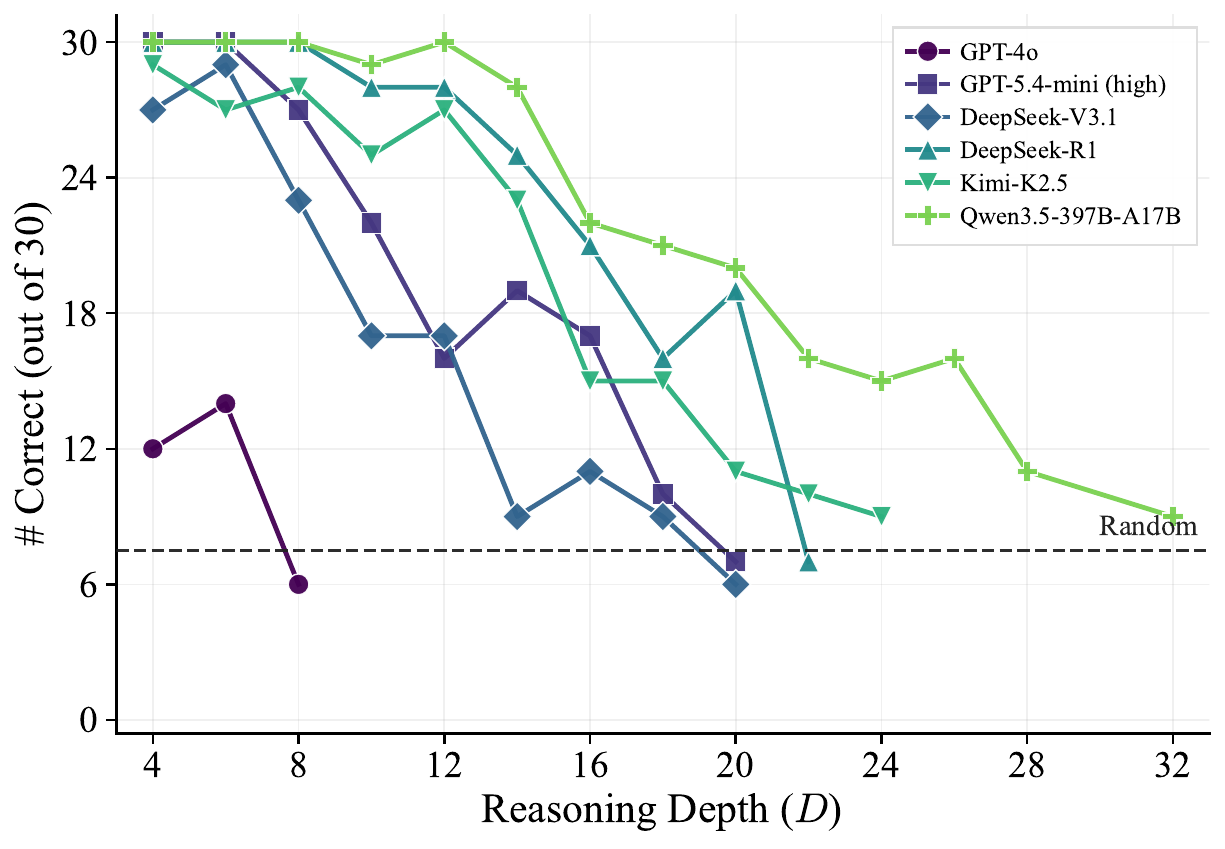}
    \caption{Number of correct answers out of 30 for frontier LLMs on \textsc{ScaleLogic} (\textit{+~Quantification} setting, $B\!=\!4$) as a function of reasoning depth. All models degrade with increasing depth.}
    \label{fig:strong_llm}
\end{figure}

To verify that our synthetic tasks present a genuine reasoning challenge even for frontier models, we evaluate six strong LLMs in the \textit{+~Quantification} setting with $B=4$ across reasoning depths $D \in \{4,6,\dots,32\}$. For each depth, we sample 30 tasks and report the number of correct answers. The evaluated models include both non-thinking LLMs (GPT-4o~\citep{hurst2024gpt} and DeepSeek-V3.1~\citep{liu2024deepseek}) and LRMs (GPT-5.4-mini~\citep{singh2025openai}, DeepSeek-R1~\citep{guo2025deepseek}, Kimi-K2.5~\citep{team2026kimi}, and Qwen3.5-397B-A17B~\citep{qwen35blog}). We use greedy decoding with temperature $T=0$ whenever temperature control is supported. For GPT-5.4-mini, whose temperature cannot be modified, we use its default setting and set the reasoning effort to ``high''. To minimize failures due to truncated outputs, we set the maximum response length to 65,536 tokens.

As shown in Figure~\ref{fig:strong_llm}, increasing reasoning depth leads to clear performance degradation across all evaluated models. Non-reasoning LLMs fail rapidly: GPT-4o and DeepSeek-V3.1 reach near-random accuracy by $D=8$ and $D=20$. Although LRMs are more robust, with Qwen3.5-397B-A17B remaining above random even at $D=32$, they also exhibit marked degradation beyond $D=12$. These results suggest that the most expressive tasks of \textsc{ScaleLogic} remain challenging even for frontier reasoning models.

Notably, Figure~\ref{fig:ood} shows that targeted RL training on \textsc{ScaleLogic} substantially closes the in-distribution gap between a 4B model and much larger frontier models evaluated zero-shot. At $D=28$, the RL-trained 4B model maintains 40\% accuracy, matched only by Qwen3.5-397B-A17B among the evaluated frontier models. Thus, the comparison highlights distribution-specific gains from targeted RL, while the continued degradation of all models at larger depths confirms that the depth axis captures a meaningful reasoning challenge.

\section{Qualitative Examples}
\label{sec:qualitative-examples}


Beyond the aggregate accuracy gains in Section~\ref{sec:downstream}, Figure~\ref{fig:qualitative_example} illustrates how RL post-training changes the model's reasoning behavior. We compare our model with the base model on MATH-500~\#80, a counting problem over a parameterized logarithmic equation whose answer is $501$. Our model solves the problem in $7/8$ rollouts, whereas the base model solves it in $0/8$.

Both models reduce the equation to $x^{2}+(4-k)x+4=0$ and consider the discriminant condition $\Delta=0$, which only gives $k\in\{0,8\}$. The base model stops there and answers $2$. In contrast, our model identifies a second case: the quadratic may have two real roots, but only one lies in the logarithm domain ($kx>0$, $x>-2$). It then splits into $k>0$ and $k<0$, uses Vieta's formulas and the sign of $f(-2)$ to verify the valid-domain cases, and concludes that every negative integer $k$ also works, yielding $500+1=501$.
This example suggests that training on \textsc{ScaleLogic} encourages a more systematic branching strategy: the model searches for hidden cases, enumerates alternatives, and verifies each branch under explicit constraints.

\newpage

\begin{tcolorbox}[
    enhanced, breakable,
    colback=gray!5, colframe=gray!50, boxrule=0.6pt, arc=2mm,
    title=\textbf{Question (MATH-500 \#80)},
    fonttitle=\small\bfseries, coltitle=black, colbacktitle=gray!18,
    left=6pt, right=6pt, top=4pt, bottom=4pt,
]
\footnotesize
Find the number of integer values of $k$ in the closed interval $[-500,500]$
for which the equation $\log(kx) = 2\log(x+2)$ has exactly one real solution.
\smallskip\\
\textbf{Gold answer:} $501$
\end{tcolorbox}

\vspace{4pt}

\begin{tcolorbox}[
    enhanced, breakable,
    colback=green!4, colframe=green!55!black, boxrule=0.7pt, arc=2mm,
    title={\textbf{Ours (Qwen3-4B + RL on synthetic logic)~~7/8 correct}},
    fonttitle=\small\bfseries, coltitle=white,
    colbacktitle=green!55!black,
    left=6pt, right=6pt, top=4pt, bottom=4pt,
]
\footnotesize

We are given the equation $\log(kx) = 2\log(x+2)$. We are to find the number of integer values of $k \in [-500,500]$ for which this equation has \textbf{exactly one real solution}.

\medskip
\textbf{Step 1: Understand the domain}\par
The logarithm is only defined for positive arguments, so $kx>0$ and $x+2>0\Rightarrow x>-2$. Depending on the sign of $k$:
\begin{itemize}\setlength{\itemsep}{0pt}
\item If $k>0$, then $x>0$.
\item If $k<0$, then $-2<x<0$.
\end{itemize}
If $k=0$, then $\log(kx)=\log 0$ is undefined, so $k=0$ is \textbf{not allowed}.

\medskip
\textbf{Step 2: Solve the equation}\par
$\log(kx)=\log((x+2)^2)$, so if both sides are defined, $kx=(x+2)^2$, giving
\[ x^2+(4-k)x+4=0. \]
For ``exactly one real solution'' we set the discriminant to zero:
\[ \Delta=(4-k)^2-16=0 \;\Rightarrow\; k\in\{0,8\}. \]
$k=0$ is invalid; only $k=8$ remains. \emph{But this seems too few.}

Wait --- but maybe there's more to it. Even if the discriminant is \textbf{not zero}, the quadratic might still have \textbf{exactly one real solution in the domain}: if the quadratic has two real roots, only one of them might lie in the domain of the original equation. So we need to be more careful.

\medskip
\textbf{Step 3: Re-express the problem}\par
The number of real solutions of the original equation equals the number of roots of $x^2+(4-k)x+4=0$ that lie in the domain. Two cases must be checked:
\begin{enumerate}\setlength{\itemsep}{0pt}
\item Discriminant $=0$, and the unique root is in the domain.
\item Discriminant $>0$, and \textbf{only one} of the two roots is in the domain.
\end{enumerate}

\medskip
\textbf{Case 1: Discriminant is zero $\Rightarrow$ one real solution}\par
$k=0$: invalid. $k=8$: $(x-2)^2=0 \Rightarrow x=2$, which lies in the domain $x>0$. So $k=8$ is valid. [$\dots$]

\medskip
\textbf{Case 2: Quadratic has two real solutions, but only one in the domain}\par
By Vieta's formulas, $x_1+x_2=k-4$ and $x_1 x_2=4>0$, so both roots share a sign.

\medskip
\textbf{Subcase A: $k>0$.}\par
Both roots are positive, so both lie in the domain $x>0$ $\Rightarrow$ \textbf{two solutions}, not acceptable. [$\dots$] Therefore no $k>0$ works.

\medskip
\textbf{Subcase B: $k<0$.}\par
Sum $k-4<-4$, product $4>0$, so both roots are \textbf{negative}. The domain is $-2<x<0$. We want exactly one root in $(-2,0)$ and the other $\le -2$.

Let $f(x)=x^2+(4-k)x+4$. Since the parabola opens upward and both roots are negative, the desired configuration is equivalent to $f(-2)<0$. Compute:
\[ f(-2)=4-2(4-k)+4=2k. \]
Hence $f(-2)<0 \Leftrightarrow k<0$. So \textbf{for every $k<0$}, the equation has exactly one real solution. [$\dots$]

\medskip
\textbf{Final Answer.}\par
The valid integers $k\in[-500,500]$ are:
\begin{itemize}\setlength{\itemsep}{0pt}
\item All negative integers from $-500$ to $-1$: $500$ values.
\item Plus $k=8$ from Case 1.
\end{itemize}
\[ \boxed{500+1=501} \]

\end{tcolorbox}

\vspace{4pt}

\begin{tcolorbox}[
    enhanced, breakable,
    colback=red!4, colframe=red!55!black, boxrule=0.7pt, arc=2mm,
    title={\textbf{Base (Qwen3-4B, no RL)~~0/8 correct}},
    fonttitle=\small\bfseries, coltitle=white,
    colbacktitle=red!55!black,
    left=6pt, right=6pt, top=4pt, bottom=4pt,
]
\footnotesize

We are given $\log(kx)=2\log(x+2)$ and we are to find the number of \textbf{integer values of $k$} in $[-500,500]$ for which the equation has \textbf{exactly one real solution}.

\medskip
\textbf{Step 1: Understand the equation}\par
Using $\log(a)=\log(b)\Rightarrow a=b$ we obtain
\[ kx=(x+2)^2.\]

[$\dots$]

\medskip
\textbf{Step 2: Solve the equation}\par
$x^2+(4-k)x+4=0$. We want exactly one real solution, which happens when the \textbf{discriminant is zero}:
\[ \Delta=(4-k)^2-16=0 \;\Rightarrow\; 4-k=\pm 4. \]
Solving: $k=0$ or $k=8$.

\textcolor{red!75!black}{So the values of $k$ for which the equation has exactly one real solution are
\[ k=0\quad\text{and}\quad k=8. \]}

\medskip
\textbf{Step 3: Check if these values are in the interval $[-500,500]$}\par
\textcolor{red!75!black}{Both $0$ and $8$ are in $[-500,500]$, so they are valid.}

\medskip
\textbf{Step 4: Are there any other values of $k$ that could make the equation have exactly one real solution?}\par
\textcolor{red!75!black}{We need to make sure that \textbf{only} when the discriminant is zero does the equation have exactly one real solution. In general, a quadratic has two real solutions if $\Delta>0$, one if $\Delta=0$, none if $\Delta<0$. So the only values of $k$ are those for which the discriminant is zero, namely $k=0$ and $k=8$.}

\medskip
\textbf{Final Answer.}
\[ \textcolor{red!75!black}{\boxed{2}} \]

\end{tcolorbox}

\medskip
\refstepcounter{figure}
\noindent Figure \thefigure:\;
MATH-500 \#80: full reasoning trace of our model (top, $7/8$ correct) and the base model (bottom, $0/8$ correct). Wrong steps in the base model are highlighted in red; ``$[\dots]$'' marks omitted verbose passages.
\label{fig:qualitative_example}